\RequirePackage{fix-cm}
\documentclass[natbib,twocolumn]{svjour3}          % twocolumn
\smartqed  % flush right qed marks, e.g. at end of proof
\usepackage{graphicx}
\usepackage{xcolor}
\usepackage{subfig}
\usepackage{multirow}%
\usepackage{color, colortbl}
\usepackage{amsmath,amssymb,amsfonts}%
\usepackage{mathrsfs}%
\usepackage[title]{appendix}%
\usepackage{xcolor}%
\usepackage{textcomp}%
\usepackage{manyfoot}%
\usepackage{booktabs}%
\usepackage{algorithm}%`
\usepackage{algorithmicx}%
\usepackage{algpseudocode}%
\usepackage{listings}%
\usepackage{tabularx}
\usepackage{makecell}
\usepackage{natbib}

\usepackage{rotating}
\usepackage{float}
\usepackage{bm}
\usepackage{enumitem}
\usepackage{balance}
\usepackage[utf8]{inputenc}
\usepackage{subcaption}
\usepackage{array}
\usepackage[left=2cm,right=2cm,top=2cm,bottom=3cm]{geometry}
\usepackage{todonotes}

\definecolor{seal}{HTML}{B711AC}
\usepackage[colorlinks=true, linkcolor=blue, filecolor=magenta, citecolor=seal]{hyperref}

\usepackage{lipsum}
\makeatletter
\def\myaddcontentsline#1#2#3{%
  \addtocontents{#1}{\protect\contentsline{#2}{#3}{sec \thesection\ at p. \thepage}{}}}
\renewcommand{\@todonotes@addElementToListOfTodos}{%
    \if@todonotes@colorinlistoftodos%
        \myaddcontentsline{tdo}{todo}{{%
            \colorbox{\@todonotes@currentbackgroundcolor}%
                {\textcolor{\@todonotes@currentbackgroundcolor}{o}}%
            \ \@todonotes@caption}}%
    \else%
        \myaddcontentsline{tdo}{todo}{{\@todonotes@caption}}%
    \fi}%
\newcommand*\mylistoftodos{%
  \begingroup
       \setbox\@tempboxa\hbox{see 9.9 at p. 99}%
       \renewcommand*\@tocrmarg{\the\wd\@tempboxa}%
       \renewcommand*\@pnumwidth{\the\wd\@tempboxa}%
       \listoftodos%
  \endgroup
}
\makeatother
\newcolumntype{C}[1]{>{\centering\arraybackslash}p{#1}}
\setlist[itemize]{leftmargin=*,noitemsep,topsep=0pt}

\usepackage{newunicodechar}
\newunicodechar{，}{,}
\newunicodechar{；}{;}
\newunicodechar{。}{.}
\title{In Search of Lost Online Test-time Adaptation: A Survey}

\author{Zixin Wang\textsuperscript{$\dag$} \and Yadan Luo\textsuperscript{*$\dag$} \and Liang Zheng\textsuperscript{$\ddag$} \and Zhuoxiao Chen\textsuperscript{$\dag$} \\ \and Sen Wang\textsuperscript{$\dag$}  \and Zi Huang\textsuperscript{$\dag$}}

\institute{
$^1$The University of Queensland, Australia \\ $^2$The Australian National University, Australia \\ $^*$ Corresponding author\\
\email{{\{zixin.wang, y.luo, zhuoxiao.chen, sen.wang, helen.huang\}@uq.edu.au}, liang.zheng@anu.edu.au} 
}

\date{Received: date / Accepted: date}
\begin{document}
\begin{sloppypar}

\clearpage

\maketitle

\begin{abstract}
This article presents a comprehensive survey of online test-time adaptation (OTTA), focusing on effectively adapting machine learning models to distributionally different target data upon batch arrival. Despite the recent proliferation of OTTA methods, conclusions from previous studies are inconsistent due to ambiguous settings, outdated backbones, and inconsistent hyperparameter tuning, which obscure core challenges and hinder reproducibility. To enhance clarity and enable rigorous comparison, we classify OTTA techniques into three primary categories and benchmark them using a modern backbone, the Vision Transformer (ViT). Our benchmarks cover conventional corrupted datasets such as CIFAR-10/100-C and ImageNet-C, as well as real-world shifts represented by CIFAR-10.1, OfficeHome, and CIFAR-10-Warehouse. The CIFAR-10-Warehouse dataset includes a variety of variations from different search engines and synthesized data generated through diffusion models. To measure efficiency in online scenarios, we introduce novel evaluation metrics, including GFLOPs, wall clock time, and GPU memory usage, providing a clearer picture of the trade-offs between adaptation accuracy and computational overhead. Our findings diverge from existing literature, revealing that (1) transformers demonstrate heightened resilience to diverse domain shifts, (2) the efficacy of many OTTA methods relies on large batch sizes, and (3) stability in optimization and resistance to perturbations are crucial during adaptation, particularly when the batch size is 1. Based on these insights, we highlight promising directions for future research. Our benchmarking toolkit and source code are available at \url{https://github.com/Jo-wang/OTTA_ViT_survey}.
\end{abstract}

% \maketitle

\section{Introduction}\label{intro}

Dataset shift \citep{quinonero2008dataset} poses a notable challenge for machine learning. Models often experience significant performance drops when confronting test data characterized by superior \textit{distribution differences} from training. Such differences might come from changes in style and lighting conditions and various forms of corruption, making test data deviate from the data upon which these models were initially trained. To mitigate the performance degradation brought by these shifts, test-time adaptation (TTA) has emerged as a promising solution. TTA aims to rectify the dataset shift issue by adapting the model to unseen distributions using unlabeled test data \citep{DBLP:journals/corr/abs-2303-15361}. Different from unsupervised domain adaptation \citep{DBLP:conf/icml/GaninL15,  DBLP:journals/ijon/WangD18}, TTA does not require access to source data for distribution alignment. Commonly used strategies in TTA include unsupervised proxy objectives, spanning techniques such as pseudo-labeling \cite{DBLP:conf/icml/LiangHF20}, graph-based learning \citep{DBLP:journals/pami/LuoWCHB23}, and contrastive learning \cite{DBLP:conf/cvpr/0001WDE22}, applied on the test data through multiple training epochs to enhance model accuracy, such as autonomous vehicle detection \citep{hegde2021uncertainty}, pose estimation \citep{lee2023tta}, video depth prediction \citep{liu2023meta}, frame interpolation \citep{choi2021test}, and medical diagnosis \citep{ma2022test, wang2022metateacher, saltori2022gipso}. Nevertheless, requiring access to the complete test set at every time step may not always align with practical use. In many applications, such as autonomous driving, adaptation is restricted to using only the \textbf{current test batch} processed in a streaming manner. Such operational restrictions make it untenable for TTA to require the full test set tenable. 

In this study, our focus is on a specific line of TTA methods, \textit{i.e.}, online test-time adaptation (OTTA), which aims to accommodate \textit{real-time changes} in the test data distribution. We provide a comprehensive overview of existing OTTA studies and evaluate the efficiency and effectiveness of these methods and their individual components. To facilitate a structured comprehension of the OTTA landscape, we categorize existing approaches into three groups: optimization-oriented OTTA, data-centric OTTA, and model-based OTTA.
\begin{itemize}
    \item \textbf{Optimization-based OTTA} focuses on various optimization methods. Examples are designing new loss functions, updating normalization layers (e.g., \texttt{BatchNorm}) during testing, using pseudo-labeling strategies, teacher-student frameworks, and contrastive learning-based approaches, \textit{etc}.
    \item \textbf{Data-based OTTA} maximizes prediction consistency across diversified test data. Diversification strategies use auxiliary data, improved data augmentation methods, and diffusion techniques and create a saving queue for test data, \textit{etc}.
    \item \textbf{Model-based OTTA} adjusts the original backbone, such as modifying specific layers or their mechanisms, adding supplementary branches, and incorporating prompts.
\end{itemize}

It is potentially useful to combine methods from different categories for further improvement. An in-depth analysis of this strategy is presented in Sec. \ref{sec:OTTA}. Note that this survey does not include the paper if source-stage customization is needed, such as \citet{DBLP:conf/acml/ThopalliTT22, DBLP:conf/cvpr/DoblerM023, DBLP:conf/cvpr/BrahmaR23, DBLP:journals/corr/abs-2206-00205, DBLP:conf/icip/AdachiYK23, DBLP:conf/iclr/LimKCC23, DBLP:conf/eccv/ChoiYCY22, DBLP:journals/corr/abs-2304-10113, DBLP:journals/corr/abs-2208-07736, DBLP:conf/cvpr/GaoZLDSW23}. As it necessitates extra operation or information at the source pre-training stage, this setup may not be feasible in some scenarios and may prevent us from comparing on a consistent standard.

\textbf{Differences from an existing survey. }\citet{DBLP:journals/corr/abs-2303-15361} provided a comprehensive overview of the vast topic of test-time adaptation (TTA), discussing TTAs in diverse configurations and their applicability in vision, natural language processing (NLP), and graph data analysis. One limitation is that the survey did not provide experimental comparisons of existing methods. \citet{DBLP:journals/corr/abs-2310-02416} studied fully test-time adaptation for some specific components, e.g., batch normalization, calibration, class re-balancing, \textit{etc}. However, it did not focus too much on analyzing the existing methods and exploring ViTs. \citet{DBLP:conf/wacv/MarsdenD024} conducted comprehensive study on universal test-time adaptation setting. In contrast, our survey concentrates on purely online TTA approaches and provides valuable insight from experimental comparisons, considering various domain shifts, hyperparameter selection, and backbone influence \citep{DBLP:conf/icml/ZhaoLAL23}.

\textbf{Contributions.} As Vision Transformer (ViT) architectures gain increasing prominence, a critical question emerges: \textit{ As Vision Transformer (ViT) architectures gain increasing prominence, a critical question arises: Do OTTA strategies, originally devised for CNNs, retain their effectiveness when applied to ViT models?} This question stems from the significant architectural differences between ViTs and conventional CNNs, such as ResNets, particularly in their normalization layers and information processing mechanisms. Given the growing adoption of ViTs, investigating their compatibility with existing OTTA strategies is essential. To thoroughly explore this question, we evaluate eight representative OTTA algorithms across diverse distribution shifts, employing a set of metrics to evaluate both effectiveness and efficiency. Below, we summarize the key contribution of this survey:

\begin{itemize}
 \item \textbf{[A focused OTTA survey]} To the best of our knowledge, this is the first focused survey on online test-time adaptation, which provides a thorough understanding of three main working mechanisms. Wide experimental investigations are conducted in a fair comparison setting.
  \item \textbf{[Comprehensive Benchmarking and Adaptation of OTTA Strategies with ViT]} We reimplemented representative OTTA baselines under the ViT architecture and testified their performance against six benchmark datasets. We drive a set of replacement rules that adapt the existing OTTA methods to accommodate the new backbone.
  \item \textbf{[Both accuracy and efficiency as evaluation Metrics]} Apart from using the traditional recognition accuracy metric, we further provide insights into various facets of computational efficiency by Giga floating-point operations per second (GFLOPs), wall clock time, and GPU memory usage. These metrics are important in real-time streaming applications and can be treated as a supplementary of \citep{DBLP:conf/wacv/MarsdenD024}.
  \item \textbf{[Real-world testbeds]} While existing literature extensively explores OTTA methods on corruption datasets like CIFAR-10-C, CIFAR-100-C, and ImageNet-C, or even real-world datasets \citep{DBLP:conf/wacv/MarsdenD024} our interests are more fall into their capability to navigate real-world dataset shifts from different angles. Specifically, we assess OTTA performance on CIFAR-10-Warehouse, a newly introduced, expansive test set of CIFAR-10. Conclusions about various domain shifts, based on the same pre-trained model, are drawn from our empirical analysis and evaluation. These shifts were previously unexplored in the existing survey \citep{DBLP:journals/corr/abs-2303-15361}.
\end{itemize}

This work aims to summarize existing OTTA methods with the aforementioned three categorization criteria and analyze some representative approaches by empirical results. Moreover, to assess real-world potential, we conduct comparative experiments to explore the portability, robustness, and environmental sensitivity of the OTTA components. We expect this survey to offer a systematic perspective in navigating OTTA's intricate and diverse solutions, enabling a clear identification of effective components. We also present new challenges as potential future research directions.

\begin{table*}[t]
\begin{center}
\caption{ \textbf{Datasets used in this survey}. We list their key statistics. 
\label{table:datasets} 
} 
\setlength{\tabcolsep}{2mm}{
\begin{tabular}{ l c  c  c  c  c } 
\toprule
Datasets &  $\#$ domains &$\#$ images   & $\#$ classes & corrupted?   & image size  \\
\midrule
\midrule
CIFAR-10-C \citep{hendrycks2018benchmarking} & 19 & 950,000 & 10 & Yes &  $32\times32$ \\
CIFAR-100-C \citep{hendrycks2018benchmarking} & 19 & 950,000 & 100 & Yes &  $32\times32$ \\
ImageNet-C \citep{hendrycks2018benchmarking} & 19 & 4,750,000 & 1000 & Yes &  $224\times224$ \\
\midrule
CIFAR-10.1 \citep{DBLP:journals/corr/abs-1806-00451} & 1 &  2,000 & 10 & {No} & $32\times32$ \\
CIFAR-10-Warehouse \citep{DBLP:journals/corr/abs-2310-04414} & 180 &608,691 & 10 & {No} & $224\times224$\\
OfficeHome \citep{DBLP:conf/cvpr/VenkateswaraECP17} & 4 & 15,500 & 65 & No & non-uniform \\
\bottomrule 
\end{tabular}}
\end{center}
% \vspace{-1.5em}
\end{table*}

\begin{figure}[t]
\centering
\includegraphics[width=0.95\linewidth]{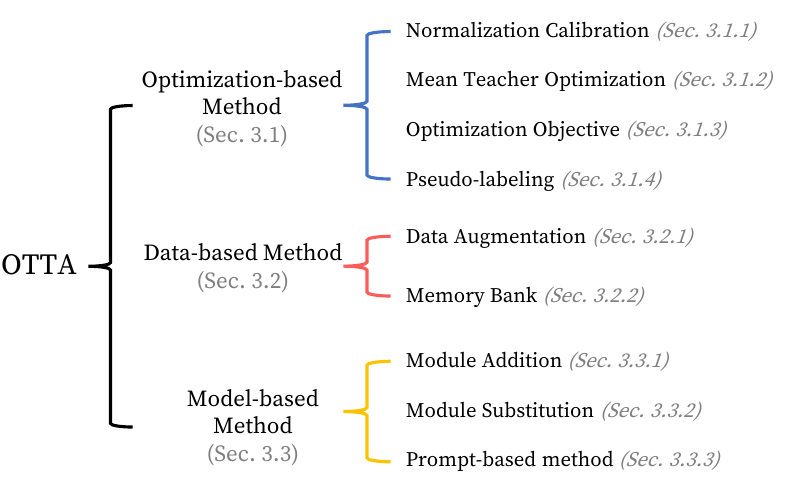}
\caption{Taxonomy of existing OTTA methods. The categories, i.e., optimization-based, data-based, and model-based, inform three mainstream working mechanisms. To provide a clear illustration, methods related to prompts are categorized into model-based methods.}\label{fig:tax}
\end{figure}

\textbf{Organization of the survey.} The rest of this survey will be organized as follows. Sec. \ref{sec:Problem} presents the problem definition and introduces widely used datasets, metrics, and applications. Using the taxonomy shown in Fig. \ref{fig:tax}, Sec. \ref{sec:OTTA} comprehensively reviews existing OTTA methods. With the new generation of backbones, Sec. \ref{sec:Exp} empirically analyzes eight state-of-the-art methods based on multiple evaluation metrics on corrupted and real-world distribution shifts. We introduce the potential future directions in Sec. \ref{sec:future} and conclude this survey in Sec. \ref{sec:conclude}.
% \vspace{-2ex}
\section{Problem Overview}\label{sec:Problem}
% \vspace{-1ex}
Online Test-time Adaptation (OTTA), with its online and real-time characteristics, represents a critical line of methods in test-time adaptation. This section %aims to elucidate the essence of OTTA by 
providing a formal definition of OTTA and delving into its fundamental attributes. Furthermore, we explore widely used datasets and evaluation methods and examine the potential application scenarios of OTTA. %
A comparative analysis is undertaken to differentiate OTTA from similar settings to ensure a clear understanding.
% \vspace{-2ex}
\subsection{Problem Definition}
In OTTA, we assume access to a trained source model and adapt the model at test time over the test input to make the final prediction. The given source model $f_{\theta^S}$ parameterized by $\theta^S$ is pre-trained on a labeled source domain $\mathcal{D}_S=\{(\boldsymbol{x}^S, \boldsymbol{y}^S)\}$, which is formed by i.i.d. sampling from the source distribution $p_S$. Unlabeled test data $\mathcal{D}_T=\{\boldsymbol{x}_1^T, \boldsymbol{x}_2^T, \boldsymbol{x}_t^T,\ldots, \boldsymbol{x}_n^T\}$ come in batches, where $t$ indicates the current time step, $n$ is the total number of time steps (i.e., number of batches). Test data often come from one or multiple different distributions $(\boldsymbol{x}_t^T, \boldsymbol{y}_t^T)\sim p_T$, where $p_S(\boldsymbol{x}, y) \neq p_T(\boldsymbol{x}, y)$ under the covariate shift assumption \citep{DBLP:conf/nips/HuangSGBS06}. During TTA, we update the model parameters for batch $t$, resulting in an adapted model $f_{\theta^t}$. Before adaptation, the pre-trained model is expected to retain its original architecture, especially the backbone, without modifying its layers or introducing new model branches during training. Additionally, the model is restricted to observing the test data only once and must produce predictions promptly online. By refining the definition of OTTA in this manner, we aim to minimize limitations associated with its application in real-world settings. Note that the model is reset to its original pre-trained state after being adapted to a specific domain. \textit{i.e.}, $\underline{f_{\theta^S}} \rightarrow f_{\theta^0} \rightarrow \underline{f_{\theta^S}} \rightarrow f_{\theta^1} \rightarrow \underline{f_{\theta^S}} \rightarrow \cdots f_{\theta^t}$. 

Due to the covariate shift between the source and test data, adapting the source model without the source data poses a significant challenge. Since there is no way to align these two sets as in unsupervised domain adaptation, one may ask, what kind of optimization objective could work under such a limited environment? Meanwhile, as the test data come at a fixed pace, how many could be a desirable amount to best fix the test-time adaptation? Or will the adaptation even work in the new era of backbones (e.g., ViTs)? Does ``Test-time Adaptation'' become a false proposition with the backbone changes? With these concerns, we unfold the OTTA methods by their datasets, evaluations, and applications and decouple their strategies, aiming to discover which one and why one component would work with the update of existing backbones.

\subsection{Datasets}
This survey mainly summarizes datasets in image classification, a fundamental problem in computer vision, while recognizing that OTTA has been applied to many downstream tasks \citep{ma2022test, ding2023maps, saltori2022gipso}.  
Testbeds in OTTA usually seek to facilitate adaptation from natural images to corrupted ones. The latter are created by perturbations such as Gaussian noise and Defocus blur. Despite including corruptions at varying severities, these synthetically induced corruptions may not sufficiently mirror the authentic domain shift encountered in real-world scenarios. Our work uses corruption \citep{croce2021robustbench}, generated images, and real-world shift datasets, summarized in Table \ref{table:datasets}. Details of each testbed are described below. 

\textbf{CIFAR-10-C} is a standard benchmark for image classification. It contains 950,000 color images, each of 32x32 pixels, spanning ten distinct classes. CIFAR-10-C retains the class structure of CIFAR-10 but incorporates 19 diverse corruption styles, with severities ranging from levels 1 to 5. This corrupted variant aims to simulate realistic image distortions or corruptions that might arise during processes like image acquisition, storage, or transmission.
    
\textbf{CIFAR-100-C} has 950,000 colored images with dimensions 32x32 pixels, uniformly distributed across 100 unique classes. The CIFAR-100 Corrupted dataset, analogous to CIFAR-10-C, integrates artificial corruptions into the canonical CIFAR-100 images.
    
\textbf{ImageNet-C} is a corrupted version of ImageNet test set \citep{krizhevsky2012imagenet}. Produced from ImageNet-1k, ImageNet-C has a similar setup to the CIFAR-10-C and CIFAR-100-C corruption types. For each domain, $5$ levels of severity are produced, with $50,000$ images per severity from $1,000$ classes.

\textbf{CIFAR-10.1} \citep{DBLP:journals/corr/abs-1806-00451} is a real-world test set of CIFAR-10. It contains roughly 2,000 images sampled from the Tiny Image dataset \citep{DBLP:journals/pieee/YangMM16}.

\textbf{CIFAR-10-Warehouse} \citep{DBLP:journals/corr/abs-2310-04414} integrates images from both diffusion model (\textit{i.e.,} Stable Diffusion-2-1 \citep{Rombach_2022_CVPR}) and targeted keyword searches across eight popular search engines. The diffusion model uses the prompt ``high quality photo of {color} {class name}", with color chosen from 12 options. The dataset comprises 37 generated and 143 real-world subsets, each containing between 300 and 8,000 images. 

\textbf{OfficeHome} is a widely used benchmark in domain adaptation and domain generalization tasks. It has $65$ classes within $4$ domains: Artistic images (Art), Clip Art, Product images (Product), and Real-World images (RealWorld). Each domain represents a significantly different style of images.

\subsection{Evaluation}
Efficiency and accuracy are crucial for online test-time adaptation to reveal the efficiency of OTTA faithfully. This survey employs the following evaluation metrics:

\noindent\textbf{Mean error (mErr)} is one of the most commonly used metrics to assess model accuracy. It computes the average error rate across all corruption types or domains. While useful, this metric usually does not provide class-specific insights in OTTA.

\noindent\textbf{GFLOPs} refers to giga floating point operations per second, which quantifies the number of floating-point calculations a model performs in a second. A model with lower GFLOPs is more computationally efficient.

\noindent \textbf{Wall-clock time} measures the actual time taken by the model to complete the adaptation. This may significantly influence the use case of an OTTA model.

\noindent \textbf{GPU memory usage} refers to the amount of memory the model uses while running on a GPU. A model with lower GPU memory usage is more applicable and can be deployed on a wider range of devices.

\subsection{Relationship with Other Tasks}
\textbf{Offline test-time adaptation (TTA)} \citep{DBLP:conf/icml/LiangHF20, DBLP:journals/pami/LiangHWHF22, DBLP:conf/cvpr/DingXT0WT22, DBLP:conf/nips/YangWWHJ21} is a technique to adapt a source pre-trained model to the target (\textit{i.e.}, test) set. This task assumes that the model has access to the entire dataset multiple times, which differs from the online test-time adaptation where the test data is given in batches.

\noindent\textbf{Continual TTA}
Contrary to the classic OTTA setup, where adaptation could be envisioned as occurring in discrete steps corresponding to distinct domain shifts, continual TTA \citep{DBLP:conf/cvpr/0013FGD22, DBLP:conf/iclr/HongLZS23, DBLP:journals/corr/abs-2303-01904, DBLP:journals/corr/abs-2304-10113, DBLP:conf/aaai/GanBLMZSL23} operates under the premise of seamless, continuously adapting to new data distributions. This process does not need a reset of the model with each perceived domain shift. Rather, it highlights the importance of a model's ability to autonomously update and refine its parameters in response to ongoing changes in the data landscape, without explicit indicators of domain boundaries.

\noindent\textbf{Gradual TTA} tackles real-world scenarios where domain shifts are gradually introduced through incoming test samples \citep{DBLP:journals/corr/abs-2208-07736, DBLP:conf/cvpr/DoblerM023}. An example is the gradual and continuous change in weather conditions. For corruption datasets, existing gradual TTA approaches assume that test data transition from severity level 1 to level 2 and then progress slowly toward the highest level. Continual and gradual TTA methods also support online test-time adaptation.

\begin{figure*}[t]
\centering
\includegraphics[width=0.96\linewidth]{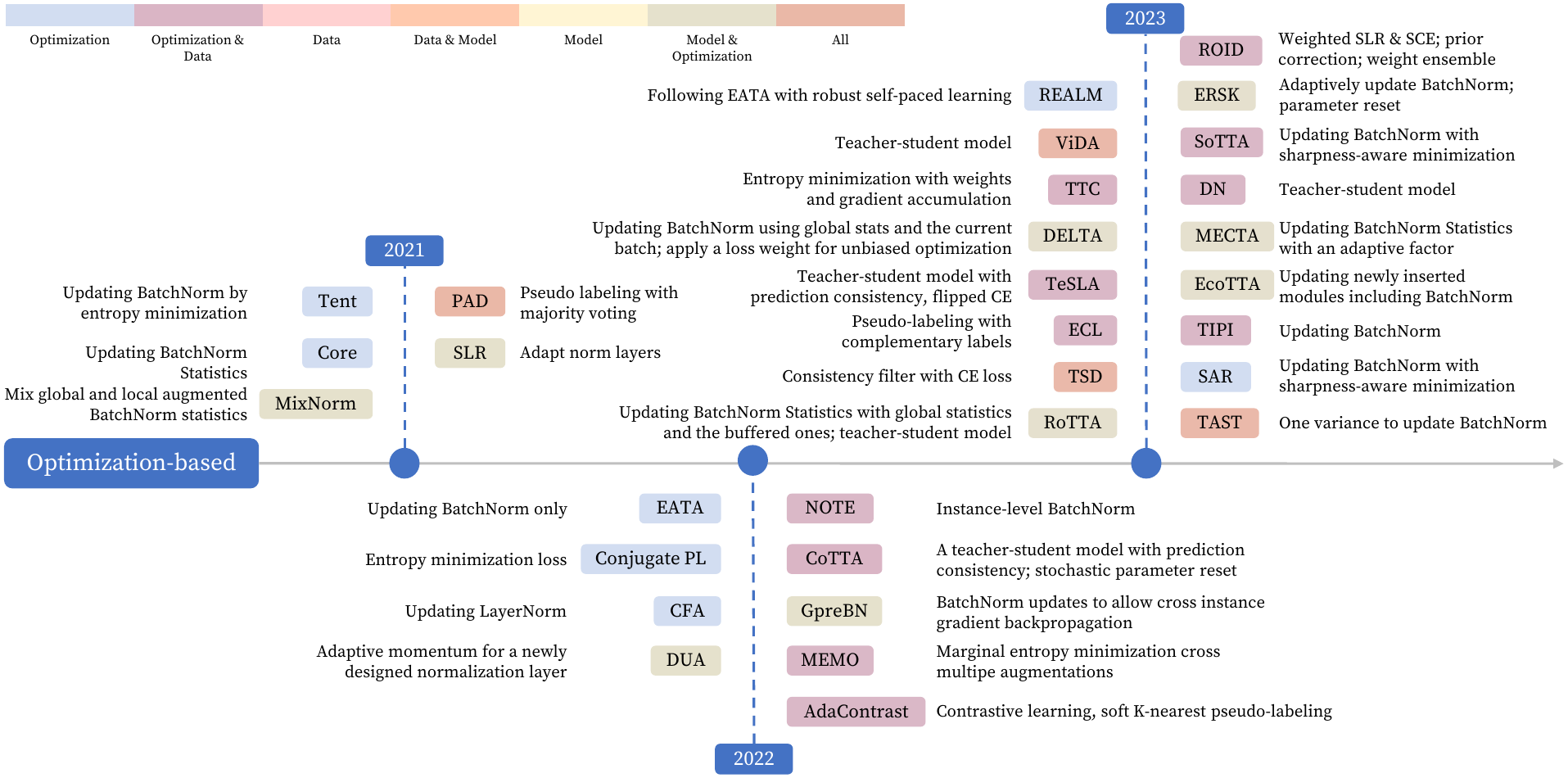}
\caption{Timeline of optimization-based OTTA methods. }\label{fig:opt-based}
\end{figure*}

\noindent\textbf{Test-time Training (TTT)} introduces an auxiliary task for both training and adaptation \citep{DBLP:conf/icml/SunWLMEH20, DBLP:conf/nips/GandelsmanSCE22}. In the training phase, the original architecture, such as ResNet101 \citep{DBLP:conf/cvpr/HeZRS16}, is modified into a ``Y"-shaped structure, where one task is image classification, and the other could be rotation prediction. During adaptation, the auxiliary task continues to be trained in a supervised manner so that model parameters are updated. The classification head output serves as the final prediction for each test sample.

\noindent\textbf{Test-time augmentation (TTAug)}  applies data augmentations to input data during inference, resulting in multiple variations of the same test sample, from which predictions are obtained \citep{DBLP:conf/iccv/ShanmugamBBG21, DBLP:conf/iconip/Kimura21}. The final prediction typically aggregates predictions of these augmented samples through averaging or majority voting. TTAug enhances model prediction performance by providing a range of data views. This technique can be applied to various tasks, including domain adaptation, offline TTA, and even OTTA, as TTAug does not require any modification of the model training process.

%\noindent
\noindent\textbf{Domain generalization} \citep{DBLP:conf/cvpr/QiaoZP20, DBLP:conf/iccv/WangLQHB21, DBLP:conf/cvpr/XuZ0W021, DBLP:journals/pami/ZhouLQXL23} aims to train models that can perform effectively across multiple distinct domains without specific adaptation to any domain. It assumes the model learns domain-invariant features that are applicable across diverse datasets. While OTTA emphasizes dynamic adaptation to specific domains over time, domain generalization seeks to establish domain-agnostic representations. The choice between these strategies depends on the specific problem considered. 

\section{Online Test-time Adaptation}\label{sec:OTTA}
Given the distribution divergence of online data from source training data, OTTA techniques are broadly classified into three categories that hinge on their responses to two primary concerns: managing online data and mitigating performance drops due to distribution shifts. \textbf{Optimization-based} methods anchored in designing unsupervised objectives typically lean towards adjusting or enhancing pre-trained models. \textbf{Model-based} approaches look to modify or introduce particular layers. On the other hand, \textbf{data-based} methods aim to expand data diversity, either to improve model generalization or to harmonize consistency across data views. According to this taxonomy, we sort out existing approaches in Table \ref{tab. otta_methods} and review them in detail below.

\subsection{Optimization-based OTTA} 
Optimization-based OTTA methods consist of three sub-categories: (1) recalibrating statistics in normalization layers, (2) enhancing optimization stability with the mean-teacher model, and (3) designing unsupervised loss functions. A timeline is illustrated in Fig. \ref{fig:opt-based}.

\subsubsection{Normalization Calibration}\label{norm-based}
In deep learning, a normalization layer aims to improve the training process and enhance the generalization capacity of neural networks by regulating the statistical properties of activations within a given layer. Batch normalization (\texttt{BatchNorm}) \citep{DBLP:conf/icml/IoffeS15}, as the most commonly used normalization layer, aims to stabilize the training process by its global statistics or a considerably large batch size. Operating by standardizing the mean and variance of activations, \texttt{BatchNorm} could also reduce the risk of vanishing or exploding gradients during training. There are also alternatives to \texttt{BatchNorm}, such as layer normalization (\texttt{LayerNorm}) \citep{DBLP:journals/corr/BaKH16}, group normalization (\texttt{GroupNorm}) \citep{DBLP:journals/ijcv/WuH20}, and instance normalization (\texttt{InstanceNorm}) \citep{DBLP:journals/corr/UlyanovVL16} A similar idea to the normalization layer is feature whitening, which also adjusts features right after the activation layer. Both strategies are commonly used in domain adaptation literature \citep{DBLP:conf/cvpr/RoySSBS019, DBLP:conf/iccv/CarlucciPCRB17}.

\begin{figure}
\centering
\includegraphics[width=0.94\linewidth]{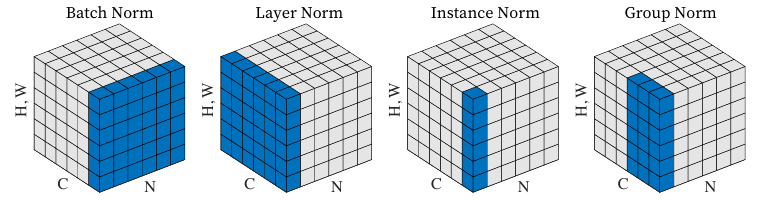}
\caption{Visualizing normalization layers \citep{DBLP:journals/ijcv/WuH20}.}\label{fig:norm}
% \vspace{-0.5cm}
\end{figure}
\noindent \textbf{Example.} Take the most commonly used \texttt{BatchNorm} as an example. Let $\bm x_i$ represent the activation for feature channel $i$ in a mini-batch. The \texttt{BatchNorm} layer will first calculate the batch-level mean $\bm \mu$ and variance $\bm \sigma ^2$ by: 
\begin{equation}
\quad\bm \mu = \frac{1}{m} \sum_{i=1}^{m} \bm x_i, \quad\bm \sigma^2 = \frac{1}{m} \sum_{i=1}^{m} (\bm x_i - \bm \mu)^2,
\end{equation}
where $m$ is the mini-batch size. Then, the calculated statistics will be applied to standardize the inputs:
\begin{equation}\label{ed: transf_BN}
\quad\hat{\bm x}_i = \frac{\bm x_i - \bm \mu}{\sqrt{\bm \sigma^2 + \epsilon}},\quad \bm y_i = \gamma \hat{\bm x}_i + \beta,
\end{equation}
where $y_i$ is the final output of the $i$-th channel from this batch normalization layer, adjusting two learnable affine parameters, $\gamma$ and $\beta$. And $\epsilon$ is used to avoid division of $0$. For the update, the running mean $\bm \mu^\text{run}$ and variance $\sigma^\text{run}$ are computed as a moving average of the mean and variance over all batches seen during training, with a momentum factor $\alpha$:
\begin{equation}
~\bm \mu^\text{run} = \alpha \bm\mu + (1-\alpha) \bm \mu^\text{run}, \quad\bm \sigma^\text{run} = \alpha \bm \sigma + (1-\alpha) \bm\sigma^\text{run}.
\end{equation}
\textbf{Motivation.} In domain adaptation, aligning batch normalization statistics is shown to mitigate performance degradation brought by covariate shifts. The underlying hypothesis suggests that information about labels is encoded within the weight matrices of each layer. At the same time, knowledge related to specific domains is conveyed through the statistics of \texttt{BatchNorm} layer. Consequently, an adaptation based on updating \texttt{BatchNorm} layer could bring enhanced performance of unseen domains \citep{DBLP:conf/iclr/LiWS0H17}. A similar idea can be used in online test-time adaptation. The assumption is, given a neural network $f$ trained on a source dataset $\mathcal{D}_S$ with \textbf{normalization parameters} $\beta$ and $\gamma$, updating $\{\gamma, \beta\}$ based on test data $\boldsymbol{x}_i^t$ at each time step $t$ will improve $f$'s robustness on the test domain.

\noindent  Building upon the assumption, initial investigations in OTTA predominantly revolved around fine-tuning only on updating the normalization layers. This strategy has a few popular variations. A common practice adjusts statistics ($\mu$ and $\sigma$) and affine parameters ($\beta$ and $\gamma$) in the \texttt{BatchNorm} layer. Note that the choice of normalization techniques, such as \texttt{LayerNorm} or \texttt{GroupNorm}, may depend on the backbone architecture and specific optimization objectives. 

\texttt{Tent} \citep{DBLP:conf/iclr/WangSLOD21} and its subsequent works, such as \citep{DBLP:conf/iclr/JangCC23}, are representative approaches within this paradigm. They update the statistics and affine parameters of \texttt{BatchNorm} for each test batch while freezing the remaining parameters. However, as seen in \texttt{Tent},  updated by minimizing soft entropy, the effectiveness of batch-level updates is dependent on data quality within each batch, introducing potential performance fluctuations. For example, noisy or poisoned data with biased statistics would significantly influence the \texttt{BatchNorm} updates.  Methods that aim at
\noindent\textbf{stabilization via dataset-level estimate} are proposed to mitigate such performance fluctuation arising from batch-level statistics. Gradient preserving batch normalization \texttt{GpreBN} \citep{DBLP:journals/corr/abs-2205-10210} is introduced that allows for cross instance gradient backpropagation by modifying the \texttt{BatchNorm} normalization factor:
\begin{equation}
\quad \hat{\bm y}_i=\frac{\frac{\bm x_i-\bm \mu_c}{\bm \sigma_c} \bar{\bm \sigma}_c+\bar{\bm\mu}_c-\bm\mu}{\bm\sigma} \gamma+\beta,
\end{equation}
where $\frac{\bm x_i-\bm\mu_c}{\bm\sigma_c}$ is the standardized input feature $\hat{\bm x}_i$, same as in Eq. \eqref{ed: transf_BN}. $\bm\sigma_c$ and $\bm\mu_c$ means stop gradient. \texttt{GpreBN} normalizes $\hat{\bm x}_i$ by arbitrary non-learnable parameters $\bm\mu$ and $\bm\sigma$. \texttt{MixNorm} \citep{DBLP:journals/corr/abs-2110-11478} mixies the statistics (produced by augmented sample inputs) of the current batch with the global statistics computed through moving average. Then, combining global-level and augmented batch-level statistics effectively bridges the gap between historical context and real-time fluctuations, enhancing performance regardless of batch size. As an alternative proposal, \texttt{RBN} \citep{DBLP:journals/corr/abs-2303-13899} considers global robust statistics from a well-maintained memory bank with a fixed momentum of moving average when updating statistics to ensure high statistic quality. Similarly, \texttt{Core} \citep{DBLP:journals/corr/abs-2110-04065} incorporates a momentum factor for moving average to fuse the source and test set statistics.

Instead of using a fixed momentum factor for the moving average, \citet{DBLP:conf/cvpr/MirzaMPB22a} propose a dynamic approach. It determines the momentum of the moving average based on a decay factor. Assuming that model performance deteriorates over time, the decay factor should progressively be considered more from the current batch as time advances to avoid biased learning by misled source statistics. \texttt{ERSK} \citep{DBLP:journals/corr/abs-2311-04991} follows a similar idea but determines its momentum by the KL divergence of \texttt{BatchNorm} statistics between source-pretrained model and the current test batch. 

\noindent\textbf{Stabilization via renormalization.} Merely emphasizing moving averages might undermine the inherent characteristics of gradient optimization and normalization when it comes to updating \texttt{BatchNorm} layers. As noted by \citet{DBLP:conf/cvpr/HuangYLD18}, \texttt{BatchNorm} primarily centers and scales activations without addressing their correlation issue of activations, where a decorrelated activation can lead to better feature representation \citep{DBLP:journals/neco/Schmidhuber92c} and generalization \citep{DBLP:journals/corr/CogswellAGZB15}. At the same time, batch size significantly influences correlated activations, which further brings limitations when batch size is small. Test-time batch renormalization module (\texttt{TBR}) in \texttt{DELTA} \citep{DBLP:conf/iclr/Zhao0X23} addresses these limitations through a renormalization process. They adjust standardized outputs using two new parameters, $r$ and $d$. $r=\frac{s g\left(\hat{\bm\sigma}^{\text {batch }}\right)}{\hat{\bm\sigma}^{\text {ema }}}$ and $ d=\frac{s g\left(\hat{\mu}^{\text {batch }}\right)-\hat{\bm\mu}^{\text {ema }}}{\hat{\bm\sigma}^{\text {ema }}}$, where $sg(\cdot)$ is stop gradient. Both parameters are computed using batch and global moving statistics, using a novel approach to maintaining stable batch statistics updating strategies (inspired by \citep{DBLP:conf/nips/Ioffe17}). Then $\hat{\bm x}_i$ is normalized further by $\hat{\bm x}_i=\hat{\bm x}_i \cdot r + d$. The above OTTA methods reset the model for each domain; this limits the model applicability to scenarios that commonly have no clear domain boundary during adaptation. \texttt{NOTE} \citep{DBLP:conf/nips/GongJKKSL22} focuses on continual OTTA under temperal correlaton, \textit{i.e.}, distribution changes over time $t$: $\left(\mathbf{x}_t, \bm y_t\right) \sim P_{\mathcal{T}}(\mathbf{x}, \bm y \mid t)$.  
Authors propose instance-level \texttt{BatchNorm} to avoid potential instance-wise variations in a domain non-identifiable paradigm.

\noindent\textbf{Stabilization via enlarging batches.} To improve the stability of adaptation, another idea is using large batch sizes. %Obviously, having a greater volume of statistical data from a larger batch contributes to the robustness of the overall statistical information. 
In fact, most methods based on batch normalization %(\texttt{BatchNorm}) 
employ substantial batch sizes such as $200$ in \citep{DBLP:conf/iclr/WangSLOD21, DBLP:journals/corr/abs-2110-11478}. Despite its effectiveness, this practice cannot deal with scenarios where data arrives in smaller quantities due to hardware (\textit{e.g.}, GPU memory) constraints, especially in edge devices.

\noindent\textbf{Alternatives to \texttt{Batchnorm}.} To avoid using large-sized batches, viable options include updating \texttt{GroupNorm} \citep{mummadi2021test} or \texttt{LayerNorm}, especially in transformer-based tasks \citep{DBLP:conf/ijcai/KojimaMI22}. The former method uses entropy for updating, while the latter further requires the assistance of sharpness-aware minimization \citep{DBLP:conf/iclr/ForetKMN21} in \citep{DBLP:conf/iclr/Niu00WCZT23}, which seeks a flat minimum of optimization. Followed by the above situation in scenarios where computational resources are limited, \texttt{MECTA} \citep{DBLP:conf/iclr/HongLZS23} introduced an innovative approach by replacing the conventional \texttt{BatchNorm} layer with a customized \texttt{MECTA} norm. This strategic change effectively mitigated memory usage concerns during adaptation, reducing the memory overhead associated with large batch sizes, extensive channel dimensions, and numerous layers requiring updates. Taking a different tack, \texttt{EcoTTA} \citep{DBLP:journals/corr/abs-2303-01904} incorporated and exclusively updated meta networks, including \texttt{BatchNorm} layers. This approach also effectively curtailed computational expenses, keeping source data discriminability while upholding robust test-time performance. Furthermore, to address the performance challenges associated with smaller batch sizes, \texttt{TIPI} \citep{nguyen2023tipi} introduced additional \texttt{BatchNorm} layers in conjunction with the existing ones. This configuration inherently maintains two distinct sets of data statistics and leverages shared affine parameters to enhance consistency across different views of test data.

\subsubsection{Mean Teacher Optimization}\label{KD}
The mean teacher model, as discussed in \citep{DBLP:conf/iclr/TarvainenV17}, presents a viable strategy to enhance optimization stability in OTTA. This approach involves initializing both the teacher and student models with a pre-trained source model. For any given test sample, weak and strong augmented versions are created. Each version is then processed by the student and teacher model correspondingly. The crux of this approach lies in employing prediction consistency, also known as consistency regularization, to update the student model. This strategy aims to achieve identical predictions from different data views, thereby reducing model sensitivity to the changes in the test data and improving prediction stability. Simultaneously, the teacher model is refined as a moving average of the student across iterations. Notably, in OTTA, the mean teacher model and \texttt{BatchNorm}-based methods are not mutually exclusive; in fact, they can be effectively integrated. Incorporating \texttt{BatchNorm} updates into the teacher-student learning framework can yield even more robust results (Sec. \ref{sec:Exp}). Similarly, the integration of the mean-teacher model with data-driven (as discussed in Sec. \ref{sec:data-based}) or model-driven (as detailed in Sec. \ref{sec:model-based}) methods shows promise for further enhancing the prediction accuracy and stability of OTTA, marking an important step forward in the field.

\noindent \textbf{Model updating strategies.} 
Following the idea of mean-teacher learning, \texttt{ViDA} \citep{DBLP:journals/corr/abs-2306-04344} utilizes to supervise the student output by the prediction from the teacher with augmented input. It further introduces high/low-rank adapters to be updated to suit continual OTTA learning. Details see Sec. \ref{sec:model-based}.
\citet{DBLP:conf/cvpr/0013FGD22} generally follows the standard consistency learning strategy but introduces a reset method: a fixed number of weights are reset to their source pre-trained states after each training iteration. This reset measure preserves source knowledge and brings robustness against misinformed updates. 

% Similarly, while  \citet{DBLP:conf/cvpr/BrahmaR23} apply parameter restoration, it further considers parameter importance by Fisher Information Matrix (\texttt{FIM}) \citep{DBLP:journals/corr/KirkpatrickPRVD16} for a given data, and only restores the parameter if the \texttt{FIM} score is less than a threshold. 

\texttt{RoTTA} \citep{DBLP:journals/corr/abs-2303-13899} adopts a different approach, focusing on updating only the customized batch normalization layer, termed \texttt{RBN}, in the student model, rather than altering all parameters.  This strategy not only benefits from consistency regularization but also integrates statistics of the out-of-distribution test data.

\noindent \textbf{Divergence in Augmentations.} Drawing inspiration from the prediction consistency strategy in the mean teacher model, \citet{DBLP:journals/corr/abs-2303-09870} propose learning adversarial augmentation to identify the most challenging augmentation policies, which drive image feature representations of towards uncertain regions near decision boundaries. 
This method not only achieves clearer decision boundaries but also enhances the separation of class-specific features, significantly improving model insensitivity to styles of unseen test data.

\subsubsection{Optimization Objective} \label{sec:optim}
\begin{figure}[t]
\centering
\includegraphics[width=0.95\linewidth]{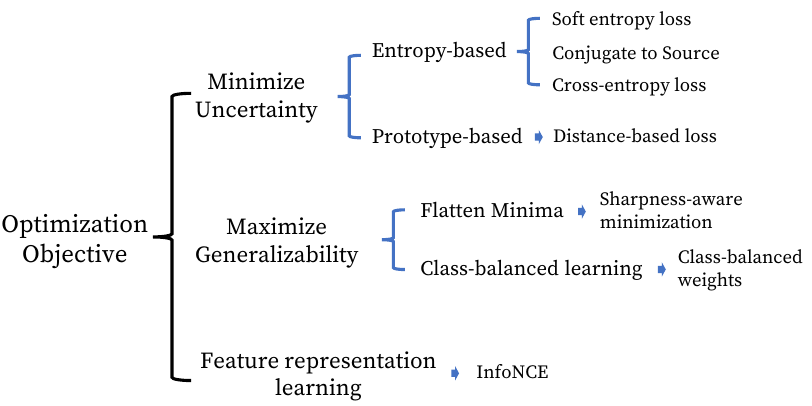}
\caption{Common optimization objectives in OTTA.}\label{fig:optim}
\end{figure}

Designing a proper optimization objective is important under the challenges of shifted test data with a limited amount. Commonly seen optimization-based Online Test-time Adaptation (OTTA) are summarized in Fig. \ref{fig:optim}. Existing literature addresses the optimization problem using three primary strategies below.

\noindent\textbf{Optimizing (increasing) confidence.} Covariate shifts typically lead to lower model accuracy, which in turn causes the model to express high uncertainty. The latter is often observed. 
As such, to improve model performance, an intuitive way is to enhance model confidence for the test data. 

\noindent \underline{Entropy-based confidence optimization.} This strategy typically aims to minimize the entropy of the softmax output vector:
\begin{equation}
\quad H(\hat{y})=-\sum_c p\left(\hat{y}_c\right) \log p\left(\hat{y}_c\right),\label{eq:ent}
\end{equation}
where $\hat{y}_c$ is the $c$-th predicted class, $p\left(\hat{y}_c\right)$ is its corresponding prediction probability. Intuitively, when the entropy of the prediction decreases, the vector will look sharper, where the confidence, or maximum confidence, increases. In OTTA, this optimization method will increase model confidence for the current batch without replying to labels and improve model accuracy. 

Two lines of work exist. One considers the entire softmax vector; the other leverages auxiliary information and uses the maximum entry of the softmax output. \texttt{Tent} is a typical method for the former.

\texttt{Tent} is a popular method that uses entropy minimization to update the affine parameters in \texttt{BatchNorm}. 
Subsequent studies have adopted and expanded upon this strategy. For example, \citet{DBLP:journals/corr/abs-2309-03964} introduces entropy minimization by incorporating self-paced learning. This addition ensures the learning process progresses at an adaptive and optimal pace. Furthermore, by integrating the general adaptive robust loss \citep{DBLP:conf/cvpr/Barron19} into self-paced learning, the proposed method achieves robustness against large and unstable loss values. \texttt{TTPR} \citep{DBLP:journals/corr/abs-2110-10232} combines entropy minimization with prediction reliability, which operates across various views of a test image to form a consistency loss. This is done by merging the mean prediction across three augmented versions, with each augmented prediction separately. \citet{DBLP:journals/corr/abs-2310-20327} introduces minimizing the entropy loss on augmentation-averaged predictions while assigning high weights for low-entropy samples. In \texttt{SAR} \citep{DBLP:conf/iclr/Niu00WCZT23}, when minimizing entropy, an optimization strategy is used so that parameters enabling `flatter' minimum regions can be found. This is shown to allow for a better model updating stability. %as detailed in Eq. \eqref{eq:SAR}.

While entropy minimization is widely used, a natural question arises: what makes soft entropy a preferred choice? To reveal the working mechanism of the loss function, \texttt{Conj-PL} \citep{DBLP:conf/nips/GoyalSRK22} designs a meta-network to parameterize it and observes that the meta-output could echo the temperature-scaled softmax output of a given model. They prove that if the cross-entropy loss is applied during source pre-training, the soft entropy loss is the most proper loss during adaptation.

A drawback of entropy minimization is that it might obtain a degenerate solution where every data point is assigned to the same class. To avoid this, \texttt{MuSLA} \citep{DBLP:conf/icip/Kingetsu0OYN22} employs mutual information of the sample $X$ and the corresponding prediction $\hat{Y}$,
\begin{equation}
\quad I_t(X ; \hat{Y})=H\left(\hat{\boldsymbol{p}}_0\right)-\frac{1}{\mathbf{M}} \sum_{t^i} H\left(\hat{\boldsymbol{p}_t^i}\right),
\end{equation}
where $\hat{\boldsymbol{p}}_0$ is the prior distribution, $M$ is mini-batch size, $i$ is the index of the sample within the batch. Maximizing $\frac{1}{\mathbf{M}} \sum_{t^i} H\left(\hat{\boldsymbol{p}_t^i}\right)$ could be seen as a regularizer to avoid same-class prediction. 

In entropy minimization, the gradient can often be dominated by low-confidence predictions. However, the cross-entropy loss can be too strict when it comes to predicted labels, leading to incorrect updates of the model even when there is a single wrong prediction. To address this trade-off, a new approach, Soft Likelihood Ratio (SLR) loss, is proposed by \citet{mummadi2021test} and further employed by \citet{DBLP:conf/wacv/MarsdenD024}. This approach emphasizes predicted classes while considering the issue raised in \texttt{MuSLA}:

\begin{equation}\label{eq:roid}
\quad \mathcal{L}_{SLR}(\hat{y}_{ti}) = - \sum_{c} w_{ti} \hat{y}_{tic} \log\left( \frac{\hat{y}_{tic}}{\sum_{j \neq c} \hat{y}_{tij}} \right),
\end{equation}

where $\hat{y}_{ti}$ is the softmax probability for the i-\textit{th} test sample at time step t. $w_{ti}$ is a \texttt{ROID}-only weight. It is a combination of diversity, considering the similarity between the recent trend of the model's prediction and the current model output, and certainty, which is the negative entropy of the output. Here, suppose the output confidence for class $c$ is small. In that case, the loss calculation will be reweighted by the summation over all other predictions $\sum_{j \neq c} \hat{y}_{tij}$ in the denominator, thus relaxing the focus on low-confidence classes.

% \underline{Mean-teacher. }
The cooperation between a teacher and a student is another possible solution to optimize prediction confidence with reliability. Here, the teacher is usually the moving averaged model of interest across iterations. \texttt{CoTTA} \citep{DBLP:conf/cvpr/0013FGD22} uses the Softmax prediction from the teacher to supervise the Softmax predictions from the student under the cross-entropy loss.

To give more precise supervision to the student, 
% the teacher model in \texttt{PETAL} \citep{DBLP:conf/cvpr/BrahmaR23} computes soft pseudo labels from the average output of different augmented views of each sample. 
\texttt{RoTTA} \citep{DBLP:conf/cvpr/YuanX023} adds a twist: the model is updated using samples stored in a memory bank. Moreover, a reweighting mechanism is introduced to prevent the model from overfitting to `old' samples in the memory bank. This reweighting prioritizes updates using `new' samples in the memory bank, ensuring a more dynamic and current learning process. Please see Sec. \ref{sec:memBank} for more details about its memory bank strategy.

Supervised by student output, \texttt{TeSLA} design its objective as a cross-entropy loss with a regularizer: 
\begin{equation}
\begin{aligned}
\quad& \mathcal{L}_{\mathrm{pl}}(\bm X, \hat{\bm Y})=-\frac{1}{B} \sum_{i=1}^B \sum_{k=1}^K f_s\left(\boldsymbol{x}_i\right)_k \log \left(\left(\hat{\boldsymbol{y}}_i\right)_k\right) \\
\quad&+\sum_{k=1}^K \hat{f}_s(\bm X)_k \log \left(\hat{f}_s(\bm X)\right)_k,
\end{aligned}
\end{equation}
where $\hat{f}_s(X)=\frac{1}{B} \sum_{i=1}^B f_s\left(\boldsymbol{x}_i\right)$ is the marginal class distribution of the student over the batch. $\hat{y}$ is the soft pseudo-label from the teacher model. $k$ is the number of classes. Except for the cross-entropy loss similar to the previous methods, it also maximizes the entropy for the averaged prediction of the student model across the batch to avoid overfitting.

A cross-entropy loss helps minimize model prediction uncertainty but may fail to give consistent uncertainty scores under different augmentations, \textit{etc}. This problem is more critical when the teacher-student mechanism is not used. To address this problem, \texttt{MEMO} \citep{DBLP:conf/nips/ZhangLF22} computes the average prediction across multiple augmentations for each test sample and then minimizes the entropy of the marginal output distribution over augmentations: 
\begin{equation}\label{eq:bald}
 \ell(\theta ; \mathbf{x}) \triangleq H\left(\bar{p}_\theta(\cdot \mid \mathbf{x})\right)=-\sum_{y \in \mathcal{Y}} \bar{p}_\theta(y \mid \mathbf{x}) \log \bar{p}_\theta(y \mid \mathbf{x}),
\end{equation}
where $\bar{p}$ denotes the averaged Softmax vector. 
Here, augmentations are randomly generated by \texttt{AugMix} \citep{DBLP:conf/iclr/HendrycksMCZGL20}. 

To encourage consistency against smaller perturbations, \citet{DBLP:conf/wacv/MarsdenD024} proposes a consistency loss based on symmetric cross-entropy loss (\texttt{SCE}) \citep{DBLP:conf/iccv/0001MCLY019}:

\begin{equation}
\begin{aligned}
\quad & \mathcal{L}_{SCE}(\hat{y}_{ti}; y_{ti}) = \\
\quad & - \frac{w_{ti}'}{2} \left( \sum_{c=1}^{C} \hat{y}_{tic} \log \tilde{y}_{tic} + \sum_{c=1}^{C} \tilde{y}_{tic} \log \hat{y}_{tic} \right).
\end{aligned}
\end{equation}
This is enabled by promoting similar outputs between test images which
have been identified as certain and diverse and an augmented view of them. Here, $\hat{y}_{ti}$ is the softmax probability for the i-\textit{th} test sample at time step t. $\tilde{y}_{t}$ is the softmax probability of the augmented view. $w_{ti}'
$ is the weight of the augmented view as in Eq. \eqref{eq:roid}. 

\noindent\underline{Prototype-based optimization}. 
Prototype-based learning \citep{DBLP:conf/cvpr/YangZYL18} is a commonly used strategy for unlabeled data by selecting representative or average for each class and classifying unlabeled data by distance-based metrics. However, its effectiveness might be limited under distribution shifts. 
Seeking a reliable prototype, \texttt{TSD} \citep{DBLP:journals/corr/abs-2303-10902} uses a Shannon entropy-based filter to find class prototypes from target samples that have high confidence. Then, a target sample is used to update the classifier-of-interest if its nearest prototypes are consistent with its class predictions from the same classifier.

\noindent\textbf{Improving generalization ability to unseen target samples.} Typically, OTTA uses the same batch of data for model update and evaluation. Here, we would like the model to perform well on upcoming test samples or have good generalization ability. A useful technique is 
sharpness-aware minimization (SAM) \citep{DBLP:conf/iclr/ForetKMN21}, where instead of seeking a minima that is `sharp' in its gradients nearby, a `flat' minima region is preferred. \citet{DBLP:conf/iclr/Niu00WCZT23} uses the following formulation to demonstrate the effectiveness of this strategy.\vspace{-2ex}
\begin{equation} \label{eq:SAR}
\quad \min _{\tilde{\Theta}} S(\mathbf{x}) E^{S A}(\mathbf{x} ; \Theta).
\end{equation}
Here, $S(\mathbf{x})$ is an entropy-based indicator function that can filter out unreliable predictions based on a predefined threshold. $E^{S A}(\mathbf{x} ; \Theta)$ is defined as: 
\begin{equation}
\quad E^{S A}(\mathbf{x} ; \Theta) \triangleq \max _{\|\epsilon\|_2 \leq \rho} E(\mathbf{x} ; \Theta+\boldsymbol{\epsilon}).
\label{eq:sharpness}
\end{equation}
This term aims to identify a weight perturbation $\epsilon$ within a Euclidean ball of radius $\rho$ that maximizes entropy. It quantifies sharpness by measuring the maximal change in entropy between $\Theta$ and $\Theta + \rho$. As such, Eq. \eqref{eq:sharpness}   %The ultimate objective involves 
jointly minimizing entropy and its sharpness. \citet{DBLP:journals/corr/abs-2310-10074} uses same idea for their optimization. 

Another difficulty affecting OTTA generalization is the class imbalance in a batch: a limited number of data for the model update would often not reflect the true class frequency. To solve this problem, Dynamic online re-weighting (\texttt{DOT}) in \texttt{DELTA} \citep{DBLP:conf/iclr/Zhao0X23} uses a momentum-updated class-frequency vector, which is initialized with equal weights for each class and then updated at every inference step based on the pseudo-label of the current sample and model weight. For a target sample, a significant weight (or frequency) for a particular class prompts \texttt{DOT} to diminish its contribution during subsequent adaptation learning, which avoids biased optimization towards frequent classes, thus improving model generalization ability.

\noindent\textbf{Feature representation learning}. Since no annotations are assumed for the test data, contrastive learning \citep{DBLP:journals/corr/abs-1807-03748} could naturally be used in test-time adaptation tasks. In a self-supervised fashion, contrastive learning is to learn a feature representation where positive pairs (data sample and its augmentations) are close and negative pairs (different data samples) are pushed away from each other. However, this requires multiple-epoch updates, which violate the online adaptation setting. To suit online learning, \texttt{AdaContrast} \citep{DBLP:conf/cvpr/0001WDE22} uses target pseudo-labels to disregard the potential same-class negative samples rather than treat all other data samples as negative. 

\subsubsection{Pseudo-labeling} \label{sec:pseudo-label}
Pseudo-labeling is a useful technique in domain adaptation and semi-supervised learning. It typically assigns labels to samples with high confidence, and these pseudo-labeled samples are then used for training. 

In OTTA, where adaptation is confined to the current batch of test data, \textbf{batch-level} pseudo-labeling is often used. % as a viable strategy. 
For example, \texttt{MuSLA} \citep{DBLP:conf/icip/Kingetsu0OYN22} implements pseudo-labeling as a post-optimization step following \texttt{BatchNorm} updates. This approach refines the classifier using pseudo-labels of the current batch, thereby enhancing model accuracy.

Furthermore, the teacher-student framework, as seen in models like \texttt{CoTTA} \citep{DBLP:conf/cvpr/0013FGD22}, \texttt{RoTTA} \citep{DBLP:conf/cvpr/YuanX023}, and \texttt{ViDA} \citep{DBLP:journals/corr/abs-2306-04344}, also adopts the pseudo-labeling strategy where the teacher outputs are used as soft pseudo-labels. With the uncertainty maintained during backpropagation, this could prevent the model from being overfitted to the incorrect predictions.

\noindent\textbf{Reliable pseudo labels} are an essential requirement. However, it is particularly challenging in the context of OTTA. On the one hand, due to the use of continuous data streams, we have limited opportunity for review. 
On the other hand, the covariate shift between the source and test sets could significantly degrade the reliability of pseudo-labels. 

To address these challenges, \texttt{TAST} \citep{DBLP:conf/iclr/JangCC23} adopts a prototype-based pseudo-labeling strategy. They first obtain the prototypes as the class centroids in the support set, where the support set is initially derived from the weights of the source pre-trained classifier and then refined and updated using the normalized features of the test data. To avoid performance degradation brought by unreliable pseudo labels, it calculates centroids only using the nearby support examples and then uses the temperature-scaled output to obtain the pseudo labels. Alternatively, \texttt{AdaContrast} \citep{DBLP:conf/cvpr/0001WDE22} uses soft $K$ nearest neighbors voting \citep{mitchell2001soft} in the feature space to generate reliable pseudo labels for target samples. On the other hand, \citet{wu2021domain} proposes to use multiple augmentations and majority voting to achieve consistent and trustworthy pseudo-labels.

\noindent\textbf{Complementary pseudo-labeling (PL).} One-hot pseudo-labels often result in substantial information loss, especially under domain shifts. To address this, \texttt{ECL} \citep{DBLP:journals/corr/abs-2301-06013} considers both maximum-probability predictions and predictions that fall below a certain confidence threshold (\textit{i.e.}, complementary labels). 
This is because of the intuition that if the model is less confident about a prediction, this prediction should be penalized more heavily. This method helps prevent the model from making aggressive updates based on incorrect but high-confidence predictions, offering a more stable approach similar to soft pseudo-label updates. 

\subsubsection{Other Approaches}
\vspace{-1ex}
Deviating from the conventional path of adapting source pre-trained models, Laplacian Adjusted Maximum likelihood Estimation (\texttt{LAME}) \citep{DBLP:conf/cvpr/BoudiafMAB22} 
focuses on refining the model output. This is achieved by discouraging the refined output from deviating from the pre-trained model while encouraging label smoothness according to the manifold smoothness assumption. The final refined prediction is obtained when the energy gap for each refinement step of a batch is small. 

To prevent loss of generalization and catastrophic forgetting, weight ensembling integrates a solution. \texttt{ROID} \citep{DBLP:conf/wacv/MarsdenD024} continuously ensemble the weights of the initial source model and the weights of the current model at time step $t$ using moving average, allowing for partial retention of the source knowledge. This approach is akin to the parameter reset strategy in \texttt{CoTTA}, commonly employed in continual adaptation tasks. Additionally, to address temporal correlation and class-imbalanced issues during adaptation, it introduces the idea of prior correction. The intuition is that if the class distribution within a batch tends to be uniform, strong smoothing is applied to ensure that no class is favored. This is indicated by the sample mean over the current softmax prediction $\hat{p}_t$. Thus, the smoothing scheme is defined as:
 \begin{equation}
\quad \bar{p}_{t} = \frac{\hat{p}_{t} + \gamma}{1 + \gamma N_{c}},
\end{equation}
where $N_c$ denotes the number of classes and $\gamma$ is an adaptive smoothing factor.
 
\subsubsection{Summary}

Optimization-based methods stand out as the most commonly seen category in online test-time adaptation, independent of the neural architecture. These methods concentrate on ensuring consistency, stability, and robustness in optimization. However, an underlying assumption of these methods is the availability of sufficient target data, which should reflect the global test data distribution. 
Addressing this aspect, the next section will focus on data-based methods, examining how they tackle the lack of accessible target data in OTTA.

%\vspace{-1ex}
\subsection{Data-based OTTA}\label{sec:data-based}
\vspace{-1ex}
With a limited number of samples in the test batch, it is common to encounter test samples with unexpected distribution changes. We acknowledge \textbf{data} might be the key to bridging this gap between the source and the test data.
In this section, we delve deeper into strategies centered around data in OTTA. We highlight various aspects of data, such as diversifying the data in each batch (Sec. \ref{sec:aug}) and preserving high-quality information on a global scale (Sec. \ref{sec:memBank}). These strategies could enhance model generalizability and tailor model discriminative capacity to the current data batch.

\begin{figure*}
\centering
\includegraphics[width=0.96\linewidth]{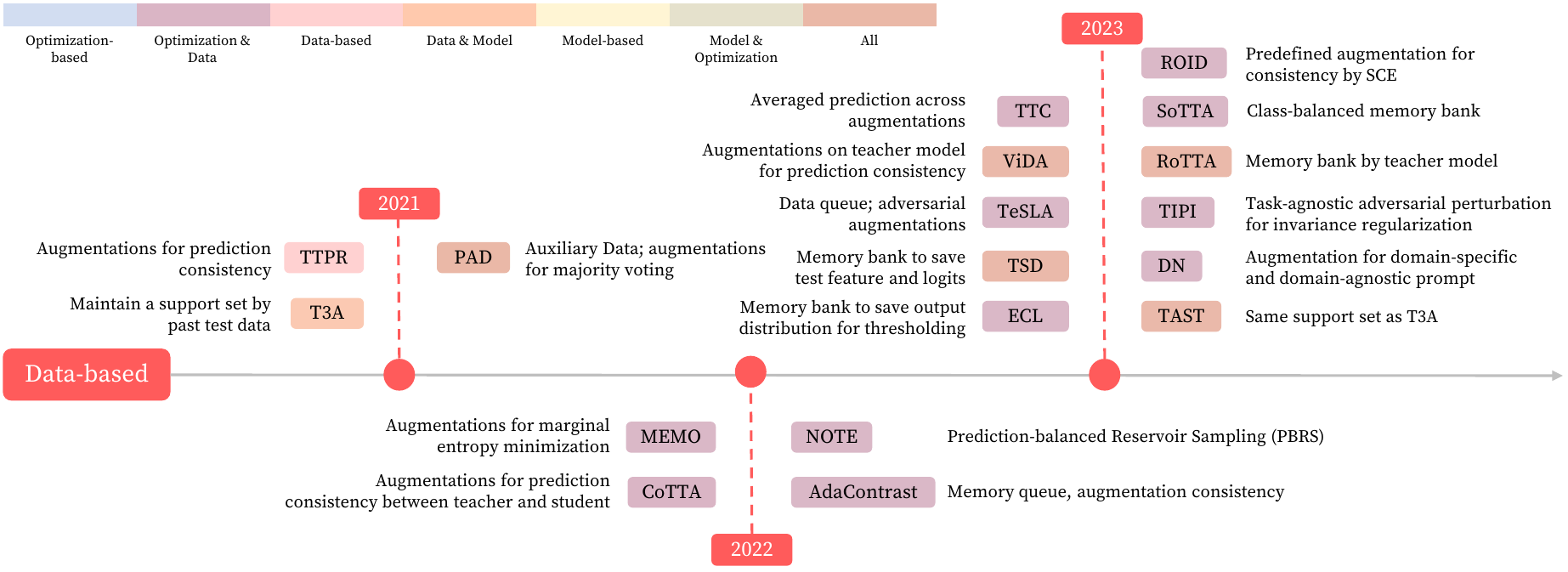}
\caption{Timeline of Data-based OTTA methods. }\label{fig:data-based}
% \vspace{-0.5cm}
\end{figure*}
\subsubsection{Data Augmentation} \label{sec:aug}
% \vspace{-1ex}
Data augmentation is important in domain adaptation \citep{DBLP:journals/ijon/WangD18} and domain generalization \citep{DBLP:journals/pami/ZhouLQXL23}, which mimics real-world variations to improve model transferability and generalizability. It is particularly useful for test-time adaptation.

\noindent\textbf{Predefined Augmentations.} 
Common data augmentation methods like cropping, blurring, and flipping are effectively incorporated into various OTTA methodologies. %showcasing their ability to mimic certain data variations. 
An example of this integration is \texttt{TTC} \citep{DBLP:journals/corr/abs-2310-20327}, which updates the model using averaged predictions from multiple augmentations. Another scenario is from the mean teacher model such as \texttt{RoTTA} \citep{DBLP:conf/cvpr/YuanX023}, \texttt{CoTTA} \citep{DBLP:conf/cvpr/0013FGD22}, and \texttt{ViDA} \citep{DBLP:journals/corr/abs-2306-04344}, applies predefined augmentations to teacher/student input, and maintains prediction consistency across different augmented views.

To ensure consistent and reliable predictions, \texttt{ROID} \citep{DBLP:conf/wacv/MarsdenD024} applies augmentation for prediction consistency by \texttt{SCE}. \texttt{PAD} \citep{wu2021domain} employs multiple augmentations of a single test sample for majority voting. This is grounded in the belief that if the majority of the augmented views yield the same prediction, it is likely to be correct, as it demonstrates insensitivity to variations in style. Instead, \texttt{TTPR} \citep{DBLP:journals/corr/abs-2110-10232} adopts KL divergence to achieve consistent predictions. For every test sample, it generates three augmented versions. The model is then refined by aligning the average prediction across these augmented views with the prediction for each view. Another approach is \texttt{MEMO}, which uses \texttt{AugMix} \citep{DBLP:conf/iclr/HendrycksMCZGL20} for test images. For a test data point, a range (usually $32$ or $64$) of augmentations from the AugMix pool $\mathcal{A}$ is generated to make consistent predictions. 

\noindent\textbf{Contextual Augmentations.}
Previously, OTTA methods often predetermine augmentation policies. Given that test distributions can undergo substantial variations in continuously evolving environments, there exists a risk that such fixed augmentation policies may not be suitable for every test sample. In \texttt{CoTTA} \citep{DBLP:conf/cvpr/0013FGD22}, rather than augmenting every test sample by a uniformed strategy, augmentations are judiciously applied only when domain differences (i.e., low prediction confidence) are detected, mitigating the risk of misleading the model. 
% \texttt{PETAL}
% \citep{DBLP:conf/cvpr/BrahmaR23} follows the idea of \texttt{CoTTA} and applies augmentations to test samples for teacher input to produce augmentation-averaged prediction only if the domain difference is substantial.

\noindent\textbf{Adversarial Augmentation. }Traditional augmentation methods always provide limited data views without fully representing the domain differences. \texttt{TeSLA} \citep{DBLP:journals/corr/abs-2303-09870} moves away from this. It instead leverages adversarial data augmentation to identify the most effective augmentation strategy. Instead of a fixed augmentation set, it creates a policy search space $\mathcal{O}$ as the augmentation pool, then assigns a magnitude parameter $m \in[0,1]$ for each augmentation. A sub-policy $\rho$ consists of augmentations and their corresponding magnitudes. To optimize the policy, the teacher model is adapted using an entropy maximization loss with a severity regularization to encourage prediction variations while avoiding the augmentation too strong to be far from the original image.

\subsubsection{Memory Bank} \label{sec:memBank}
Going beyond augmentation strategies that could diversify the data batch, the memory bank is a powerful tool to preserve valuable data information for future memory replay. Setting up a memory bank involves two key considerations: (1). Determining \textbf{which data} should be stored in the memory bank. This requires identifying samples that are valuable for possible replay during adaptation.
 (2). \textbf{The management} of the memory bank. This includes strategies for adding new instances and removing old ones from the bank. 

Memory bank strategies generally fall into time-uniform and class-balanced categories. Notably, many methods choose to integrate both types to maximize effectiveness. In addressing the challenges posed by both temporally correlated distributions and class-imbalanced issues, \texttt{NOTE} \citep{DBLP:conf/nips/GongJKKSL22} introduces the Prediction-Balanced Reservoir Sampling (\texttt{PBRS}) to save sample-prediction pairs. The ingenuity of \texttt{PBRS} lies in its fusion of two distinct sampling strategies: time-uniform and prediction-uniform. The time-uniform approach, reservoir sampling (\texttt{RS}), aims to obtain uniform data over a temporal stream. Specifically, for a sample $x$ being predicted as class $k$, we randomly sample a value $p$ from a uniform distribution $[0,1]$. Then, if $p$ is smaller than the proportion of class $k$ in whole memory bank samples, pick one randomly from the same class and replace it with the new one $x$. Instead, the prediction-uniform saving strategy (\texttt{PB}) prioritizes the predicted labels to ascertain the majority class within the memory. Upon identification, it supplants a randomly selected instance from the majority class with a fresh data sample, thereby ensuring a balanced representation. The design of \texttt{PBRS} ensures a more harmonized distribution of samples across both time and class dimensions, fortifying the model's adaptation capabilities.

A similar strategy is employed in \texttt{SoTTA} \citep{DBLP:journals/corr/abs-2310-10074} to facilitate class-balanced learning. Each high-confidence sample-prediction pair is stored when the memory bank has available space. If the bank is full, the method opts to replace a sample either from one of the majority classes or from its class if it belongs to the majority. This ensures a more equitable class distribution and strengthens the learning process against class imbalances. Another work, \texttt{RoTTA} \citep{DBLP:conf/cvpr/YuanX023}, offers a category-balanced sampling with timeliness and uncertainty (\texttt{CSTU}) module, dealing with the batch-level shifted label distribution. In \texttt{CSTU}, the author proposes a category-balanced memory bank $M$ with a capacity of $N$. In the memory bank, data samples $x$ are stored alongside their predicted labels $\hat{y}$, a heuristic score $\mathcal{H}$, and uncertainty metrics $\mathcal{U}$. Here, the heuristic score is calculated by:
\begin{equation}
\quad \mathcal{H}=\lambda_t \frac{1}{1+\exp (-\mathcal{A} / \mathcal{N})}+\lambda_u \frac{\mathcal{U}}{\log \mathcal{C}}
\end{equation}
where $\lambda_t$ and $\lambda_u$ is the trade-off between time and uncertainty, $\mathcal{A}$ is the age (i.e., how many iterations this sample has been stored in the memory bank) of a sample stored in the memory bank. $\mathcal{C}$ is the number of classes, $\mathcal{N}$ is the capacity of the memory bank, and $\mathcal{U}$ is the uncertainty measurement, which is implemented as the entropy of the sample prediction. The score $\mathcal{H}$ is then used to decide whether a test sample should be saved into the memory bank for each class. As the lower heuristic score is always preferred, its intuition is to maintain fresh (i.e., lower age $\mathcal{A}$), balanced, and certain (i.e., lower $\mathcal{U}$) test samples. thereby enhancing adaptability during online operations.

To avoid the negative impact of the batch-level class distribution, \texttt{TeSLA} \citep{DBLP:journals/corr/abs-2303-09870} incorporates an online queue to hold class-balanced, weakly augmented sample features and their corresponding pseudo-labels. To enhance the correctness of pseudo-label predictions, each test sample is compared with its closest matches within the queue. Similarly, \texttt{TSD} \citep{DBLP:journals/corr/abs-2303-10902} is dedicated to preserving sample embeddings and their associated logits in a memory bank for trustworthy predictions. Initialized by the weights from a source pre-trained linear classifier \citep{DBLP:conf/nips/IwasawaM21}, this memory bank is subsequently employed for prototype-based classification. 

Contrastive learning is well-suited for OTTA, as discussed in \ref{sec:optim}. However, this approach can be challenging for online learning, especially when it pushes away feature representations of data from the same class. Unlike conventional methods where the feature space can be revisited multiple times, \texttt{AdaContrast} \citep{DBLP:conf/cvpr/0001WDE22} offers an innovative solution. It keeps all previously encountered key features and pseudo-labels in a memory queue to avoid forming `push-away' pairs from the same class. This method speeds up the learning process and reduces the risk of error accumulation in data from the same class, thereby improving the efficiency and precision of the learning process.

\texttt{ECL} \citep{DBLP:journals/corr/abs-2301-06013} represents a novel shift away from traditional methods by incorporating a memory bank about output distributions for setting thresholds on complementary labels. The memory bank is also periodically refreshed using the newly updated model parameters, ensuring its relevance and effectiveness.

\subsubsection{Summary}
Data-based techniques are particularly useful for handling online test sets that may be biased or have unique stylistic constraints. However, these techniques often increase computational demands, posing challenges in online scenarios. The following section will focus on an alternative strategy: how architectural modifications can offer distinct advantages in Online Test-Time Adaptation.
\vspace{-2ex}
\subsection{Model-based OTTA}\label{sec:model-based}
Model-based OTTA concentrates on adjusting the model architecture to address distribution shifts. The changes made to the architecture generally involve either \textbf{adding new components} or \textbf{replacing existing blocks}. This category is expanded to include developments in prompt-based techniques. It involves adapting prompt parameters or using prompts to guide the adaptation process.

\subsubsection{Module Addition}
\begin{figure*}[t]
\centering
\includegraphics[width=0.96\linewidth]{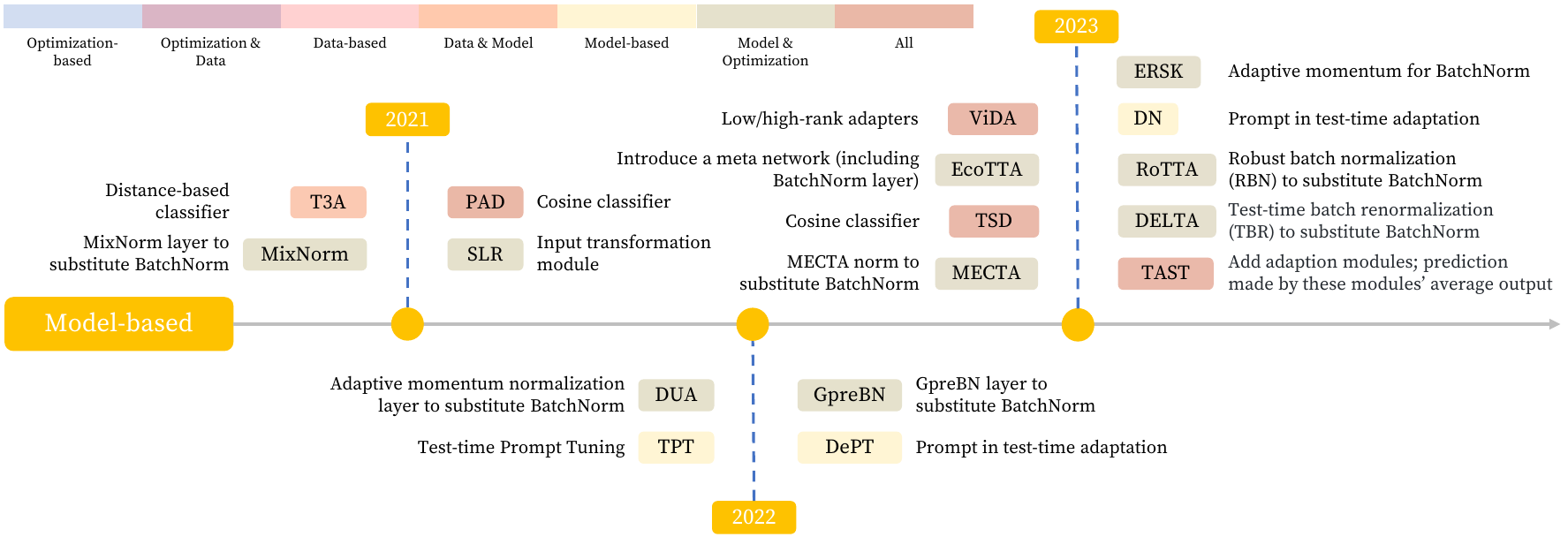}
\caption{Timeline of Model-based OTTA methods. }\label{fig:model-based}
% \vspace{-0.5cm}
\end{figure*}

\noindent\textbf{Input Transformation.} In an effort to counteract domain shift, \citet{mummadi2021test} introduce to optimize an input transformation module $d$, along with the \texttt{BatchNorm} layers as discussed in Sec. \ref{sec:optim}. This module is built on the top of the source model $f$, i.e., $g = f \circ d$. Specifically, $d(x)$ is defined as:
\begin{equation}
\quad d(x) = \gamma \cdot [\tau x + (1 - \tau) r_\psi(x)] + \beta,
\end{equation}
where $\gamma$ and $\beta$ are channel-wise affine parameters. The component $r_\psi$ denotes a network designed to have the same input and output shape, featuring $3 \times 3$ convolutions, group normalization, and ReLU activations. The parameter $\tau$ facilitates a convex combination of the unchanged and transformed input $r_\psi(x)$. 

\noindent\textbf{Adaptation Module.} To stabilize predictions during model updates, \texttt{TAST} \citep{DBLP:conf/iclr/JangCC23} integrates $20$ adaptation modules to the source pre-trained model. Based on \texttt{BatchEnsemble} \citep{DBLP:conf/iclr/WenTB20}, these modules are appended to the top of the pre-trained feature extractor. The adaptation modules are updated multiple times independently by merging their averaged results with the corresponding pseudo-labels for a batch of data.

In the case of continual adaptation, promptly detecting and adapting to changes in data distribution is inevitable to deal with catastrophic forgetting and the accumulation of errors. To realize this, \texttt{ViDA} \citep{DBLP:journals/corr/abs-2306-04344} utilizes the idea of low/high-rank feature cooperation. Low-rank features retain general knowledge, while high-rank features better capture distribution changes. To obtain these features, the authors introduce two adapter modules correspondingly parallel to the linear layers (if the backbone model is ViT). 

Additionally, since distribution changes in continual OTTA are unpredictable, strategically combining low/high-rank information is crucial. Here, the authors use MC dropout \citep{gal2016dropout} to assess model prediction uncertainty about input $x$. This uncertainty is then used to adjust the weight given to each feature. Intuitively, if the model is uncertain about a sample, the weight of domain-specific knowledge (high-rank feature) is increased, and conversely, the weight of domain-shared knowledge (low-rank feature) is increased. This helps the model dynamically recognize distribution changes while preserving its decision-making capabilities.

\subsubsection{Module Substitution}
Module substitution typically refers to swapping an existing module in a model with a new one. The commonly used techniques are about:

\noindent\textbf{Classifier.} Cosine-distance-based classifier \citep{DBLP:journals/jmlr/ChenGGRC09} offers great flexibility and interpretability by leveraging similarity to representative examples for decision-making. Employing this, \texttt{TAST} \citep{DBLP:conf/iclr/JangCC23} formulates predictions by assessing the cosine distance between the sample feature and the support set. 
\texttt{TSD} \citep{DBLP:journals/corr/abs-2303-10902} employs a similar classifier, assessing the features of the current sample against those of its K-nearest neighbors from a memory bank. \texttt{PAD} \citep{wu2021domain} uses a cosine classifier for predicting augmented test samples in its majority voting process. \texttt{T3A} \citep{DBLP:conf/nips/IwasawaM21} relies on the dot product between templates in the support set and input data representations for classification.

\noindent In the context of updating \textbf{\texttt{BatchNorm}} statistics, any alteration to \texttt{BatchNorm} that extends beyond the standard updating approach can be classified under this category. This includes techniques such as \texttt{MECTA} norm \citep{DBLP:conf/iclr/HongLZS23}, \texttt{MixNorm} \citep{DBLP:journals/corr/abs-2110-11478}, \texttt{RBN} \citep{DBLP:journals/corr/abs-2303-13899}, and \texttt{GpreBN} \citep{DBLP:journals/corr/abs-2205-10210}, etc. To maintain focus and avoid redundancy, these specific methods and their intricate details will not be extensively covered again in this section.

\subsubsection{Prompt-based Method}
The rise of vision-language models (VLMs) demonstrates their remarkable ability in zero-shot generalization. One typical example is \texttt{CLIP} \citep{DBLP:conf/icml/RadfordKHRGASAM21}, which calculates the cosine similarity between text and image features to enable zero-shot inference. To obtain the text feature, a set of predefined prompts is required to combine with the semantic class candidates.\footnote{To ensure a clear understanding of this new direction, we merge all the details of the three methods into this section. However, you can check their components in the timeline tables.}

Borrows the idea of prompting from VLMs. ``Decorate the Newcomers" (\texttt{DN}) \citep{DBLP:conf/aaai/GanBLMZSL23} employs prompts as supplementary information added onto the image input. To infuse the prompts with relevant information, it employs a student-teacher framework in conjunction with a frozen source pre-trained model to capture both domain-specific and domain-agnostic prompts. For acquiring domain-specific knowledge, it optimizes the cross-entropy loss between the outputs of the teacher and student models. Additionally, \texttt{DN} introduces a parameter insensitivity loss to mitigate the impact of parameters prone to domain shifts. This strategy aims to ensure that the updated parameters, which are less sensitive to domain variations, effectively retain domain-agnostic knowledge. Through this approach, \texttt{DN} balances learning new, domain-specific information while preserving crucial, domain-general knowledge.

\citet{DBLP:journals/corr/abs-2210-04831} introduces a novel method (\texttt{DePT}). Its process starts by segmenting the transformer into multiple stages, then introducing learnable prompts at the initial layer of each stage, concatenated with image and CLS tokens. During adaptation, \texttt{DePT} utilizes a mean-teacher model to update the learnable prompts and the classifier in the student model. For the student model, updates are made based on the cross-entropy loss calculated between pseudo labels and outputs from strongly augmented student output. Notably, these pseudo labels are generated from the student model, using the averaged predictions of the top-k nearest neighbors of the student's weakly augmented output within a memory bank. In terms of the teacher-student interaction, to counter potential errors from incorrect pseudo labels, \texttt{DePT} implements an entropy loss between predictions made by strongly augmented views of both the student and teacher models. Additionally, the method minimizes the mean squared error between the combined prompts of the student and teacher models at the output layer of the Transformer. Furthermore, to ensure that different prompts focus on diverse features and to prevent trivial solutions, \texttt{DePT} also maximizes the cosine distance among the combined prompts of the student.

 VLMs often underperform for domain-specific data. While attempting to address this, traditional fine-tuning strategies typically compromise the model's generalization power by altering its parameters. In response, with the idea of online adaptation, \textbf{Test Time Prompt Tuning} emerges as a solution. Deviating from the conventional methods, it fine-tunes the prompt, adjusting only the context of the model's input, thus preserving the generalization power of the model. One representative is \texttt{TPT} \citep{DBLP:conf/nips/ShuNHYGAX22}. It generates $N$ randomly augmented views of each test image and updates the prompting parameter by minimizing the entropy of the averaged prediction probability distribution. Additionally, a confidence selection strategy is proposed to filter out the output with high entropy to avoid noisy updating brought by unconfident samples. By updating the learnable parameter of the prompt, it could be easier to adapt the model to the new, unseen domains.

\begin{figure*}[!ht]
\centering
\subfloat[CIFAR-10-C $\&$ 10.1 Dataset]{{\includegraphics[width=0.32\textwidth]{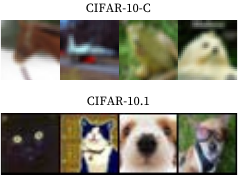} }}%
\subfloat[ImageNet-C Dataset]{{\includegraphics[width=0.32\textwidth]{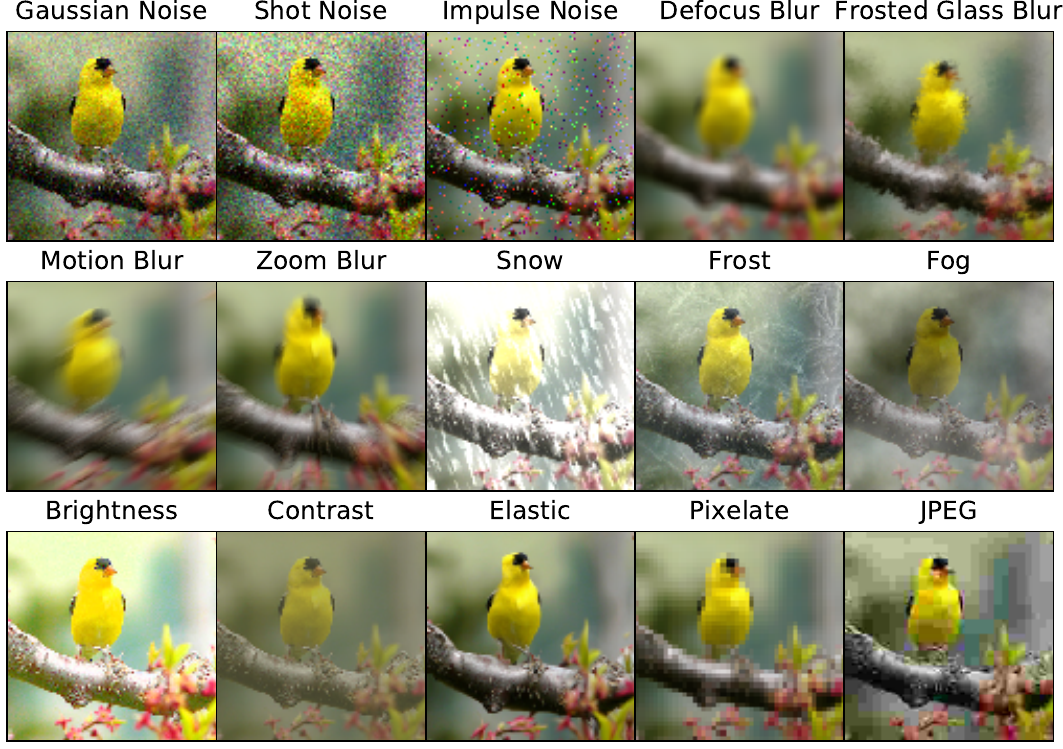} }}%
\subfloat[CIFAR-10-Warehouse Dataset]{{\includegraphics[width=0.32\textwidth]{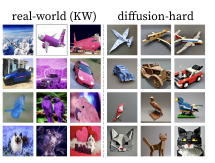} }}
\caption{The exemplars of the adopted datasets. The datasets include color variations, synthetic data, and types of corruption.}\label{fig:image-samples}
\end{figure*}

\subsubsection{Summary}
Model-based OTTA methods have shown effectiveness but are less prevalent than other groups, mainly due to their reliance on particular backbone architectures. For example, layer substitution mainly based on \texttt{BatchNorm} in the model makes them inapplicable to ViT-based architectures. 

A critical feature of this category is its effective integration with prompting strategies. This combination allows for fewer but more impactful model updates, leading to greater performance improvements. Such efficiency makes model-based OTTA methods especially suitable for complex scenarios.

\section{Empirical Studies} \label{sec:Exp}

% online test-time adaptation是一个实践性很强的task，而现实世界更注重
Existing OTTA methods predominantly use WideResNet \citep{DBLP:conf/bmvc/ZagoruykoK16} or ResNet \citep{DBLP:conf/cvpr/HeZRS16} for experiments, overlooking the evolution of backbones in recent years. In this study, we explore the possibility of decoupling OTTA methods from their conventional CNN backbones. We specifically focus on adapting these methods to the Vision Transformer model \citep{DBLP:conf/iclr/DosovitskiyB0WZ21}. % to assess their backbone-agnostic ability. 
Our work presents strategies for adapting methods, initially developed for CNNs, to work effectively with ViT architectures, thereby examining their flexibility under backbone changes.

\noindent\textbf{Baselines.} 
We evaluate eight OTTA methods, using a standardized testing protocol for fair comparison. We use six diverse datasets: three corrupted datasets (CIFAR-10-C, CIFAR-100-C, and ImageNet-C), two real-world shifted datasets (CIFAR-10.1 and OfficeHome), and one comprehensive dataset (CIFAR-10-Warehouse). CIFAR-10-Warehouse plays a pivotal role in our evaluation, featuring a broad array of subsets including real-world variations from different search engines with different colors and images created through the diffusion model. Specifically, we use the Google split in CIFAR-10-Warehouse to assess the capability of OTTA model to handle color shifts alongside mixed object styles. Furthermore, we evaluate the Diffusion split to assess the effectiveness of OTTA on artificially generated image samples, which have become increasingly popular in recent years.

\subsection{Implementation Details}
\noindent\textbf{Optimization details.} 
We use PyTorch for implementation on NVIDIA RTX A6000. The foundational backbone for all approaches is ViT-base-patch16-224 \citep{DBLP:conf/iclr/DosovitskiyB0WZ21}\footnote{https://github.com/huggingface/pytorch-image-models}.
When using CIFAR-10-C, CIFAR-10.1, and CIFAR-10-Warehouse as target domains, we train the source model on CIFAR-10 with $8,000$ iterations, including a warm-up phase spanning 1,600 iterations. The training uses a batch size of 64 and the stochastic gradient descent (SGD) algorithm with a learning rate of 3e-2. We use an identical configuration to train the source model on CIFAR-100, with an extended training duration of $16,000$ iterations and a warm-up period spanning $4,000$ iterations. The source model on the ImageNet-1k dataset is acquired from the Timm repository \footnote{\texttt{vit\_base\_patch16\_224.orig\_in21k\_ft\_in1k}}. For the OfficeHome dataset, we train the source model on the clipart domain for $3,500$ with $200$ iterations of warmup. Additionally, we apply basic data augmentation techniques, including random resizing and cropping, across all methods. We use the common optimization setup, employing the Adam optimizer with a momentum term $\beta$ of $0.9$ and a learning rate of 1e-3. Resizing and cropping techniques are applied as a default preprocessing step for all datasets. Then, an input normalization $(0.5, 0.5, 0.5)$ is adopted to mitigate potential performance fluctuations arising from external factors beyond the algorithm's core operations.
\begin{figure*}[t]
    \centering
    \includegraphics[width=0.95\linewidth]{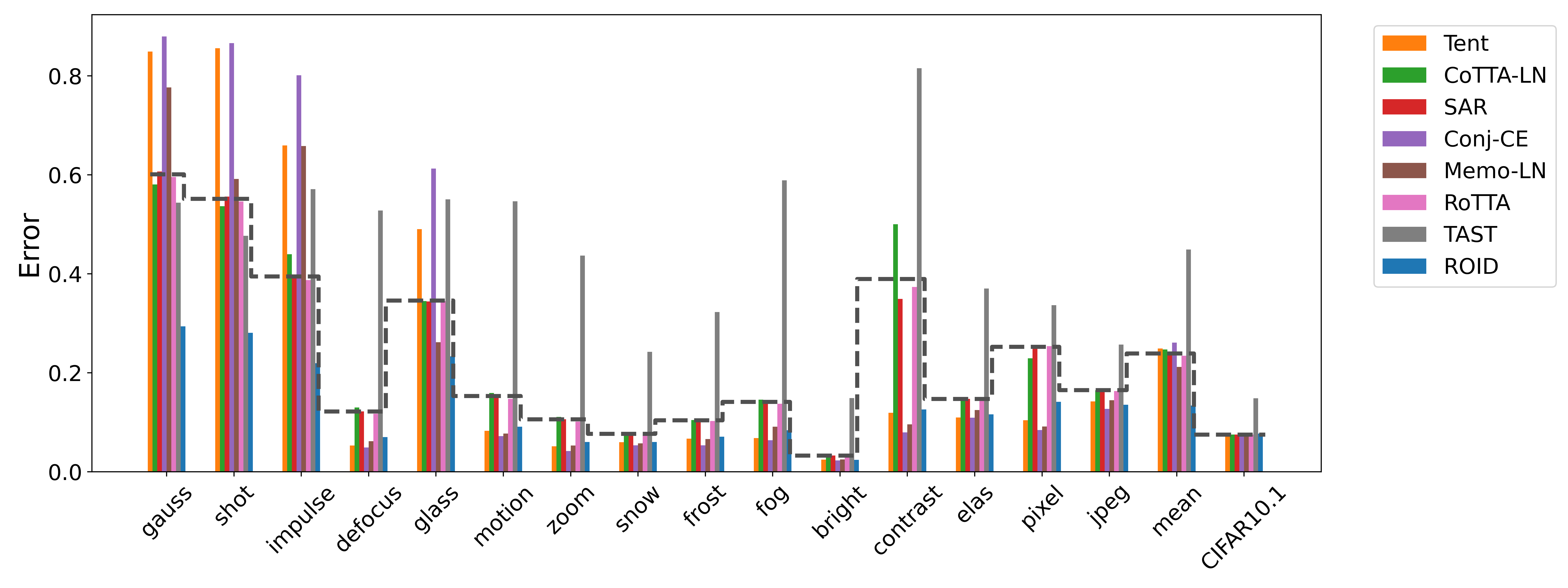}
    \includegraphics[width=0.95\linewidth]{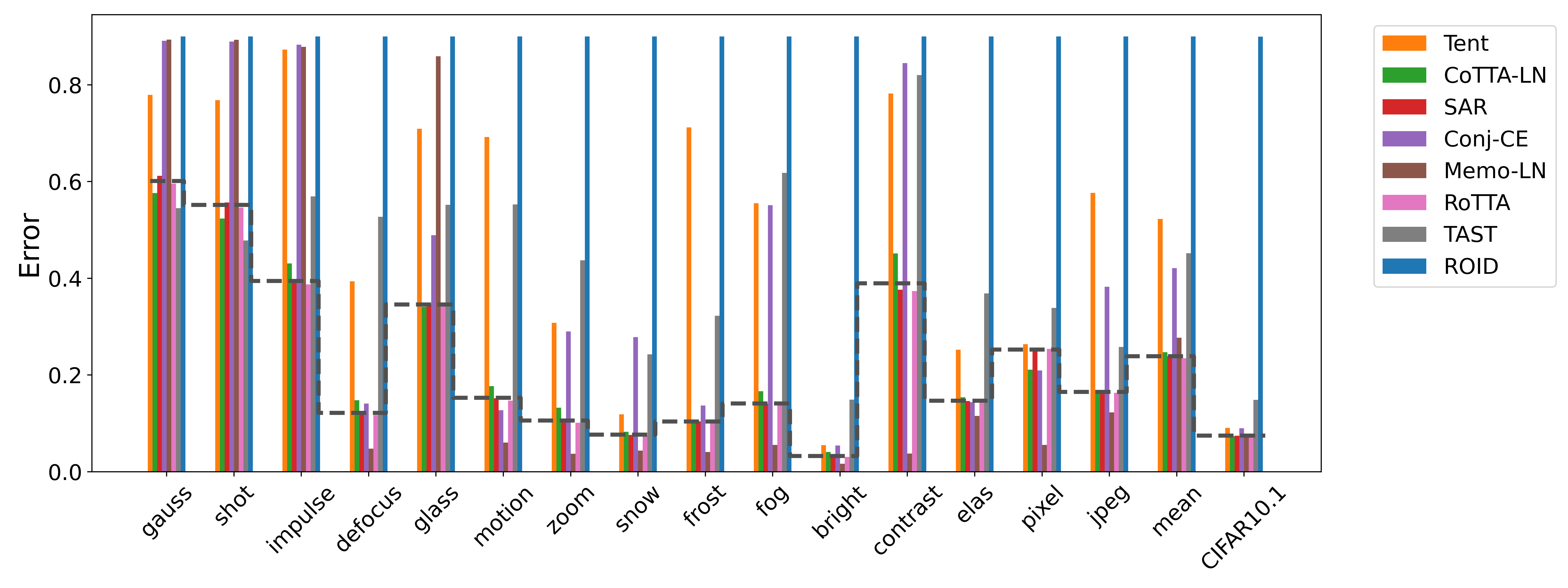}
    \caption{Comparison of the OTTA performance on the CIFAR-10-C (at severity level $5$) and CIFAR-10.1 datasets with ViT-base-patch16-224. The top and bottom plots show the experiments conducted with batch size $16$ and $1$ based on the \texttt{LayerNorm} updating strategy, respectively. The dotted line represents the source-only performance.}
    \label{fig:CIFAR-10}
\end{figure*}

\noindent\textbf{Component substitution.} To adapt core methods for use with the Vision Transformer (ViT), we develop a series of strategies:
\begin{itemize} 
    \item \textbf{Switch to \texttt{LayerNorm}}: In light of the absence of the \texttt{BatchNorm} layer in ViT, it is natural that all \texttt{BatchNorm}-based strategies in the original implementations would be automatically moved onto \texttt{LayerNorm}.
    \item \textbf{Disregard \texttt{BatchNorm} mixup}: Removing statistic mixup strategy originally designed for \texttt{BatchNorm}-based methods because \texttt{LayerNorm} is designed to normalize each data point independently.
    \item \textbf{Sample Embedding Changes}: For OTTA methods that rely on feature representations, an effective solution is to use the class embedding (e.g., the first dimension of the ViT model feature) as the image feature.
    \item \textbf{Pruning Incompatible Components}: Any elements incompatible with the ViT framework are identified and removed. 
\end{itemize}
Above are useful techniques for integrating important OTTA methods into ViT.  %expanding their application to  thereby broadening their application to this advanced model architecture. 
Note that these solutions are not only limited to the OTTA methods. Rather, they can be viewed as a broader set of guidelines that can be applied where there is a need for changing backbone architectures.

\noindent\textbf{Baselines}: We select eight methods that represent the eight OTTA categories mentioned in this paper, respectively. %, ensuring diversity and representation. 
They include:

\begin{enumerate}
    \item \texttt{Tent}: [optimization-based] A fundamental OTTA method rooted in \texttt{BatchNorm} updates for both statistics and affine parameters with entropy minimization. To reproduce it on ViTs, we replace its \texttt{BatchNorm} updates with a \texttt{LayerNorm} updating strategy.
    \item  \texttt{CoTTA} [optimization $\&$ data] uses the mean-teacher model, where the teacher model is updated by the moving average of the student. During adaptation, soft entropy is optimized and combined with selective augmentation. For each iteration, it also applies parameter reset (i.e., partially reset the model parameter to the source pre-trained version). While it requires updating the entire student network (i.e., \texttt{CoTTA-ALL}), we further assess \texttt{CoTTA-LN} that only update \texttt{LayerNorm} on its student model. We also deconstruct its parameter reset strategy, resulting in another two variants: updating \texttt{LayerNorm} without parameter reset (\texttt{CoTTA$^*$-LN}), and full network updating without parameter reset (\texttt{CoTTA$^*$-ALL}).
    \item  \texttt{SAR} [optimization-based] follows the same strategy as \texttt{Tent} while using sharpness-aware minimization (\texttt{SAM}) \cite{DBLP:conf/iclr/ForetKMN21} for flat minima during optimization. 
    \item  \texttt{Conjugate-PL}: [optimization-based] As the source model is optimized by the cross-entropy loss, this method is called \texttt{Conj-CE}. It is similar to \texttt{Tent} but allows the model to interact with the data twice for each iteration: once for updating \texttt{LayerNorm} (just like \texttt{Tent}) and another for prediction. 
    \item  \texttt{MEMO}: [optimization $\&$ data] For each test sample, \texttt{MEMO} applies various data augmentations and adjusts the model parameters to minimize the entropy across the model marginal output distributions from these augmentations. To ensure consistency and avoid unexpected performance fluctuations, we omit all data normalization processes from its set of augmentations. At the same time, we evaluate the performance of \texttt{MEMO} on two versions, \texttt{LayerNorm} update and full update, resulting in \texttt{MEMO-LN} and \texttt{MEMO-ALL} correspondingly.
    \item \texttt{RoTTA} [optimization, data $\&$ model] uses a memory bank to store class-balanced data, considering uncertainty and the ``age'' of each saved data sample. It also introduces a time-aware reweighting strategy to enhance adaptation stability. For evaluation, given that \texttt{BatchNorm} is not compatible with ViTs, we exclude the \texttt{RBN} module from its implementation.
    \item \texttt{TAST} [optimization, data $\&$ model] integrates multiple adaptation modules into the source pre-trained model. Based on \texttt{BatchEnsemble} \cite{DBLP:conf/iclr/WenTB20}, these modules are appended to the top of the pre-trained feature extractor. The adaptation modules are then updated multiple times independently by merging their averaged results with the corresponding pseudo-labels for a batch of data. To accommodate the ViT architecture (especially ViT-base), we use the class embedding from the first dimension as the feature representation.
    \item \texttt{ROID}\footnote{For fair comparison, we do not use gradient accumulation.} [optimization $\&$ data] is a method for not only typical OTTA but also universal TTA, capable of handling temporal correlation and domain non-stationarity. It incorporates weighted SLR loss with the Symmetric cross-entropy loss, alongside weight ensemble and prior correction mechanisms to ensure efficacy in complex scenarios. 
\end{enumerate}
 Despite the broad range of available OTTA methods, we believe that a detailed examination of this carefully selected subset will yield valuable insights. Our empirical study is designed to explore the following key questions.

\subsection{Does OTTA still work with ViT?}
To assess the transferability and adaptability of the selected OTTA methods, we compare them against the source-only  baseline (i.e., direct inference) on vision transformers. For consistency and controlled variable comparison, each bar chart included in our analysis is plotted based on the \texttt{LayerNorm} updating strategy.

\subsubsection{On CIFAR-10-C and CIFAR-10.1 Benchmarks} \label{sec:cifar10c}
We evaluate the CIFAR-10-C and CIFAR-10.1 datasets with batch sizes $1$ and $16$ and show results in Fig. \ref{fig:CIFAR-10}. We discuss our observations in three aspects.

\textbf{Corruption Types.} Across both batch sizes, most methods exhibit higher error rates in response to noise-induced corruptions, as illustrated in Fig. \ref{fig:CIFAR-10} and Fig. \ref{fig:bs_cifar10c}. In contrast, these methods tend to perform better when dealing with structured corruptions, such as snow, zoom, or brightness. This pattern suggests that various corruption types reflect different degrees of divergence from the corruption domain to the source dataset. Especially, adapting to noise corruption poses a significant challenge for confidence optimization-based methods (such as \texttt{Tent}, \texttt{MEMO-LN}, and \texttt{Conj-CE}), regardless of the batch size. This difficulty can be linked to the substantial domain gap and the unpredictable nature of noise patterns discussed earlier. Although these strategies aim to increase the model's confidence, they are not equipped to directly correct erroneous predictions. It is worth noting that \texttt{ROID} surpasses all other baselines on noise-based corruptions. This may attributed to the certainty- and diversity-weighted SCE loss, which is designed to address noise issues while avoiding trivial solutions. CIFAR 10.1 stands out as an exception, as it is not associated with any specific corruption. Instead, it represents real-world data that share the same label set as CIFAR 10. Compared to other corruption domains — except the brightness domain — CIFAR 10.1 exhibits a smaller domain gap, as evidenced by the lower error rate achieved through direct inference. In this context, most methods yield results comparable to direct inference. Notably, \texttt{Conj-CE} excels with a batch size of 16, while \texttt{CoTTA$*$-LN}, \texttt{CoTTA-LN}, and \texttt{SAR} stand out when the batch size is reduced to 1. 

\textbf{Batch Sizes.} Batch size is found to aid pure optimization-based methods like \texttt{Tent}, \texttt{Conj-CE}, and \texttt{MEMO-LN} in outperforming direct inference (i.e., source only). Similarly, batch size significantly influences the performance of \texttt{ROID}. It achieves the highest performance, with a significant margin over others, in terms of mean error on CIFAR-10-C when the batch size is 16. Nevertheless, it makes almost entirely erroneous predictions under a batch size of 1. As detailed in Section \ref{sec:bs-selection}, our analysis leads to the conclusion that larger batch sizes tend to stabilize loss optimization, which is beneficial for adaptation processes in these methods. This phenomenon is also evident in CIFAR-10.1, where entropy-based methods such as \texttt{Tent} and \texttt{Conj-CE} face challenges at smaller batch sizes. 

However, methods that incorporate prediction reliability can effectively navigate the limitations associated with smaller batch sizes. For instance, \texttt{SAR} capitalizes on sharpness-aware minimization to enhance its robustness. Similarly, \texttt{CoTTA-LN} employs selective augmentation in tandem with a reset strategy, averting erroneous optimization paths. In addition, \texttt{RoTTA} employs a well-designed memory bank to ensure batch-level stability.
Conversely, \texttt{TAST} underperforms compared to the source-only baseline, even when utilizing larger batch sizes. This could be due to its methodology being specifically tailored for the ResNet family, which may not translate as effectively to other architectures.

% The memory bank in \texttt{RoTTA} contributes to maintaining global information, rendering it more batch-agnostic. From a different angle, \texttt{SAR} achieves flat minima, which ensures model optimization for stability and prevents biased learning during adaptation. \texttt{MEMO} also displays impressive performance in certain domains, even with batch sizes as small as $1$. 

\begin{table*}[!h]\caption{Classification error rate ($\%$) for the standard CIFAR-100 $\rightarrow$ CIFAR-100-C online test-time adaptation task. Results are
evaluated on ViT-base-patch16-224 with the largest corruption severity level 5. Here, SO means source only.} \label{tab. cifar-100-c-exp} % MEMO=MEMO-LN
    \centering
    \resizebox{1\linewidth}{!}{% 
    \begin{tabular}{l cccc cccc cccc cccc}
    \toprule
        Method  & Glass & Fog &Defoc. &Impul. &Contr. &Gauss. &Elastic. &Zoom &Pixel &Frost &Snow &JPEG &Motion &Shot &Brit. &Mean  \\ 
        \midrule
        SO & 62.39	& 34.23 & 32.01 & 76.94 & 60.32 & 76.05 & 28.86 & 25.25 & 39.73 & 24.20 & 19.79 & 35.62 & 32.56 & 72.86 & 7.32 & 41.88 \\
 \midrule
    \multicolumn{17}{c}{\centering \textcolor{blue}{Batch size = 16}}\\
    \texttt{Tent} & 73.69 & 35.40 & 16.33 & 97.06 & 85.94 & 91.95 & 24.4 & 15.37 & 26.78 & 23.54 & 17.44 & 33.36 & 22.09 & 93.71 & 5.53 & 44.17 \\
    \texttt{CoTTA-LN} & 62.68 & 41.84 & 38.77 & 87.01 & 74.04 & 80.53 & 34.62 & 30.46 & 36.46 & 26.47 & 22.08 & 39.70 & 39.89 & 77.25 & 7.58 & 46.63\\
    \texttt{CoTTA-ALL} & 68.75 & 79.77 & 45.56 & 98.58 & 97.81 & 94.80 & 65.83 & 53.62 & 38.73 & 37.63 & 24.49 & 67.40 & 86.00 & 94.18 & 13.19 & 64.42\\
    \texttt{CoTTA$*$-LN} & 63.12 & 42.25 & 39.10 & 87.11 & 73.93 & 80.57 & 34.94 & 30.71 & 36.75 & 26.30 & 22.42 & 39.79 & 40.22 & 77.47 & 7.60 & 46.82 \\
    \texttt{CoTTA$*$-ALL} & 91.58 & 91.85 & 83.09 & 98.61 & 97.91 & 97.39 & 91.57 & 90.03 & 86.26 & 83.19 & 79.11 & 90.57 & 93.47 & 97.00 & 72.50 & 89.61\\
    \texttt{SAR} & 63.09 & 31.41 & 30.72 & 82.17 & 64.76 & 79.59 & 27.39 & 24.36 & 34.76 & 24.10 & 19.61 & 34.01 & 29.26 & 77.55 & 7.32 & 42.01 \\
    \texttt{Conj-CE} & 86.24 & 37.12 & 13.06 & 97.87 & 94.18 & 97.49 & 22.73 & 11.07 & 18.64 & 33.83 & 14.13 & 31.64 & 16.06 & 97.03 & 4.86 & 45.06 \\
    \texttt{MEMO-LN} & 83.03 & 68.62 & 17.49 & 96.33 & 30.39 & 94.71 & 25.19 & 16.58 & 28.71 & 19.93 & 17.13 & 34.18 & 20.07 & 92.96 & 5.38 & 43.38 \\
    \texttt{RoTTA} & 62.22 & 33.34 & 30.92 & 76.12 & 58.78 & 75.42 & 28.51 & 24.41 & 39.61 & 24.10 & 19.55 & 35.49 & 31.59 & 72.36 & 7.14 & 41.30 \\
    \texttt{TAST} & 70.09 & 65.01 & 58.18 & 79.09 & 84.37 & 76.04 & 50.82 & 52.14 & 49.78 & 45.77 & 41.75 & 48.93 & 58.68 & 72.76 & 20.65 & 58.27\\
    \texttt{ROID} & 54.30 & 24.63 & 19.05 & 57.87 & 33.67 & 60.84 & 24.96 & 17.25 & 28.49 & 19.64 & 17.74 & 32.55 & 22.15 & 57.20 & 5.88 & 31.75\\
    
        \midrule
    \multicolumn{17}{c}{\centering \textcolor{blue}{Batch size = 1}}\\
    \texttt{Tent} & 94.66 & 63.74 & 20.36 & 98.33 & 96.17 & 97.72 & 42.98 & 50.11 & 23.65 & 65.41 & 29.05 & 69.70 & 83.09 & 96.28 & 8.43 & 62.65 \\
    \texttt{CoTTA-LN} & 61.94 & 42.16 & 39.05 & 82.87 & 69.13 & 79.16 & 33.02 & 31.30 & 35.29 & 24.84 & 21.57 & 37.19 & 39.32 & 75.95 & 9.32 & 45.47 \\
    \texttt{CoTTA-ALL} & 96.95 & 93.78 & 95.34 & 98.90 & 97.82 & 98.86 & 92.47 & 93.33 & 96.36 & 95.66 & 98.42 & 95.79 & 96.81 & 98.68 & 98.04 & 96.48 \\
    \texttt{CoTTA$*$-LN} & 66.87 & 46.04 & 42.94 & 85.11 & 71.35 & 82.93 & 39.81 & 35.38 & 38.61 & 30.09 & 27.39 & 45.03 & 44.50 & 79.71 & 11.59 & 49.82 \\
    \texttt{CoTTA$*$-ALL} & 98.62 & 98.36 & 98.26 & 98.87 & 98.60 & 98.95 & 98.11 & 97.64 & 98.63 & 98.25 & 98.62 & 98.38 & 98.04 & 98.90 & 98.49 & 98.45 \\
    \texttt{SAR} & 62.39 & 33.95 & 31.29 & 80.15 & 60.48 & 77.82 & 28.63 & 24.83 & 38.60 & 24.12 & 19.87 & 35.51 & 31.73 & 74.89 & 7.29 & 42.10 \\
    \texttt{Conj-CE} & 97.07 & 73.10 & 18.54 & 98.67 & 97.21 & 98.25 & 43.50 & 10.90 & 31.95 & 46.68 & 14.56 & 82.12 & 16.89 & 98.18 & 4.75 & 55.49 \\
    \texttt{MEMO-LN} & 98.08 & 86.99 & 10.42 & 98.69 & 96.71 & 98.64 & 24.14 & 9.35 & 12.94 & 91.24 & 61.54 & 64.93 & 13.12 & 98.63 & 4.45 & 57.99 \\
    \texttt{RoTTA} & 62.22 & 33.34 & 30.92 & 76.11 & 58.78 & 75.41 & 28.51 & 24.40 & 39.63 & 24.10 & 19.55 & 35.49 & 31.59 & 72.36 & 7.14 & 41.30 \\
    \texttt{TAST} & 70.52 & 65.33 & 58.90 & 79.33 & 84.50 & 76.18 & 51.21 & 52.46 & 49.93 & 46.23 & 41.92 & 49.08 & 59.31 & 72.95 & 20.79 & 58.58 \\
    \texttt{ROID}& 98.99 & 98.99 & 98.99 &99.00 & 98.99 & 99.00 & 98.99 & 98.99 & 98.99 & 99.00 & 99.00 & 98.99 & 98.99 & 99.00 & 99.00 & 98.99 \\
        \bottomrule
    \end{tabular}}
\end{table*}
\subsubsection{On CIFAR-100-C Benchmark} \label{sec:cifar100c}

The performance on the CIFAR-100-C dataset exhibits a trend similar to that observed in the CIFAR-10-C dataset, as shown in Table \ref{tab. cifar-100-c-exp}.

\textbf{Number of Classes. } As CIFAR-100-C shares the same corruption setup as CIFAR-10-C, it is important to understand the differences between the CIFAR-10-C and CIFAR-100-C datasets. CIFAR-10-C consists of only 10 classes, making it possible for all classes to be represented within a single batch if its size is 16. In contrast, CIFAR-100-C, with its 100 classes, introduces a distinct challenge for online streaming adaptation. This scenario is particularly problematic for methods like \texttt{Tent}, which rely on updating the \texttt{LayerNorm} parameters within the current batch for subsequent use. While none can surpass the source-only performance, \texttt{Tent} experiences a 1.33$\%$ accuracy reduction on CIFAR-10-C and a more pronounced 3.94$\%$ reduction on CIFAR-100-C.

\textbf{Batch Sizes. }Besides, a noteworthy finding from the CIFAR-100-C experiments is that \texttt{RoTTA}, with a mean error rate of 41.30$\%$ at any experimented batch size, consistently outperforms direct inference (which has a mean error rate of 41.88$\%$). This achievement is likely due to \texttt{RoTTA}'s ability to maintain label diversity within its memory bank, emphasizing the importance of preserving a wide and varied information spectrum to tackle batch-sensitive and complex adaptation tasks effectively. This trend is similarly observed in Fig. \ref{fig:bs_cifar10c} and Fig. \ref{fig:imagenet}, further reinforcing the robustness of \texttt{RoTTA}. Moreover, although \texttt{SAR} and \texttt{TAST} underperform direct inference, they demonstrate stability across various batch sizes. Considering \texttt{RoTTA}, these methods represent the primary approaches to handling different batch sizes: 1. stable optimization, as the flat minima are more resilient to gradient fluctuations; and 2. information preservation, which can provide additional insights for each batch, reducing model sensitivity to batch sizes. Nevertheless, \texttt{ROID} and \texttt{CoTTA-ALL} exhibits significant performance variations with respect to batch sizes.

\subsubsection{On Imagenet-C Benchmark}
% when image size -> 224, if it is the same observation?
\begin{figure*}
    \centering
    \includegraphics[width=0.95\linewidth]{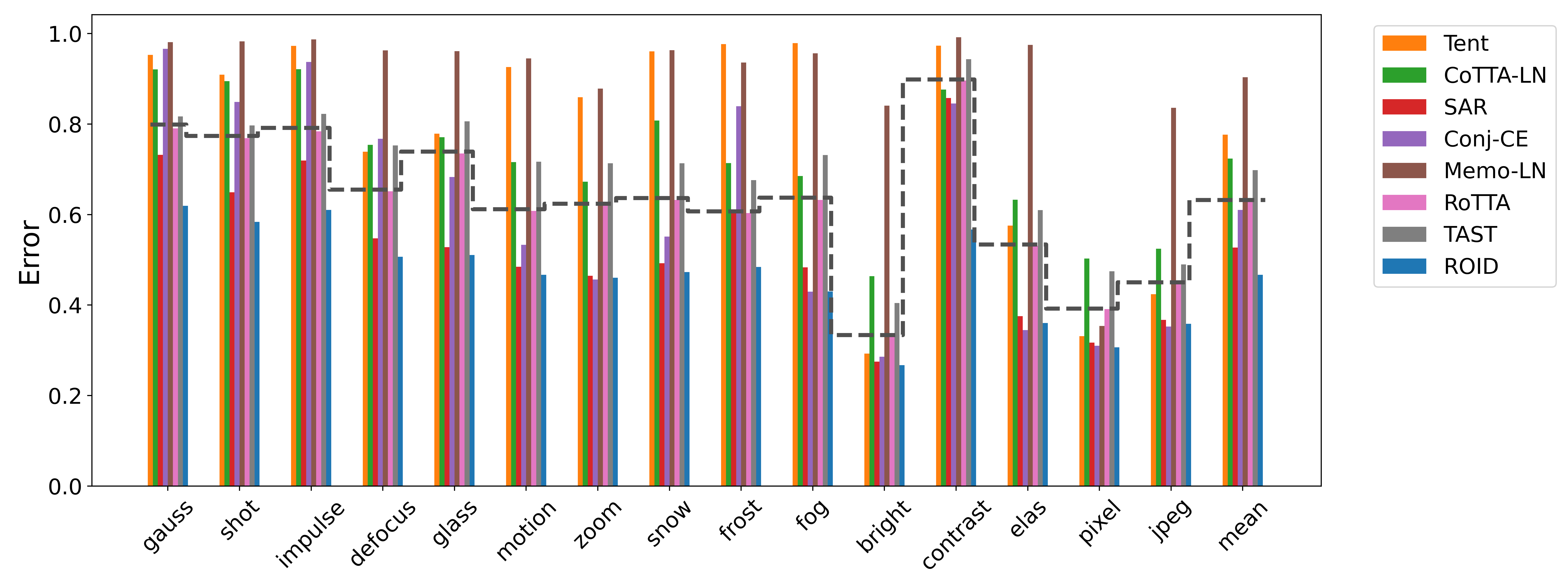}
    \includegraphics[width=0.95\linewidth]{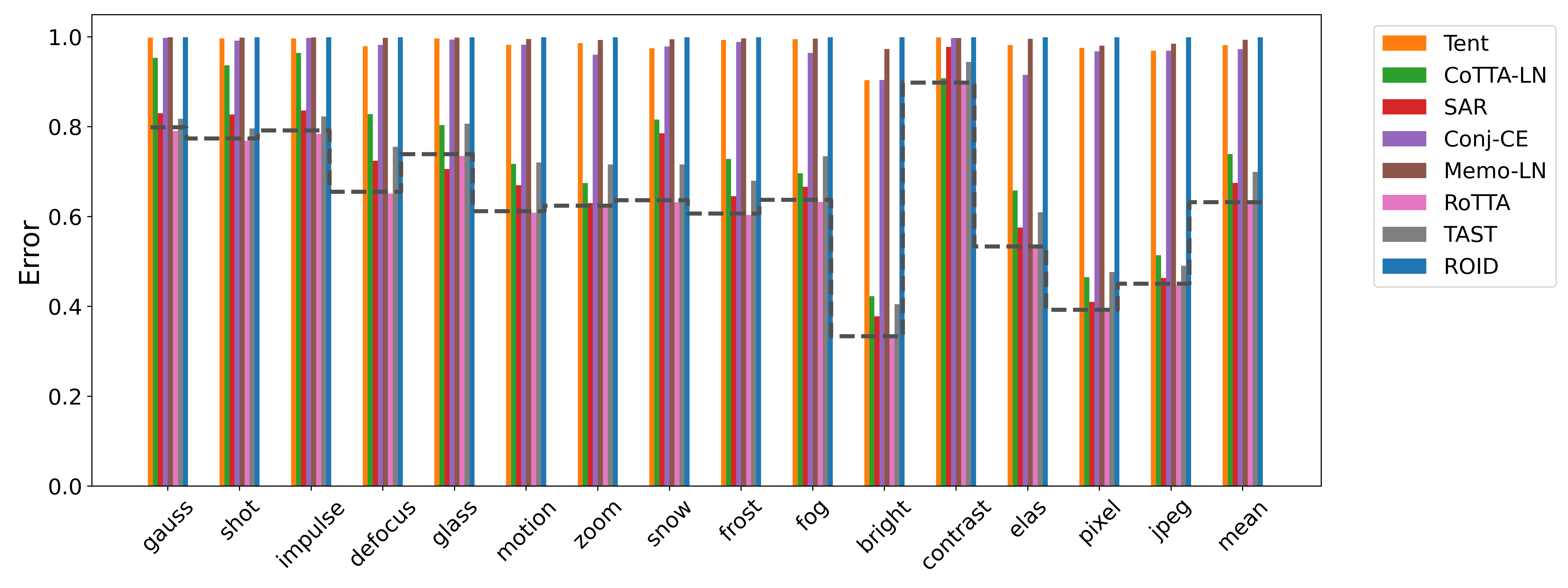}
    \caption{Comparison of the OTTA performance on the ImageNet-C severity level $5$. The upper and bottom plots show the experiments conducted with batch size $16$ and $1$ based on the \texttt{LayerNorm} updating strategy, respectively. The dotted line represents the source-only performance.}
    \label{fig:imagenet}
\end{figure*}

When comparing performance across CIFAR-10-C, CIFAR-100-C, and ImageNet-C, it becomes apparent that ImageNet-C exhibits notably poorer performance overall. This could be attributed to the larger number of classes in the ImageNet-C dataset, as a similar trend is observed when comparing it with CIFAR-10-C and CIFAR-100-C.

\textbf{Batch Sizes. }When batch size is minimized to 1, optimization-based methods exhibit error rates exceeding 95$\%$ across all domains. This also applies to \texttt{ROID} which mainly relies on model optimization. An exception to this trend is \texttt{SAR}, which produces a similar trend as in CIFAR-100-C. However, it produces a larger performance gap in ImageNet-C than CIFAR-100-C compared with direct inference. With different numbers of classes, this further indicates that datasets with increased complexity and a higher degree of challenge exhibit greater sensitivity to batch-size alterations.
%%%% here start

\textbf{Corruption Types. }  Varying domain gaps emerge across datasets when employing the same mechanism to generate corrupted data. Nevertheless, overall consistency remains, likely influenced by the training process of the source model. Additionally, corruptions representing natural domain gaps (e.g., brightness, zoom, snow) consistently appear simpler than noise-based corruptions (e.g., Gaussian noise, shot noise, etc.). Surprisingly, contrast exhibits a notably high error rate for both direct inference and most methods, indicating a substantial domain gap compared to other corruption types. In this case, \texttt{ROID} appears to excel in addressing it. In contrast, brightness appears to have the smallest domain gap with the source data.

\textbf{Adaptation Strategy. }As depicted in Fig. \ref{fig:imagenet}, when the batch size is set to 16, \texttt{SAR}, \texttt{Conj-CE}, \texttt{RoTTA}, and \texttt{ROID} outperform the source-only model in terms of mean error. In contrast, \texttt{Tent}, \texttt{MEMO-LN}, and \texttt{CoTTA-LN} demonstrate significantly poor results. The error rate within each domain exhibits a similar trend. Notably, \texttt{Conj-CE}, which conducts an additional inference of each batch for final prediction compared with \texttt{Tent}, markedly surpasses \texttt{Tent} across most domains and in mean error. This suggests a significant inter-batch shift in ImageNet-C, such as class differences, indicating that strategies based on optimizing the current batch for predicting the next batch (such as \texttt{Tent}) may be less effective.

\subsubsection{On CIFAR-10-Warehouse Benchmark}

In our study, the CIFAR-10-Warehouse dataset emerges as an indispensable resource for assessing OTTA methods, perfectly aligning with the CIFAR-10 label set to facilitate comprehensive comparisons across a spectrum of distribution shifts. Our detailed examination of two distinct domains within this dataset — real-world shifts and diffusion synthesis shifts — aims to probe deeper into the resilience and adaptability of OTTA methods amidst diverse distribution challenges.

\textbf{\underline{Google split. }}Diverging from previous CIFAR-10 variances, which primarily integrated artificially induced corruptions (i.e., CIFAR-10-C) or relied on sample differences (i.e., CIFAR 10.1), the Google split of the CIFAR-10-Warehouse offers an unparalleled perspective. Sourced from Google search queries, this segment includes 12 subdomains that exhibit a range of color variations within the CIFAR-10 labels. It constructs an essential benchmark to measure contemporary OTTA methods' proficiency against real-world distribution shifts.

\begin{figure*}
    \centering
    \includegraphics[width=0.95\linewidth]{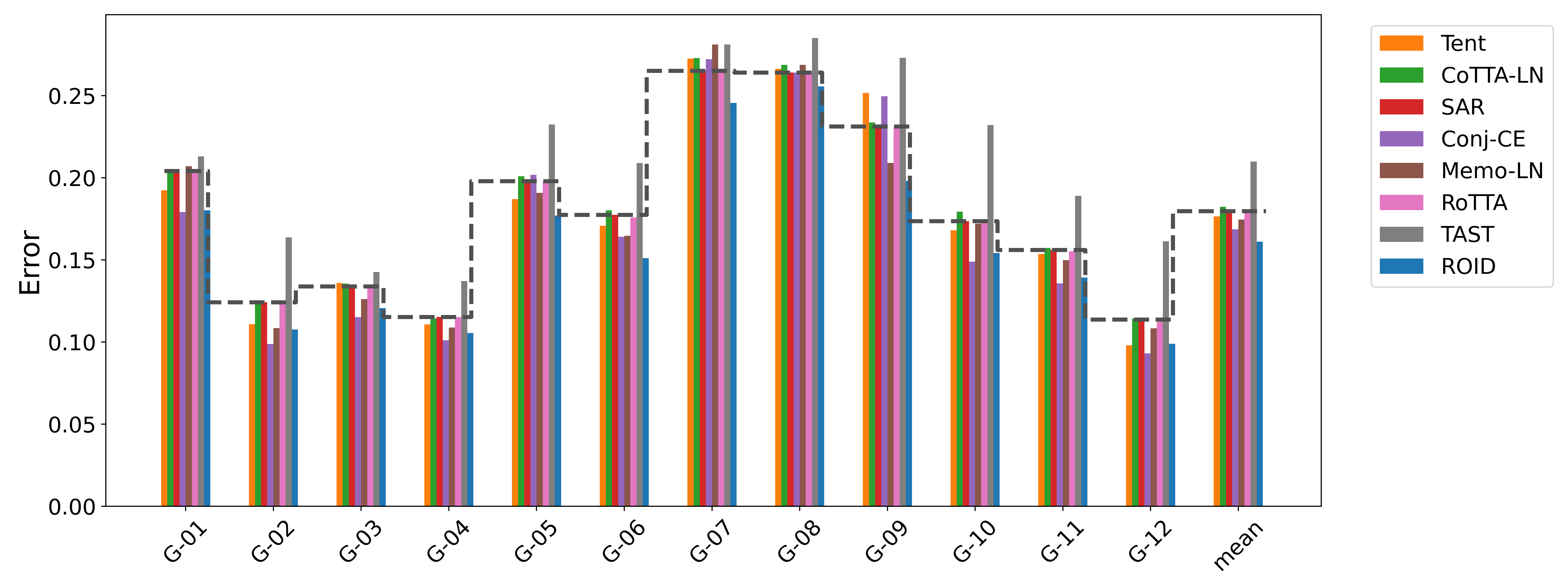}
    \includegraphics[width=0.95\linewidth]{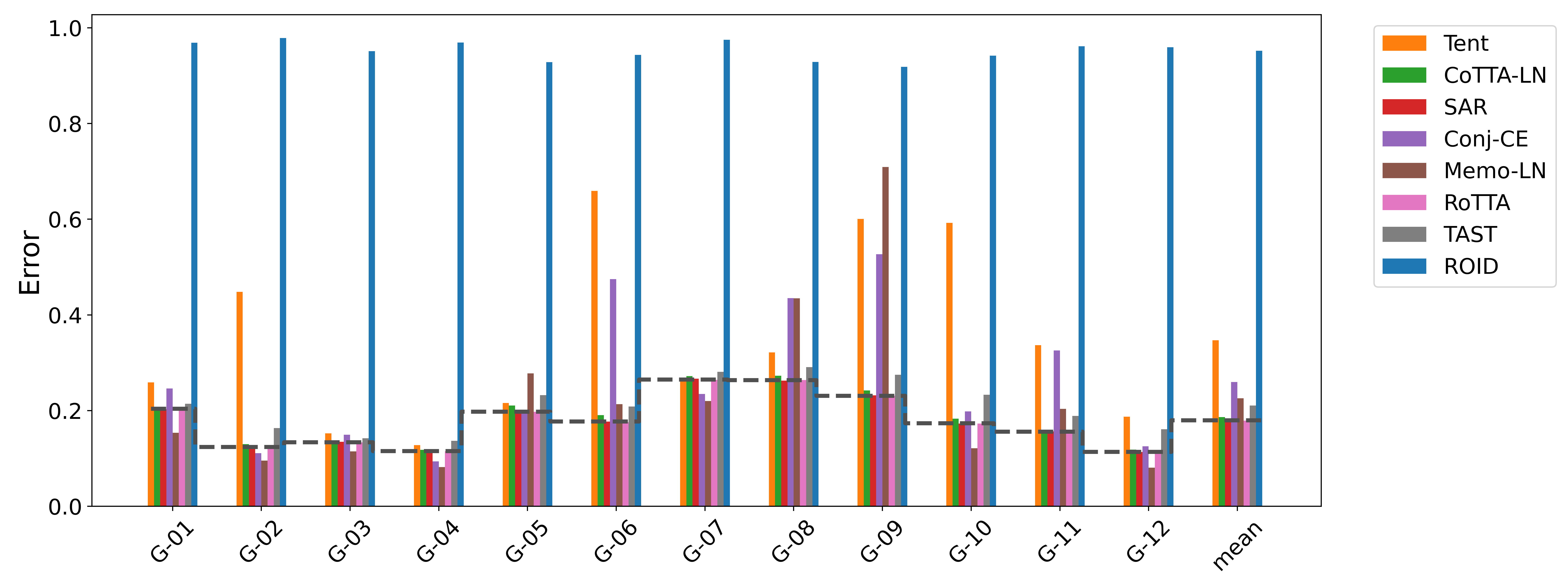}
    \caption{Comparison of the OTTA performance on the \textbf{Google split} of CIFAR-10-Warehouse. The upper and bottom plots show the experiments conducted with batch size $16$ and $1$ based on the \texttt{LayerNorm} updating strategy. The dotted line represents the source-only performance.}
    \label{fig:CIFAR-10-W}
\end{figure*}

\textbf{Cross-dataset Comparison: Real-world \textit{v.s.} Corruptions. } Comparing the empirical results for both the Google split and CIFAR-10-C, a notable difference is that the real-world shift presented in CIFAR-10-Warehouse exhibits a smaller domain gap globally, as indicated by the performances of direct inference. Concerning subdomains, the real-world shifts also demonstrate lower variance compared to corruption domains. We can infer that dealing with attacks or noises in a streaming manner is the most challenging adaptation task, as these two datasets are based on the same source pre-trained model.

\textbf{Batch Sizes.} Regarding the differences in batch sizes depicted in Fig. \ref{fig:CIFAR-10-W}, it is observed that when the batch size is 16, five OTTA methods based on \texttt{LayerNorm} updating either match or surpass the performance of direct inference. This observation suggests that the \texttt{LayerNorm} updating strategy is generally effective in handling real-world shifts. However, when the batch size is reduced to 1, methods that solely rely on entropy minimization, such as \texttt{Tent} and \texttt{Conj-CE}, experience performance degradation across most domains. This decline may be attributed to the instability of optimization for single-sample batches. Additionally, the fact that \texttt{ROID} performs similarly to corruption-based datasets indicates its sensitivity to changes in batch size.

\textbf{Adaptation Strategy.} \texttt{RoTTA}, \texttt{TAST} and \texttt{SAR} demonstrate exceptional stability regardless of batch sizes, which is also shown in CIFAR-100-C. Furthermore, \texttt{ROID} exhibits a smaller margin for surpassing performance compared to corruption-based datasets, indicating its robustness against data noise.

\begin{table*}[!ht]\caption{Classification error rate ($\%$) for the CIFAR-10 $\rightarrow$ CIFAR-10-Warehouse online test-time adaptation task. Results are
evaluated on ViT-base-patch16-224 with the diffusion split.} \label{tab. cifar-w-d-exp}
    \centering
    \resizebox{1\linewidth}{!}{% 
    \begin{tabular}{l cccc cccc cccc c}
    \toprule
     Diffusion & DM-01 & DM-02 & DM-03 & DM-04 & DM-05 & DM-06 & DM-07 & DM-08 & DM-09 & DM-10 & DM-11 & DM-12 & Mean \\ 
        \midrule
        Source-Only & 1.84	&1.36&	3.05&	2.32&	5.16&	3.80&	3.18&	2.82&	1.18&	4.41	&2.43	&2.23& 2.82\\
 \midrule
    \multicolumn{14}{c}{\centering \textcolor{blue}{Batch size = 16}}\\
    \texttt{Tent} & 1.61 & 1.05 & 2.55 & 1.36 & 7.02 & 2.41 & 1.91 & 2.57 & 1.09 & 4.32 & 1.68 & 2.30 & 2.49 \\
    \texttt{CoTTA-LN} & 1.84 & 1.36 & 3.05 & 2.34 & 5.16 & 3.86 & 3.21 & 2.82 & 1.16 & 4.39 & 2.48 & 2.21 & 2.82 \\
    % \texttt{CoTTA-ALL} & 13.80 & 15.07 & 9.46 & 10.41 & 44.77 & 17.86 & 21.18 & 13.93 & 1.43 & 44.20 & 2.73 & 20.25 & 17.92 \\ % this is the update all setting
    \texttt{SAR} & 1.84 & 1.36 & 3.05 & 2.32 & 5.16 & 3.80 & 3.18 & 2.82 & 1.18 & 4.41 & 2.43 & 2.23 & 2.82 \\
    \texttt{Conj-CE} & 1.59 & 0.86 & 1.77 & 0.86 & 5.96 & 1.73 & 1.77 & 2.07 & 0.64 & 2.52 & 1.21 & 1.98 & 1.91 \\
    % \texttt{Conj-Poly} & 1.59 & 0.86 & 1.77 & 0.86 & 5.96 & 1.73 & 1.77 & 2.07 & 0.64 & 2.52 & 1.21 & 1.98 & 1.91 \\
    \texttt{MEMO-LN} & 1.50 & 1.02 & 2.14 & 0.86 & 5.39 & 2.05 & 2.46 & 2.36 & 0.82 & 3.77 & 1.84 & 1.86 & 2.17 \\ % this is the memo-ln
     % \texttt{MEMO-ALL} & 90.91 & 90.91 & 90.91 & 90.91 & 90.91 & 90.89 & 90.91 & 90.89 & 90.91 & 90.89 & 90.66 & 90.91 & 90.88 \\
    \texttt{RoTTA} & 1.80 & 1.36 & 3.05 & 2.23 & 5.07 & 3.71 & 3.18 & 2.80 & 1.14 & 4.34 & 2.43 & 2.14 & 2.77 \\
    \texttt{TAST} & 1.48 & 1.77 & 2.68 & 1.68 & 4.98 & 2.41 & 3.09 & 2.96 & 1.73 & 4.14 & 3.11 & 2.84 & 2.74 \\
    \texttt{ROID} & 1.64 & 0.93 & 2.32 & 1.27 & 4.00 & 2.16 & 2.43 & 2.14 & 0.89 & 3.55 & 1.73 & 1.77 & 2.07 \\
        \bottomrule
    \end{tabular}}
\end{table*}

\textbf{\underline{Diffusion split. }} Furthermore, we incorporate the Diffusion split into our analysis. Created using stable diffusion \cite{Rombach_2022_CVPR}, this domain introduces a novel examination of artificially generated samples. Considering the increasing prevalence of diffusion-generated imagery, this evaluation offers unique insights into the capability of OTTA methods to adapt to the rising tide of generated images encountered in real-world applications. 

\textbf{Cross-dataset Comparison: Corruption, Real-world, or Diffusion? }In terms of mean error, the diffusion split has the smallest domain gap among CIFAR-10-based datasets. This is even true when compared to the CIFAR 10.1 dataset, suggesting that OTTA methods based on ViT can generally handle diffusion-based images.

\textbf{Adaptation Strategy. }
The outcomes in Table \ref{tab. cifar-w-d-exp} indicate that, on average, each OTTA method performs equally well or better than the baseline direct inference in terms of mean error. However, some methods still exhibit poor performance in certain domains. For instance, while \texttt{TAST} delivers satisfactory results on DM-01, it fails to perform on DM-02, DM-08, DM-09, and DM-12 domains. Similarly, \texttt{Tent}, \texttt{Conj-CE}, and \texttt{MEMO-LN} methods fall short on DM-05.

In contrast, \texttt{RoTTA} and \texttt{ROID} demonstrate their capability to surpass direct inference outcomes for every domain. Similarly, \texttt{SAR} enables the model to reach a region in the optimization landscape that is less sensitive to data variations, resulting in stable predictions.

\begin{table}[h]
\centering
\caption{Classification error rate ($\%$) for the OfficeHome online test-Time adaptation task using a source model trained on Clipart with ViT-base-patch16-224. SO denotes source-only. } 
% before iter 16000
\resizebox{0.95\linewidth}{!}{% 
\begin{tabular}{l | c c c c}
\toprule
Method  & Art & Product & RealWorld & Mean \\
\midrule
SO & 39.56 & 34.42 & 29.54 & 34.51 \\
\midrule
& \multicolumn{4}{c}{\centering \textcolor{blue}{Batch size = 16}}\\
\texttt{Tent} & 39.10 & 33.63 & 28.69 & 33.81\\
\texttt{CoTTA-ALL} & 72.23 & 86.30 & 79.55 & 79.36 \\ 
\texttt{SAR} & 39.02 & 33.30 & 28.85 & 33.72\\
\texttt{Conj-CE} & 43.47 & \textbf{32.28} & \textbf{27.01} & 34.26\\
\texttt{MEMO-LN} & 43.14 & 34.74 & 30.25 & 36.04\\
\texttt{RoTTA} & 39.56 & 34.42 & 29.47 & 34.48\\
\texttt{TAST} & 39.47 & 32.30 & 30.02 & 33.93\\
\texttt{ROID} & \textbf{37.74} & 33.03 & 28.23 & \textbf{33.00}\\
 \midrule
& \multicolumn{4}{c}{\centering \textcolor{blue}{Batch size = 1}}\\
\texttt{Tent} & 39.60 & 67.97 & 47.56 & 51.71\\
\texttt{CoTTA-ALL} & 96.25 & 97.63 & 97.41 & 97.10\\ 
\texttt{SAR} & \textbf{39.27} & 34.65 & 29.56 & 34.49\\
\texttt{Conj-CE} & 68.40 & 69.59 & 30.30 & 56.09\\
 \texttt{MEMO-LN} & 48.17 & 94.05 & 91.51 & 93.39\\
 \texttt{RoTTA} & 39.56 & 34.44 & \textbf{29.47} & 34.49\\
\texttt{TAST} & 39.43 & \textbf{32.28} & 30.11 & \textbf{33.94}\\
\texttt{ROID} & 96.95 & 98.22 & 98.03 & 97.73\\
\bottomrule
\end{tabular}}\vspace{-2.5ex}
\label{table:officehome}
\end{table}

\subsubsection{On OfficeHome Benchmark}

Our benchmark dataset, OfficeHome, includes four domains. In our experiments, we use the clipart domain as the source dataset and apply OTTA methods to the remaining three domains: Art, Product, and RealWorld.

\textbf{Subdomains. }
The Art domain seems the most difficult to adapt to from clipart based on the given pre-trained source model. However, even with minimal batch size, \texttt{SAR}, \texttt{RoTTA}, \texttt{TAST}, \texttt{ROID} show competitive results against the baseline in the Art domain. Conversely, in the Product domain, reducing batch sizes significantly impacts the performance of several methods. The RealWorld domain presents the most significant challenge for adaptation, with few methods surpassing direct inference across different batch sizes.

\textbf{Batch Sizes. }\texttt{SAR}, \texttt{RoTTA}, and \texttt{TAST} consistently outperform the source-only baseline irrespective of batch size, indicating stable performance. On the other hand, \texttt{Tent}, \texttt{Conj-CE}, and \texttt{ROID} show a decline in performance at a batch size of 1, suggesting a dependency on larger batch sizes for optimal results. 

\textbf{Adaptation Strategy. }
It is observed that both \texttt{CoTTA-ALL} and \texttt{MEMO-LN} do not perform well across all domains. This suggests that these models rely heavily on augmentations that are not effective enough to bridge the domain gap in OfficeHome. As a result, it is crucial to further examine current augmentation strategies to address domain discrepancies more efficiently, especially for some specific online environments.

\textbf{Conclusion.} Based on our extensive experiments, most OTTA methods exhibit similar behavioral patterns across various datasets. This consistency indicates the potential of contemporary OTTA techniques in effectively managing diverse domain shifts. Of particular note are two methods, namely \texttt{RoTTA} and \texttt{SAR}, which highlight the significance of optimization insensitivity and information preservation, respectively. Additionally, the effectiveness of \texttt{ROID} is demonstrated when the batch size is reasonable, showcasing its capability.
\begin{figure*}[t]
\centering
\subfloat{{\includegraphics[width=0.33\textwidth]{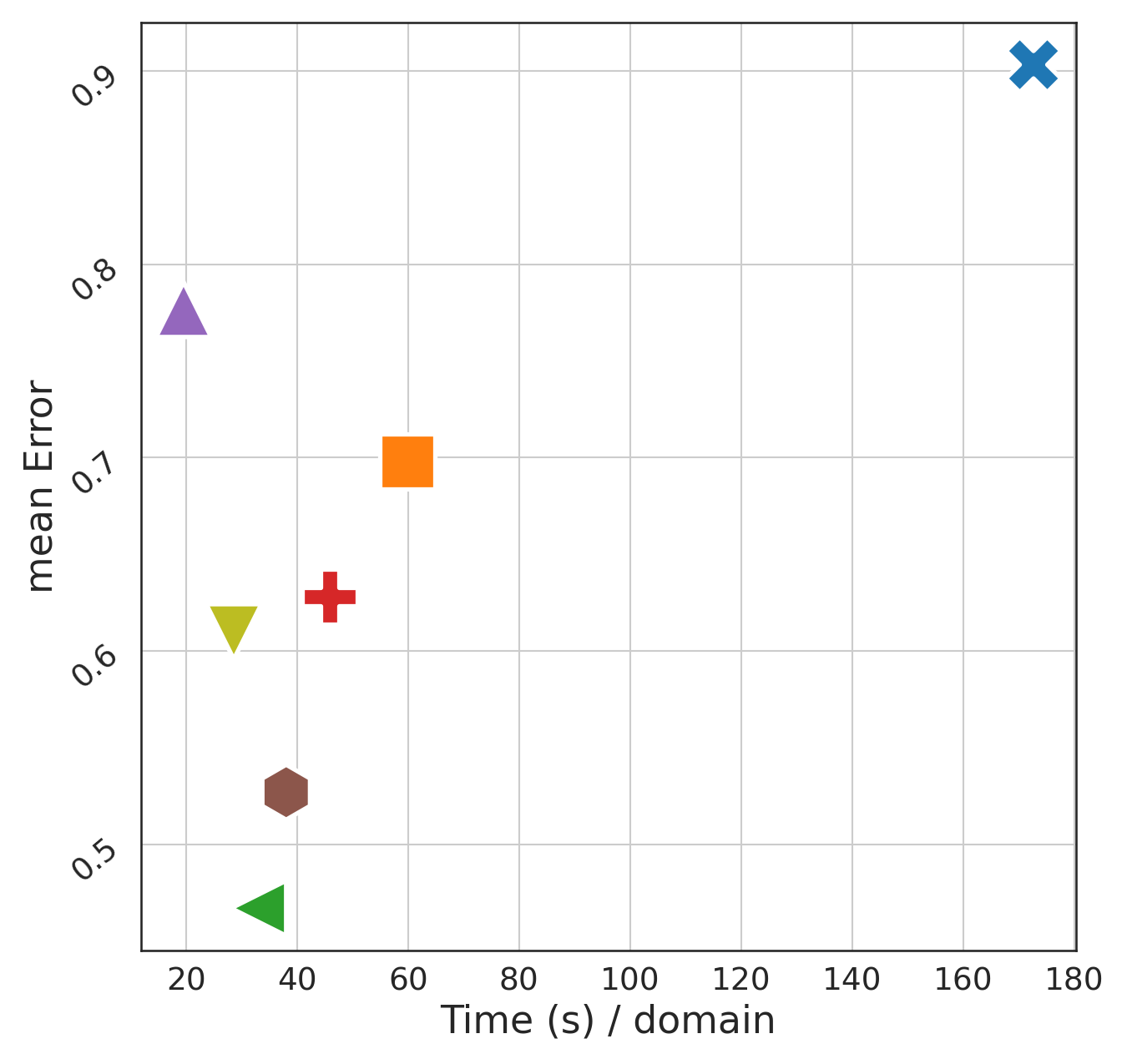} }}%
\subfloat{{\includegraphics[width=0.33\textwidth]{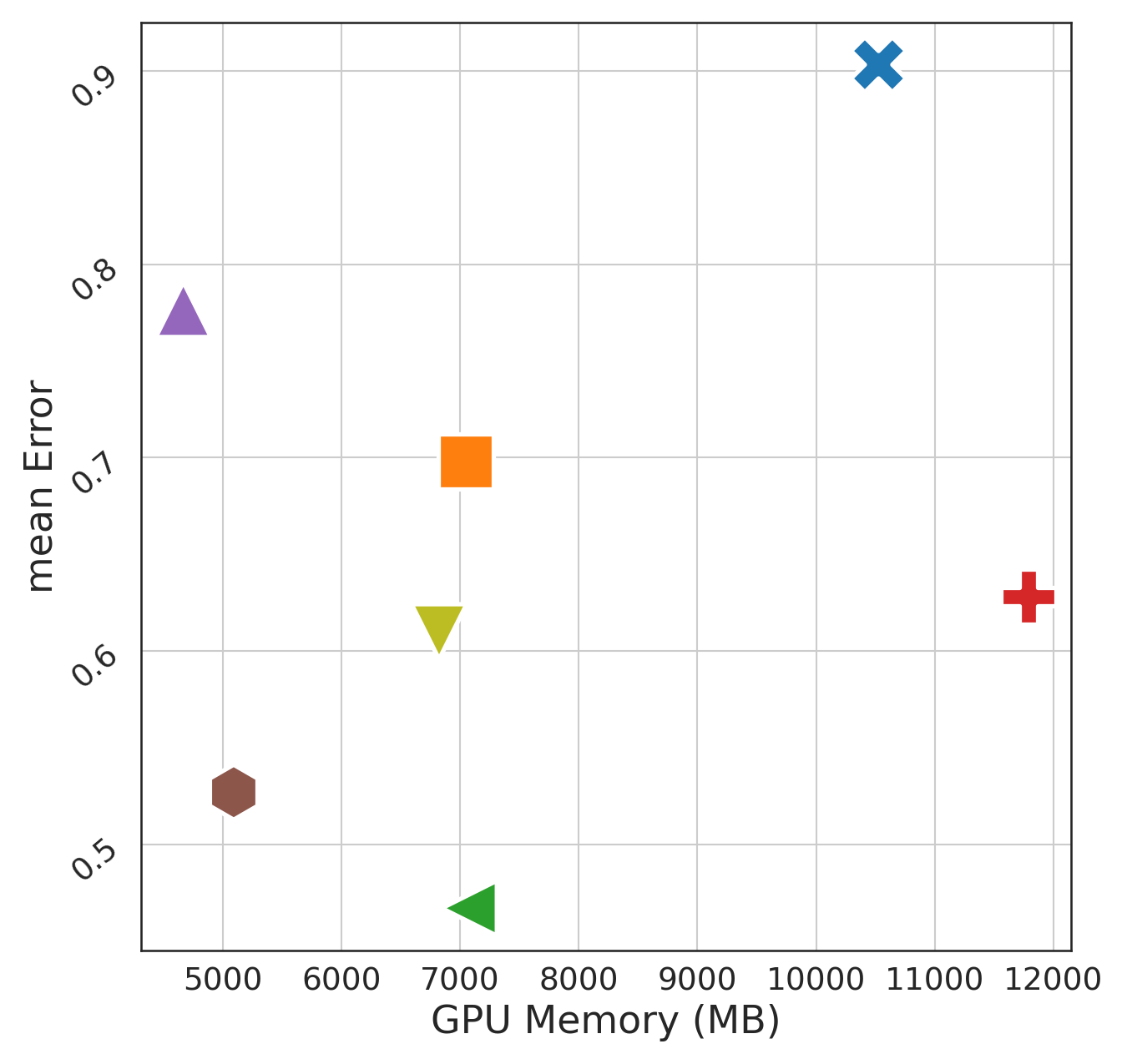} }}
\subfloat{{\includegraphics[width=0.33\textwidth]{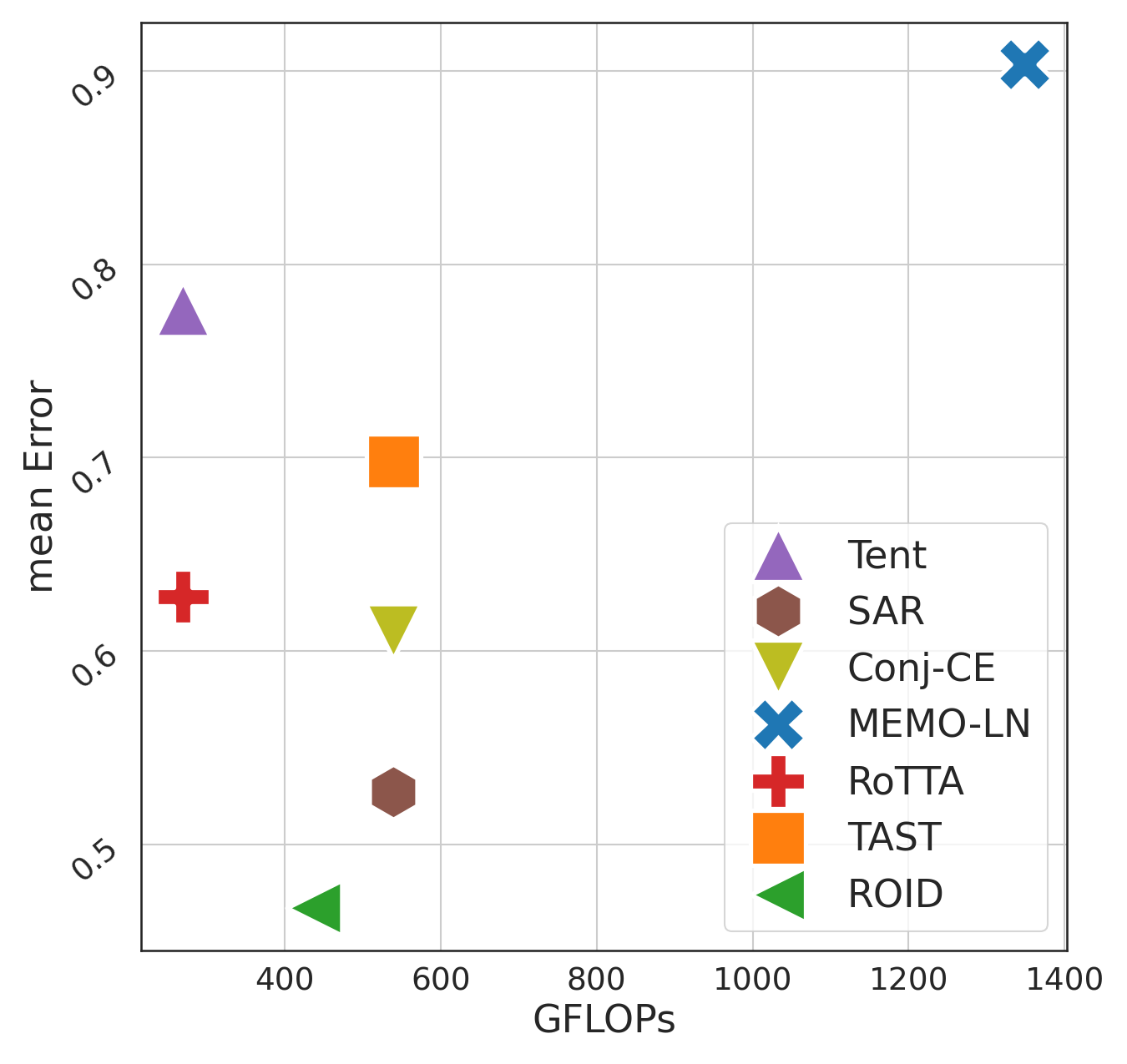} }}%
\caption{Mean error vs (a) average wall-clock time per domain, (b) GPU memory usage, and (c) GFLOPs. All experiments are conducted on a severity level of $5$ on the ImageNet-C dataset.} \label{fig:error_vs_flops}
\end{figure*}

% \vspace{-1em}
\subsection{Is OTTA efficient?} 
To assess the efficacy of OTTA algorithms, especially within the constraints of hardware limitations, we employ three primary evaluation metrics: wall clock time, GPU memory usage, and GFLOPs.

As depicted in Fig. \ref{fig:error_vs_flops}, lower values across these metrics are indicative of superior performance. Our analysis reveals that within the context of the ImageNet-C dataset, \texttt{MEMO-LN} exhibits suboptimal performance, accompanied by elevated computational demands, which are deemed disadvantageous. 

Meanwhile, \texttt{RoTTA} demonstrates commendable results, characterized by reduced GFLOPs and processing time; however, its memory bank demands greater GPU memory. This requirement may necessitate further hardware provisions or memory optimizations for deployment in practical applications. Note that in our analyses, \texttt{CoTTA-ALL} is excluded to maintain the integrity and informativeness of the comparative showcase. This decision is made since \texttt{CoTTA-ALL} has unexpectedly high GFLOPs and wall clock time cost, along with high error rates, which made the comparative landscape appear skewed. 

Conversely, \texttt{SAR} maintains low mean error rates while ensuring computational efficiency across all efficiency metrics, especially compared with \texttt{Tent}. Similarly, \texttt{Conj-CE} attains significant error reduction with slightly increased resource consumption, performing inference for each batch directly after every iteration. \texttt{ROID}, additionally, achieves a balance between effectiveness and efficiency, as evidenced by its notably low time consumption and GFLOPs, alongside moderate GPU memory usage, while still delivering a superbly low error rate.

\begin{figure}[t]
    \centering
    \includegraphics[width=0.95\linewidth]{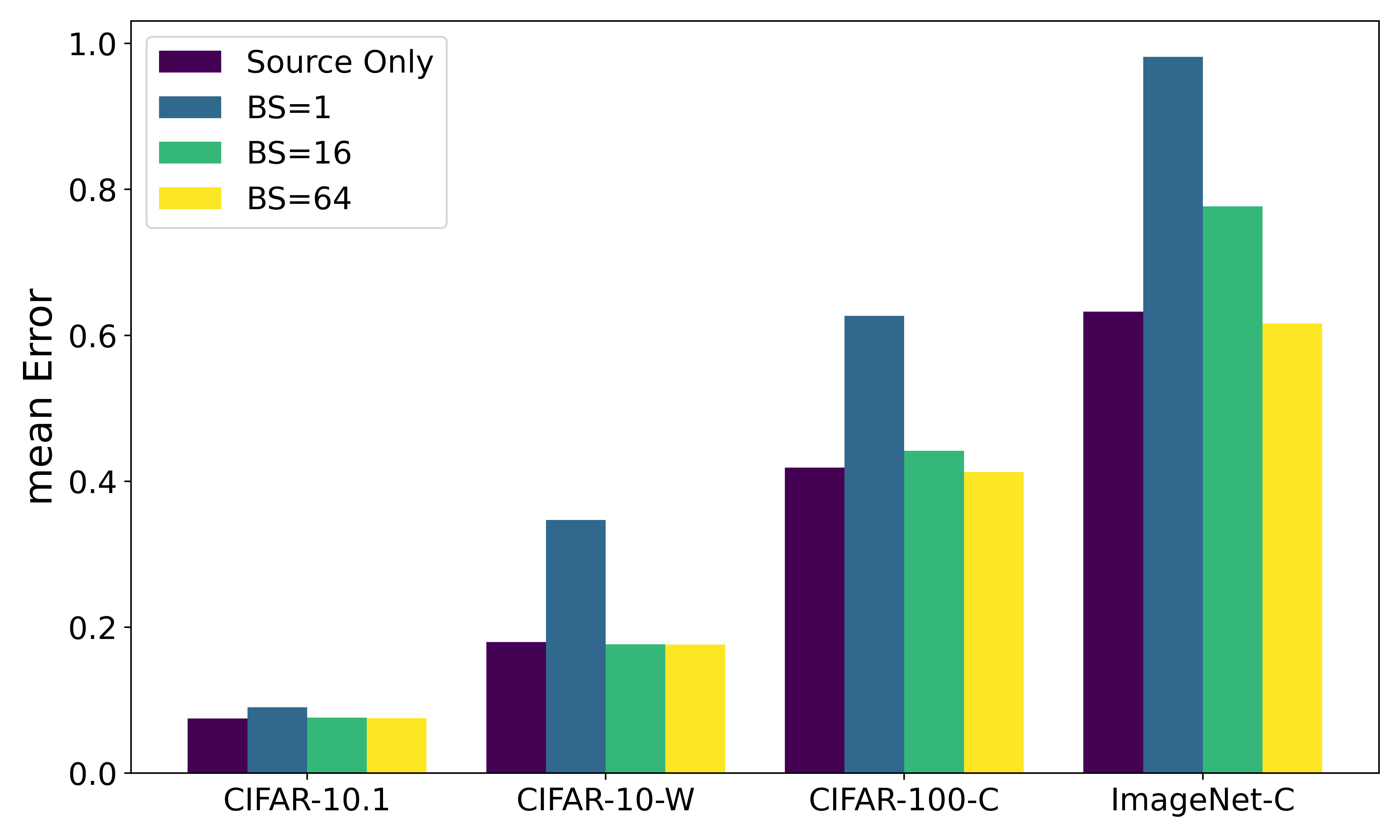}
    \caption{Impact of varying batch sizes for \texttt{Tent} on CIFAR-10.1, CIFAR-10-Warehouse google split, CIFAR-100-C, and ImageNet-C.}
    \label{fig:batch_all_dataset}
\end{figure}

\subsection{Is OTTA sensitive to Hyperparameter Selection?} \label{sec:bs-selection}

In this section, we investigate whether OTTA methods will be impacted by various hyperparameters. We conduct our experimental study on ImageNet-C with batch size 16 using Vit-base-patch16-224 to assess optimizers, learning rates, and schedulers.
\begin{figure*}
    \centering
    \includegraphics[width=0.95\linewidth]{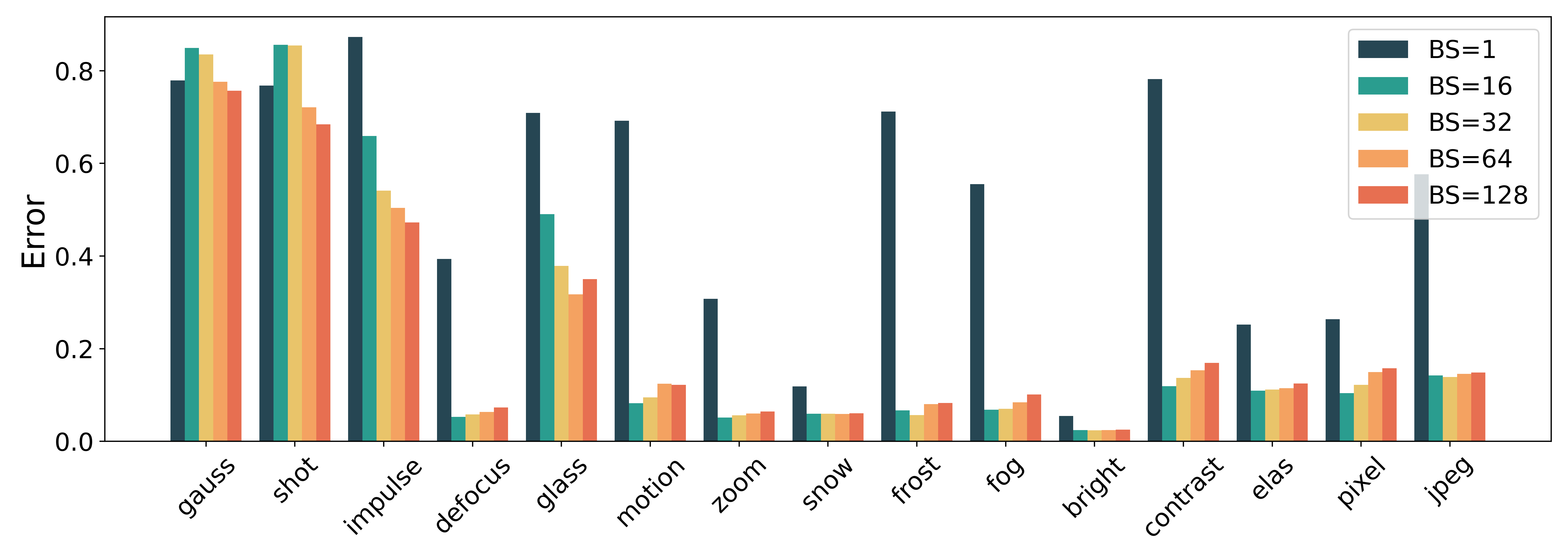}
    \caption{Impact of varying batch sizes on CIFAR-10-C severity level $5$. The base model is \texttt{Tent} optimized on \texttt{LayerNorm}.}
    \label{fig:bs_cifar10c}
\end{figure*}
\subsubsection{Batch Size Matters, but only to an extent.}
Fig. \ref{fig:bs_cifar10c} examines the impact of different batch sizes on \texttt{Tent} in the CIFAR-10-C dataset. It reveals that performance significantly varies with batch sizes from $1$ to $16$ across most corruptions. However, this variability diminishes with larger batch sizes ($16$ to $128$), indicating a reduced influence of \texttt{LayerNorm} updating on batch size, in contrast to traditional \texttt{BatchNorm} settings. This pattern is consistent across other datasets, as depicted in Fig. \ref{fig:batch_all_dataset}. Nevertheless, \textbf{batch size remains crucial for stabilizing the optimization process}. For example, a batch size of $16$ outperforms a batch size of $1$ in confidence optimization methods in the Google split of the CIFAR-10-Warehouse dataset. 

However, \textbf{Larger batch sizes are essential for complex datasets} like CIFAR-100-C and ImageNet-C, where direct inference struggles, emphasizing the need to tailor batch sizes to the data complexity. Additionally, Fig. \ref{fig:bs_cifar10c} suggests that \textbf{increasing batch sizes is not effective against challenging corruptions}, such as Gaussian and Shot noise. This indicates the necessity for more advanced adaptation strategies beyond mere batch size adjustments in complex learning scenarios. Additionally, it is worth noting that gradient accumulation \citep{DBLP:conf/wacv/MarsdenD024} could be treated as a solution to allow single-sample adaptation.

\begin{table*}[!htb]
\caption{The Comparison of model update strategies across five benchmarks: \texttt{Layernorm} only v.s. full model. $^*$ indicates the variant of \texttt{CoTTA} with no parameter reset mechanism.}\label{tab:layer_matter}
\resizebox{1\linewidth}{!}{% 
\begin{tabular}{lccccccccccccccc}
    \toprule
         Method & \multicolumn{3}{c}{\centering CIFAR-10-C} & \multicolumn{3}{c}{\centering CIFAR-10.1}  & \multicolumn{3}{c}{\centering CIFAR-100-C} & \multicolumn{3}{c}{\centering ImageNet-C} & \multicolumn{3}{c}{\centering CIFAR-Warehouse}  \\%& Input adaptation & Visual Prompt Tuning \\
       \cmidrule(lr){2-4}\cmidrule(lr){5-7}\cmidrule(lr){8-10}\cmidrule(lr){11-13}\cmidrule(lr){14-16}
          &LN &ALL &$\Delta$ Err &LN &ALL & $\Delta$ Err &LN &ALL &$\Delta$ Err &LN &ALL &$\Delta$ Err &LN &ALL &$\Delta$ Err  \\
        \midrule 
         \texttt{CoTTA}  &0.2471 & 0.5247 & \textcolor{green}{-52.91\%} &0.0750 &0.1155 &\textcolor{green}{-35.06\%} &0.4663 &0.6442 &\textcolor{green}{-27.62\%} &0.7237 &0.8516 &\textcolor{green}{-15.02\%} &0.1823 &0.2804 &\textcolor{green}{-34.99\%}   \\
          \texttt{CoTTA$*$}  &0.2468 & 0.7708 &\textcolor{green}{-67.98\%} &0.0750 &0.1640 &\textcolor{green}{-54.27\%} &0.4682 &0.8961 &\textcolor{green}{-47.75\%} &0.7229 &0.9054 &\textcolor{green}{-20.16\%} &0.1820&0.5463&\textcolor{green}{-66.68\%} \\
          \midrule
        \texttt{MEMO} &0.2116  &0.8999 &\textcolor{green}{-76.49\%}  &0.0780 &0.9000 &\textcolor{green}{-91.33\%} &0.4338 &0.9900 &\textcolor{green}{-56.18\%} &0.9032 &0.9995 &\textcolor{green}{-96.35\%} &0.1746 &0.8955 &\textcolor{green}{-80.50\%} \\

  \bottomrule
\end{tabular}}
\end{table*}

% \vspace{0.5cm}
\subsubsection{Optimization Layer Matters!}
To assess the critical role of \texttt{LayerNorm}, we compare \texttt{LayerNorm} update with the full model update, as summarized in Table. \ref{tab:layer_matter}. This ablation study primarily focused on \texttt{CoTTA} and \texttt{MEMO}, evaluating the impact of optimizing \texttt{LayerNorm} alone.
A notable observation is, for all methods, \texttt{LayerNorm} update plays an important role in gaining high performance, underscoring its effectiveness in boosting model performance by avoiding significant forgetting of the source knowledge. This is also proven by \texttt{Tent}, \texttt{SAR}, and \texttt{Conj-CE}, as shown in Fig. \ref{fig:layer-ablation} in the appendix.

\subsubsection{Optimizer matter?}
We evaluate Adam and SGD across eight methods with a batch size of 16 on the ImageNet-C dataset. Our findings, detailed in Table \ref{tab:sgd}, reveal several key observations. Firstly, methods that adapt to test data under soft supervision, such as soft entropy, appear more susceptible to changes in the optimizer. For instance, switching to SGD resulted in a significant mean error reduction of $9.84\%$ for \texttt{Tent}. Similarly, when updated by soft entropy, SGD enabled \texttt{CoTTA-ALL} to achieve a mean error difference of over $10\%$. This pattern was also observed in \texttt{Conj-CE} and \texttt{MEMO-LN}. This disparity likely stems from the superior generalization capabilities of SGD, as discussed by \cite{DBLP:conf/nips/WilsonRSSR17}. The notable exception within this group is \texttt{SAR}, which utilizes Sharpness-Aware Minimization (\texttt{SAM}) to find stable minima. This approach may diminish the impact of the choice between optimizers.

Another distinct category includes \texttt{RoTTA} and \texttt{TAST}. These methods, significantly reliant on their memory banks or the averaging of predictions across multiple adaptation modules, demonstrate reduced sensitivity to the choice of optimization strategy. \texttt{ROID} instead exhibits high sensitivity in terms of the optimizer changes. Here, Adam shows its stability due to its adaptive learning rate and moment estimation for noisy and non-stationary objectives.

\begin{table*}[!ht]
  \caption{Comparison between Adam and SGD (highlighted in\colorbox{red!10}{red}) on ImageNet-C using ViT-base-patch16-224 at severity level 5 and Batch Size 16, SO denotes source-only.} 
  \label{tab:sgd}
  \resizebox{1\linewidth}{!}{% 
    \begin{tabular}{l cccc cccc cccc cccc}
    \toprule
    Method & Glass & Fog &Defoc. &Impul. &Contr. &Gauss. &Elastic. &Zoom &Pixel &Frost &Snow &JPEG &Motion &Shot &Brit. &Mean \\
    \midrule
    Source-Only & 73.92 & 63.76 & 65.54 & 79.18 & 89.84 & 79.88 & 53.38 & 62.42 & 39.26 & 60.70 & 63.66 & 45.06 & 61.20 & 77.38 & 33.38 & 63.24 \\
     \midrule
    \texttt{Tent} & 77.86 & 97.88 & 73.88 & 97.24 & 97.28 & 95.28 & 57.54 & 85.92 & 33.12 & 97.64 & 96.06 & 42.40 & 92.60 & 90.88 & 29.30 & 77.66 \\
    \rowcolor{red!10} \texttt{Tent} & 74.86 & 95.34 & 65.28 & 87.18 & 95.04 & 87.58 & 38.44 & 50.74 & 31.82 & 92.78 & 88.74 & 36.92 & 56.98 & 87.98 & 27.64 & 67.82 \\
    \texttt{CoTTA-ALL} & 73.72 & 94.82 & 92.80 & 99.14 & 92.94 & 99.06 & 78.06 & 84.14 & 79.40 & 88.92 & 96.70 & 54.48 & 86.58 & 98.10 & 58.48 & 85.16 \\
    \rowcolor{red!10} \texttt{CoTTA-ALL} & 76.80 & 69.22 & 75.64 & 91.60 & 87.22 & 92.20 & 62.90 & 67.06 & 50.60 & 71.36 & 81.18 & 51.46 & 70.78 & 89.58 & 45.78 & 72.23 \\
    \texttt{SAR} & 52.78 & 48.34 & 54.74 & 71.92 & 85.74 & 73.18 & 37.54 & 46.48 & 31.70 & 61.14 & 49.24 & 36.74 & 48.48 & 64.92 & 27.52 & 52.70 \\
    \rowcolor{red!10} \texttt{SAR} & 53.34 & 45.68 & 52.52 & 82.40 & 98.30 & 72.24 & 37.72 & 47.82 & 31.42 & 49.90 & 47.82 & 37.10 & 48.86 & 77.32 & 27.52 & 54.00 \\
    \texttt{Conj-CE} & 68.30 & \textbf{42.96} & 76.74 & 93.70 & 84.52 & 96.60 & \textbf{34.46} & \textbf{45.64} & 31.00 & 83.90 & 55.12 & \textbf{35.28} & 53.32 & 84.88 & 28.60 & 61.00 \\
    \rowcolor{red!10} \texttt{Conj-CE} & 53.38 & 44.32 & 51.96 & \textbf{57.62} & 61.60 & \textbf{60.62} & 37.02 & 47.24 & 31.42 & 50.14 & 48.22 & 36.34 & 47.66 & \textbf{57.96} & 28.38 & 47.59 \\
    \texttt{MEMO-LN} & 96.10 & 95.62 & 96.26 & 98.68 & 99.18 &	98.08 & 97.48 & 87.80 & 35.40 & 93.58 & 96.30 & 83.56 & 94.50 & 98.26 & 84.06 & 90.32 \\
     \rowcolor{red!10} \texttt{MEMO-LN} & 93.16 & 88.46 & 96.10 & 97.60 & 98.66 & 96.72 & 95.86 & 85.20 & 33.10 & 94.42 & 94.70 & 57.30 & 89.48 & 98.66 & 33.36 & 83.52 \\
    % \hline
    \texttt{RoTTA} & 73.50 & 63.24 & 65.14 & 78.40 & 89.58 & 79.02 & 53.10 & 61.94 & 39.12 & 60.34 & 63.24 & 44.62 & 60.80 & 76.88 & 33.12 & 62.80 \\
    \rowcolor{red!10} \texttt{RoTTA} & 73.68 & 64.00 & 65.56 & 76.72 & 89.86 & 77.72 & 53.46 & 62.28 & 39.46 & 60.98 & 63.92 & 44.26 & 60.92 & 75.18 & 33.42 & 62.76 \\
    \texttt{TAST} & 80.60 & 73.16 & 75.26 & 82.22 & 94.32 & 81.68 & 60.98 & 71.34 & 47.48 & 67.62 & 71.34 & 48.98 & 71.68 & 79.66 & 40.46 & 69.79 \\ % updated version for tast
    \rowcolor{red!10} \texttt{TAST} & 80.54 & 73.04 & 75.32 & 82.18 & 94.30 & 81.74 & 60.90 & 71.28 & 47.46 & 67.52 & 71.42 & 48.88 & 71.80 & 79.66 & 40.22 & 69.75 \\
    \texttt{ROID} & \textbf{51.06} & 43.02 & \textbf{50.68} & 61.02 & \textbf{56.70} & 61.94 & 36.04 & 46.02 & \textbf{30.66} & \textbf{48.44} & \textbf{47.30} & 35.88 & \textbf{46.70} & 58.38 & \textbf{26.72} & \textbf{46.70}\\
    \rowcolor{red!10} \texttt{ROID} & 98.30 & 98.08 & 98.64 & 99.48 & 99.80 & 99.50 & 98.04 & 99.18 & 78.34 & 98.96 & 99.54 & 98.82 & 99.06 & 98.54 & 43.00 & 93.82\\
    \bottomrule
    \end{tabular}}
\end{table*}
\subsubsection{Learning Rate. } 
\begin{figure}[t]
    \centering
    \includegraphics[width=0.95\linewidth]{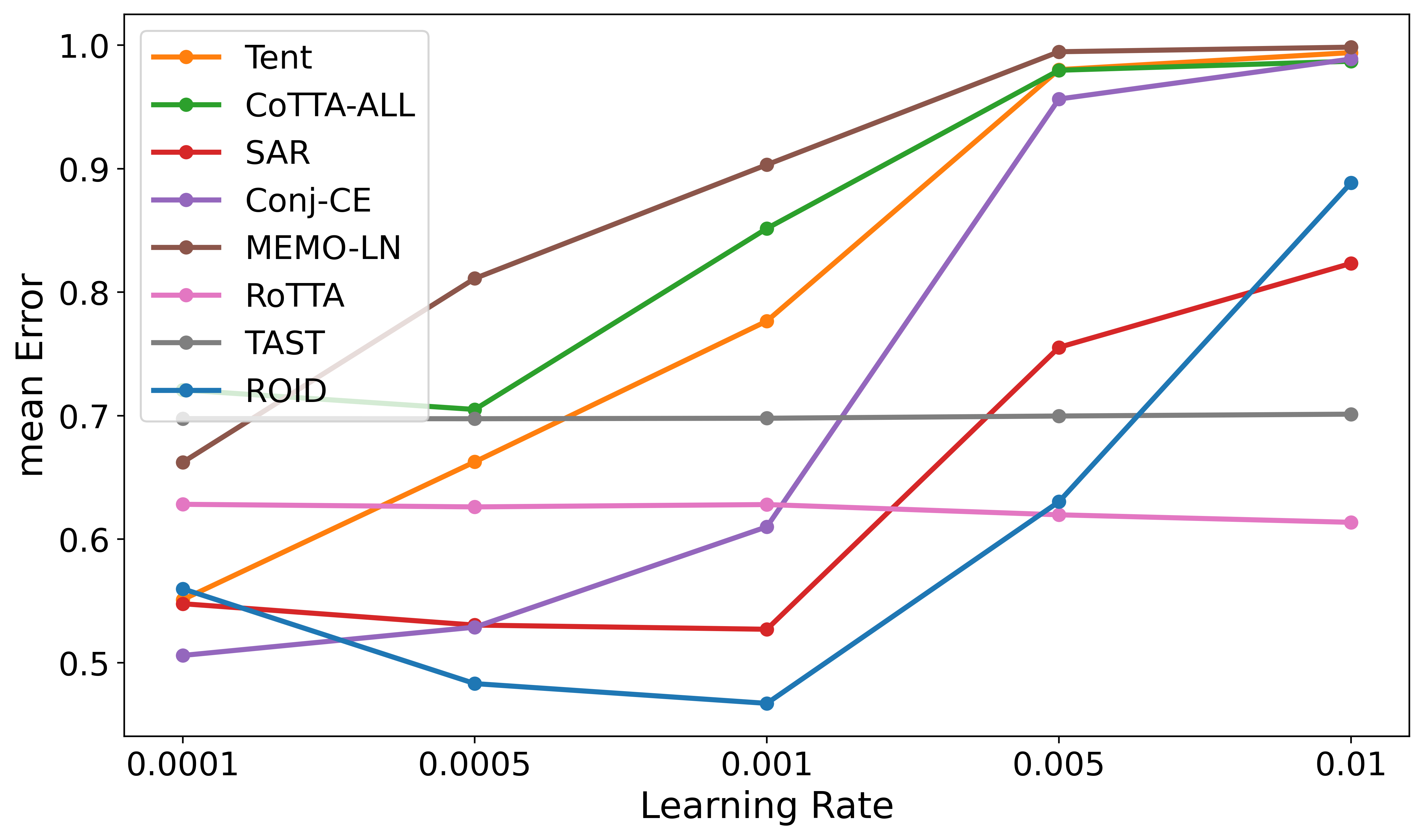}
    \caption{Impact of varying learning rates on ImageNet-C severity level $5$. }
    \label{fig:lr-ablation}
\end{figure}

To further explore the impact of hyperparameters, we examine the learning rate within the range [0.0001, 0.0005, 0.001, 0.005, 0.01] on the ImageNet-C dataset, as detailed in Figure \ref{fig:lr-ablation}. Results indicate that lower learning rates benefit online adaptation.

The empirical results shown in Fig. \ref{fig:lr-ablation} confirm a general idea: a high learning rate can cause quick overfitting of the current batch in online adaptation. Lower rates enhance model stability, enabling smoother pattern adaptation. Conversely, a large learning rate could disrupt the finely learned knowledge from the source model thus causing performance degradation. 

However, this is not universally applicable. Methods such as \texttt{RoTTA} and \texttt{TAST} consistently demonstrate stable performance across varying learning rates. By incorporating more information for prediction, these methods alleviate the impact of batch-specific label variations, thereby bolstering model stability. Moreover, \texttt{CoTTA-ALL}, \texttt{SAR}, and \texttt{ROID} attain their optimal performance at specific learning rate values, excluding the lowest one, underscoring the continued relevance of studying learning rates to optimize adaptation performance.

\subsubsection{Scheduler. }We examine the effectiveness of three different learning rate schedulers to ascertain their influence on performance. Compared with the default learning rate setup of 0.001 (Def) and a minimum learning rate of 0.0001 (Min) for the Adam optimizer, they are:
    \begin{itemize}
        \item ExponentialLR (Exp): divides the learning rate every iteration by the decay rate of 0.9, adjusting from the first domain to the last.
        \item CosineAnnealingWarmRestarts (Cos): periodically resets the learning rate to a higher value and then decreases it following a part of the cosine curve. Here, we gradually decrease the learning rate to 0.0001 for each domain, with the cycle length matching the number of iterations per domain.
        \item CycleLR (Cyc): cyclically varying the learning rate between two boundary values over a set number of training iterations. Here, we put it to begin with a maximum learning rate of 0.001 in each domain and decay to 0.0001 by the final iteration of the domain.
    \end{itemize}
    
 \begin{table}[h]
\centering
\caption{Variations of learning rate schedulers on ImageNet-C severity level 5 with batch size 16. }
% before iter 16000
\resizebox{0.9\linewidth}{!}{% 
\begin{tabular}{l | c c c c c}
\toprule
Method  & Exp & Cos & Cyc & Def & Min\\
\midrule
\texttt{Tent} & 57.01 & 72.07 & 69.30 & 77.66 & \textbf{55.10} \\
\texttt{CoTTA-ALL} & 77.23 & 82.99 & 77.48 & 85.16 & \textbf{72.08} \\ 
\texttt{SAR} & 58.47 & \textbf{52.43} & 53.99 & 52.70 & 54.77 \\
\texttt{Conj-CE} & 52.64 & 54.98 & 53.03 & 61.00 & \textbf{50.58} \\
\texttt{MEMO-LN} & \textbf{61.03} & 92.59 & 92.99 & 90.32 & 66.22 \\
\texttt{RoTTA} & 62.97 & 62.91 & 62.91 & \textbf{62.80} & 62.82 \\
\texttt{TAST} & 69.79 & 69.79 & 69.79 & 69.79 & \textbf{69.76} \\
\texttt{ROID} & 59.93 & 48.13 & 47.87 & \textbf{46.70} & 55.98\\
\bottomrule
\end{tabular}}
\label{table:scheduler}
\end{table}
% \vspace{-1em}
As shown in Table \ref{table:scheduler}, cyclical adjustments such as Cos and Cyc of \texttt{Tent} and \texttt{Conj-CE} can surpass the default setup (Def) but do not perform as well as the minimum learning rate (Min), indicating the need for a lower learning rate from the outset of the adaptation process. Meanwhile, \texttt{RoTTA} and \texttt{TAST} demonstrate very stable performance across different learning rate strategies, showcasing their robustness. Notably, \texttt{RoTTA} exceeds the source-only performance (i.e. 63.24$\%$) under any condition. Additionally, \texttt{ROID} demonstrates superior performance when utilizing the default learning rate without a scheduler.

One exception is \texttt{MEMO-LN}, where ExponentialLR (Exp) is the only one that competes with or outperforms the minimum learning rate setup. Furthermore, as illustrated in Fig. \ref{fig:lr-ablation}, \texttt{MEMO-LN} produces significant differences in performance among the schedulers and learning rates, indicating its high sensitivity of learning rate. 

\begin{table}[!h]
\centering
\caption{ViT \textit{v.s.} SwinTransformer on ImageNet-C severity level 5. } 
% before iter 16000
\resizebox{1\linewidth}{!}{% 
\begin{tabular}{l | c c c c c c c c c}
\toprule
  & SO & \texttt{Tent} & \texttt{CoTTA-ALL} & \texttt{SAR} & \texttt{Conj-CE} & \texttt{MEMO-LN} & \texttt{RoTTA} & \texttt{TAST} & \texttt{ROID}\\
\midrule
\multicolumn{10}{c}{\centering \textcolor{blue}{Batch size = 16}}\\
ViT & 63.24 & 77.66 & 85.16 & 52.70 & 61.00 & 90.32 & 62.80 & 69.79 & \textbf{46.70}\\
Swin & 54.61 & 78.48 & 88.80 & 55.44 & 83.77 & 85.21 & 54.38 & 65.53 & \textbf{31.75}\\
\midrule
\multicolumn{10}{c}{\centering \textcolor{blue}{Batch size = 1}}\\
ViT & 63.24 & 98.16 & 98.72 & 67.49 & 97.26 & 99.32 & \textbf{62.81} & 69.93 & 99.91\\
Swin & 54.61 & 96.98 & 99.35 & 56.18 & 95.61 & 99.02 & \textbf{54.39} & 65.84 & 99.90\\
\bottomrule
\end{tabular}}\vspace{-2.5ex}
\label{table:swin-vit}
\end{table}

\subsection{Is OTTA effective with other vision transformer variants?}

We further evaluate the effectiveness of selected OTTA methods on the SwinTransformer\footnote{swin\_base\_patch4\_window7\_224.ms\_in22k\_ft\_in1k} as a point of comparison to the foundational ViT, as shown in Table \ref{table:swin-vit}.

Switching the backbone architecture to SwinTransformer can greatly improve performance. This is evidenced by the direct inference of SwinTransformer, which outperforms all ViT outputs except for \texttt{SAR} and \texttt{ROID}. This emphasizes the crucial role of backbone architecture in achieving high performance and suggests that improvements in backbone designs can sometimes overshadow the enhancements brought about by advanced OTTA methodologies. The transition to a new backbone necessitates a reevaluation of the efficacy of individual OTTA methods. Notably, among those, the architecture-agnostic methods \texttt{RoTTA} (with the \texttt{RBN} module removed) and \texttt{ROID} outperform the source-only performance under a batch size of 16 using SwinTransformer. Furthermore, \texttt{RoTTA} maintains stability even with a batch size of 1, highlighting its robustness and adaptability across different architectures.

In light of these findings, there appears to be a dynamic interplay between backbone architectures and OTTA methods. Therefore, it is crucial to evaluate OTTA strategies in the context of rapid advancements in underlying transformer models.

\section{Future Directions}\label{sec:future}
Our initial evaluations of the Vision Transformer revealed that many Online Test-Time Adaptation methods are not fully optimized for this architecture, resulting in suboptimal outcomes. Based on these findings, we propose several key attributes for an ideal OTTA approach, suitable for future research and tailored to advanced architectures like ViT:

\begin{itemize}
    \item \textbf{Refining OTTA in Realistic Settings:} Future OTTA methods should undergo testing in realistic environments such as practical TTA \citep{DBLP:conf/wacv/MarsdenD024}, weaken the domain boundaries, employing advanced architectures, practical testbeds, and reasonable batch sizes. This approach aims to gain deeper and more relevant insights.
    \item \textbf{Addressing Multimodal Challenges and Exploring Prompting Techniques:} With the evolution towards foundation models like \texttt{CLIP} \citep{DBLP:conf/icml/RadfordKHRGASAM21}, OTTA is confronted with new challenges. These models may face shifts across various modalities, necessitating innovative OTTA strategies that extend beyond reliance on images alone. Exploring prompt-based methods could offer significant breakthroughs in OTTA.
    \item \textbf{Hot-swappable OTTA:} Keeping pace with the rapid evolution of backbone architectures is crucial. Future OTTA methods should focus on adaptability and generalizability to seamlessly integrate with evolving architectures.
    \item \textbf{Stable and Robust Optimization for OTTA:} Stability and robustness in optimization remain paramount. Given that larger batch sizes demonstrate limited effectiveness in ViT, future research should investigate more universal optimization improvements. Such advancements aim to consistently enhance model performance, independent of external factors like batch size.
\end{itemize}

\section{Conclusion}\label{sec:conclude}
In this survey, we thoroughly examine Online Test-Time Adaptation (OTTA), detailing existing methods, relevant datasets, benchmarks for evaluation, and their implementations. Comprehensive experiments are conducted to evaluate the effectiveness and efficiency of current OTTA methods when applied to vision transformers. Our observations indicate that noise-synthesized domain shifts often present more significant challenges compared with other types of shifts, such as those encountered in real-world scenarios or diffusion environments. Additionally, the presence of a large number of classes in a dataset can lead to noticeable discrepancies between batches, potentially impacting OTTA models' ability to retain consistent knowledge. This may result in learning difficulties and an increased risk of severe forgetting. To address these challenges, we find that updating the normalization layer with a memory bank or optimization flatness, combined with proper batch size selection, can effectively stabilize the adaptation process and mitigate forgetting. We hope this survey could serve as a foundational reference, providing valuable insights for researchers and practitioners keen to explore the dynamic field of OTTA.
\\

\noindent \textbf{Data Availability Statement:} 
The data that support the findings of this study are openly available as summarized in Table \ref{table:datasets}. This open-sourced repository\footnote{\url{https://github.com/Jo-wang/OTTA_ViT_survey}} includes all the implementation code and experimental configurations of this work. 

\balance{
\bibliographystyle{spbasic}
\bibliography{main.bib}}
%% If required, the content of the .bbl file can be included here once the bbl is generated

\begin{table*}[!htb]
% \makeatletter\def@captype{table}\makeatother
\caption{The summary of existing OTTA published in the top-tier conferences before Dec-2023. }\label{tab. otta_methods}
\small
\resizebox{1\linewidth}{!}{% 
\begin{tabular}{@{\extracolsep\fill}p{0.4cm}  p{4.7cm}  c c c c c  c c c c c}
    \toprule
        Year & Method & \multicolumn{3}{c}{\centering Model-based} & \multicolumn{2}{c}{\centering Data-based}  & \multicolumn{5}{c}{\centering Optimization-based} \\
         \cmidrule(lr){3-5}\cmidrule(lr){6-7}\cmidrule(lr){8-12} &  & \multicolumn{1}{c}{\centering add. } & \multicolumn{1}{c}{\centering subst. }  & \multicolumn{1}{c}{\centering prompt} & \multicolumn{1}{c}{\centering mem. } & \multicolumn{1}{c}{\centering aug. } & \multicolumn{1}{c}{\centering loss. }& \multicolumn{1}{c}{\centering pl.} & \multicolumn{1}{c}{\centering t-s. } & \multicolumn{1}{c}{\centering norm. } & \multicolumn{1}{c}{\centering other} \\
       % \centering \cmidrule(lr){3-5}\cmidrule(lr){6-7}\cmidrule(lr){8-11}
       %     & &add. &subst. &prompt &mem. &aug. &loss. &pl. &t-s. &norm. \\
        \midrule
         2021 &\texttt{Tent} \citep{DBLP:conf/iclr/WangSLOD21} & & & & & & \checkmark & & & \checkmark & \\
         2021 &\texttt{MixNorm} \citep{DBLP:journals/corr/abs-2110-11478} & & \checkmark & & & & & & &\checkmark  & \\
         2021 &\texttt{PAD} \citep{wu2021domain} & & \checkmark & & & \checkmark & & \checkmark & &  & \\
         2021 &\texttt{TTPR} \citep{DBLP:journals/corr/abs-2110-10232} & & & & & \checkmark & \checkmark & & &  & \\
         2021 &\texttt{Core} \citep{DBLP:journals/corr/abs-2110-04065} & & & & & & & & & \checkmark &\\
         2021 &\texttt{SLR} \citep{mummadi2021test} & \checkmark & & & & & \checkmark & & & \checkmark &\\
         2021 &\texttt{T3A} \citep{DBLP:conf/nips/IwasawaM21} & &\checkmark & & \checkmark & & & & & & \\
         2022 &\texttt{NOTE} \citep{DBLP:conf/nips/GongJKKSL22} & & & & \checkmark & & & & & \checkmark &\\
         2022 &\texttt{CoTTA} \citep{DBLP:conf/cvpr/0013FGD22} & & & & & \checkmark & & & \checkmark &  &\\
         2022 &\texttt{Conj-PL} \citep{DBLP:conf/nips/GoyalSRK22} & & & & & & \checkmark & \checkmark & & \checkmark &\\
        % 2022 &\texttt{LAME} \citep{DBLP:conf/cvpr/BoudiafMAB22} & & & &  & & & & \\
        2022 &\texttt{GpreBN} \citep{DBLP:journals/corr/abs-2205-10210} & & \checkmark & & & & & & & \checkmark &\\
        2022 &\texttt{CFA} \citep{DBLP:conf/ijcai/KojimaMI22} & & & & & & & & & \checkmark  &\\
        2022 &\texttt{DUA} \citep{DBLP:conf/cvpr/MirzaMPB22a} & & \checkmark & & & \checkmark & & & & \checkmark  &\\
        2022 &\texttt{MEMO} \citep{DBLP:conf/nips/ZhangLF22} & & & & & \checkmark & \checkmark & & & \checkmark &\\
        2022 &\texttt{AdaContrast} \citep{DBLP:conf/cvpr/0001WDE22}  & & & & \checkmark & \checkmark & \checkmark & \checkmark & & &\\
        2022 & \texttt{TPT} \citep{DBLP:conf/nips/ShuNHYGAX22} & & & \checkmark & & & & & & &\\
        2022 & \texttt{DePT} \citep{DBLP:journals/corr/abs-2210-04831} & & & \checkmark & & & & & & & \\
        2022 & \texttt{LAME} \citep{DBLP:conf/cvpr/BoudiafMAB22} & & & & & & & & & & \checkmark\\
        2023 & \texttt{DN} \citep{DBLP:conf/aaai/GanBLMZSL23}  & & & \checkmark & & & & & & & \\
        % \hline
        2023 & \texttt{TAST} \citep{DBLP:conf/iclr/JangCC23} & \checkmark & \checkmark & & \checkmark & & & & & \checkmark & \\
        2023 & \texttt{MECTA} \citep{DBLP:conf/iclr/HongLZS23} & & \checkmark & & & & & & & \checkmark &\\
        2023 & \texttt{DELTA} \citep{DBLP:conf/iclr/Zhao0X23} & & \checkmark & & & & & & & \checkmark &\\
        2023 & \texttt{TeSLA} \citep{DBLP:journals/corr/abs-2303-09870} & & & & \checkmark & \checkmark & \checkmark & & \checkmark & &\\
        2023 & \texttt{RoTTA} \citep{DBLP:conf/cvpr/YuanX023} &  & \checkmark & & \checkmark & \checkmark & & & \checkmark & \checkmark &\\
        2023 & \texttt{EcoTTA} \citep{DBLP:journals/corr/abs-2303-01904} & \checkmark & & & & & & & & \checkmark &\\
        2023 & \texttt{TIPI} \citep{nguyen2023tipi} & & & & & \checkmark & & & & \checkmark &\\
        2023 & \texttt{TSD} \citep{DBLP:journals/corr/abs-2303-10902} & & \checkmark &  & \checkmark & & \checkmark & & & &\\
        2023 & \texttt{SAR} \citep{DBLP:conf/iclr/Niu00WCZT23} & & & & & & \checkmark & & & \checkmark &\\
        2023 & \texttt{ECL} \citep{DBLP:journals/corr/abs-2301-06013} & & & & \checkmark & & & \checkmark & & & \\
        2023 & \texttt{REALM} \citep{DBLP:journals/corr/abs-2309-03964} & & & & & & \checkmark & & & \checkmark &\\ 
        2023 & \texttt{TTC} \citep{DBLP:journals/corr/abs-2310-20327} & & & & & \checkmark & \checkmark & & & &\\
        2023 & \texttt{SoTTA} \citep{DBLP:journals/corr/abs-2310-10074} & & & & \checkmark & & \checkmark & & & \checkmark &\\
        2023 & \texttt{ViDA} \citep{DBLP:journals/corr/abs-2306-04344} & \checkmark & & & & \checkmark & & & \checkmark & &\\ 
        2023 & \texttt{ERSK} \citep{DBLP:journals/corr/abs-2311-04991} & & \checkmark & & & & & & & \checkmark & \\ 
        2023 & \texttt{ROID} \citep{DBLP:conf/wacv/MarsdenD024} & & & & & \checkmark & \checkmark & & & \checkmark & \checkmark\\
  \bottomrule
\end{tabular}}
\end{table*}

\newpage
\appendix
\section{Appendix}
This appendix offers tables about the categorization of OTTA methods and experiments conducted on the eight selected methods across the chosen datasets.

\begin{figure}[!ht]
\centering
\includegraphics[width=0.9\linewidth]{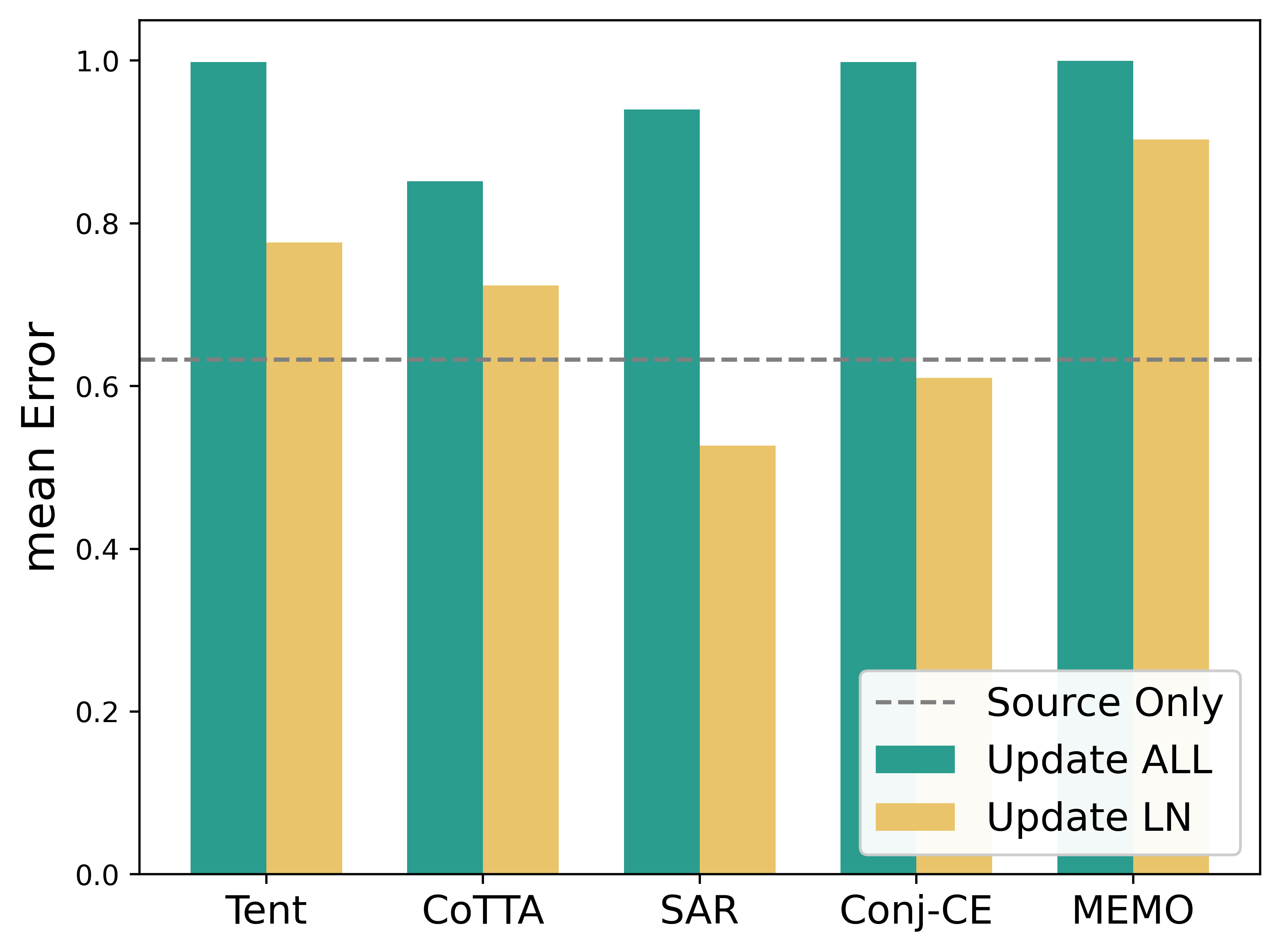}
\caption{Fully update v.s. \texttt{LayerNorm} updates on ImageNet-C severity level $5$ with batch size 16.} 
\label{fig:layer-ablation}
\end{figure}

\begin{itemize}
    \item Table \ref{tab. otta_methods}: The summary of existing OTTA methods. 
    \item Table \ref{tab. cifar-10-c-exp}: Experiments on the CIFAR-10 $\rightarrow$ CIFAR-10-C task.
    \item Table \ref{tab:vit_base_16_224}: Experiments on the ImageNet-1k $\rightarrow$ ImageNet-C task.
    \item Table \ref{tab. cifar-w-exp}: Experiments on the CIFAR-10 $\rightarrow$ CIFAR-10-Warehouse Google split task.
    \item Fig. \ref{fig:layer-ablation}: Ablation study on \texttt{LayerNorm} update strategy.
\end{itemize}

\begin{table*}[!ht]\caption{Classification error rate ($\%$) for the standard CIFAR-10 $\rightarrow$ CIFAR-10-C/CIFAR 10.1 online test-time adaptation task. Results are evaluated on ViT-base-patch16-224 with the largest corruption severity level 5. Here, SO denotes source only.}\label{tab. cifar-10-c-exp}
    \centering
    \resizebox{1\linewidth}{!}{% 
    \begin{tabular}{l cccc cccc cccc ccccc }
    \toprule
        Method  & Glass & Fog &Defoc. &Impul. &Contr. &Gauss. &Elastic. &Zoom &Pixel &Frost &Snow &JPEG &Motion &Shot &Brit. &Mean  & 10.1\\%& GFLOPs \\ 
        \midrule
        SO & 34.60 &14.14 & 12.20 & 39.47 & 38.95 & 60.12 & 14.70 & 10.61 & 25.26 & 10.39 & 7.67 & 16.51 & 15.30 
 & 55.17 & 3.26 & 23.89 & 7.50\\%& 269.81  \\ 
    \midrule
    \multicolumn{17}{c}{\centering \textcolor{blue}{Batch size = 16}}\\
    \texttt{Tent} &49.02 & 6.81 & 5.29 & 65.93 & 11.91 & 84.93 & 10.96 & 5.12 & 10.41 & 6.68 & 5.96 & 14.21 & 8.25 & 85.60 & 2.43 & 24.90 & 7.60\\%& 269.81 \\ 
    \texttt{CoTTA-LN} & 34.50 & 14.57 & 12.97 & 43.96 & 50.01 & 58.04 & 15.05 & 11.09 & 22.92 & 10.43 & 7.75 & 16.46 & 15.92 & 53.65 & 3.27 & 24.71 & 7.50\\%& 809.42 \\ 
    \texttt{CoTTA-ALL} & 71.68 & 51.28 & 38.25 & 82.60 & 79.30 & 85.57 & 31.22 & 16.51 & 39.95 & 36.75 & 37.38 & 37.37 & 81.00 & 88.54 & 9.72 & 52.47 & 11.55\\%& 809.42 \\ 
    \texttt{CoTTA$*$-LN} & 34.58 & 14.60 & 13.02 & 43.71 & 49.71 & 58.00 & 15.03 & 11.09 & 22.96 & 10.45 & 7.67 & 16.53 & 15.93 & 53.68 & 3.24 & 24.68 & 7.50\\%& 809.42 \\ 
    \texttt{CoTTA$*$-ALL} & 77.17 & 78.07 & 70.48 & 85.98 & 81.41 & 86.57 & 74.54 & 66.80 & 68.95 & 78.34 & 72.30 & 79.23 & 79.62 & 88.83 & 67.91 & 77.08 & 16.40\\%& 809.42 \\ 
    \texttt{SAR} & 34.39 & 14.13 & 12.18 & 39.38 & 34.92 & 60.67 & 14.70 & 10.61 & 25.17 & 10.39 & 7.67 & 16.51 & 15.28 & 55.59 & 3.26 & 23.66 & 7.50\\%& 539.61 \\
    \texttt{Conj-CE} & 61.26 & 6.36 & 4.89 & 80.12 & 7.94 & 87.96 & 10.92 & 4.20 & 8.43 & 5.32 & 5.33 & 12.71 & 7.18 & 86.63 & 2.27 & 26.10 & 7.35\\%& 539.61\\ 
    \texttt{MEMO-LN} & 26.15 & 9.08 & 6.17 & 65.80 & 9.58 & 77.68 & 12.48 & 5.29 & 9.14 & 6.59 & 5.71 & 14.45 & 7.71 & 59.15 & 2.49 & 21.16 & 7.80\\%& 809.42\\ 
    \texttt{RoTTA} & 34.47 & 13.72 & 11.70 & 38.72 & 37.35 & 59.55 & 14.29 & 10.14 & 25.38 & 10.25 & 7.42 & 16.31 & 14.70 & 54.59 & 3.07 & 23.44 & 7.50\\%& 269.81 \\ 
    \texttt{TAST} & 55.06 & 58.89 & 52.80 & 57.12 & 81.54 & 54.39 & 37.02 & 43.68 & 33.65 & 32.28 & 24.21 & 25.69 & 54.63 & 47.69 & 14.87 & 44.90 & 14.85\\%& 268.45 \\
    \texttt{ROID} & 23.31 & 8.39 & 7.00 & 21.91 & 12.59 & 29.36 & 11.58 & 6.00 & 14.11 & 7.07 & 6.01 & 13.52 & 9.10 & 28.06 & 2.40 & 13.36 & 7.70 \\
        \midrule
    \multicolumn{17}{c}{\centering \textcolor{blue}{Batch size = 1}}\\
    \texttt{Tent} & 70.90 & 55.50 & 39.38 & 87.28 & 78.20 & 77.92 & 25.22 & 30.77 & 26.38 & 71.19 & 11.85 & 57.65 & 69.19 & 76.82 & 5.48 & 52.25 & 9.05\\%& 16.86  \\
    \texttt{CoTTA-LN} & 34.09 & 16.63 & 14.79 & 43.05 & 45.14 & 57.63 & 15.39 & 13.23 & 21.11 & 10.15 & 8.25 & 16.88 & 17.68 & 52.36 & 4.07 & 24.70 & 7.30\\%& 50.59  \\
    \texttt{CoTTA-ALL} & 87.81 & 89.90 & 86.13 & 89.17 & 88.82 & 89.55 & 84.91 & 87.77 & 86.30 & 87.42 & 89.17 & 88.67 & 89.06 & 89.22 & 81.65 & 87.70 & 68.25\\%& 50.59 \\
    \texttt{CoTTA$*$-LN} & 35.88 & 18.35 & 16.96 & 46.27 & 47.31 & 58.21 & 17.49 & 15.12 & 20.95 & 10.78 & 9.60 & 18.15 & 19.72 & 52.15 & 4.83 & 26.12 & 7.30\\%& 50.59 \\
    \texttt{CoTTA$*$-ALL} & 89.79 & 89.76 & 86.30 & 89.63 & 89.66 & 89.78 & 89.17 & 89.12 & 89.32 & 88.04 & 89.44 & 89.18 & 88.76 & 89.67 & 87.83 & 89.03 & 79.20\\%& 50.59 \\
    \texttt{SAR} & 34.63 & 14.04 & 12.03 & 39.55 & 37.62 & 61.17 & 14.63 & 10.61 & 24.85 & 10.38 & 7.62 & 16.63 & 15.20 & 55.67 & 3.27 & 23.86 & 7.45\\%& 33.73 \\
    \texttt{Conj-CE} & 48.90 & 55.11 & 14.09 & 88.28 & 84.49 & 89.11 & 14.39 & 29.01 & 20.95 & 13.70 & 27.84 & 38.25 & 12.71 & 88.95 & 5.43 & 42.08 & 9.00\\%& 33.73 \\
    \texttt{MEMO-LN} & 85.91 & 5.55 & 4.77 & 87.84 & 3.76 & 89.33 & 11.54 & 3.70 & 5.52 & 4.08 & 4.36 & 12.28 & 6.01 & 89.32 & 1.65 & 27.71 & 7.80\\%& 556.48 \\
    \texttt{RoTTA} & 34.46 & 13.72 & 11.71 & 38.72 & 37.36 & 59.55 & 14.29 & 10.12 & 25.39 & 10.24 & 7.43 & 16.32 & 14.71 & 54.62 & 3.07 & 23.45 & 7.50\\%& 16.86 \\
    \texttt{TAST} & 55.16 & 61.79 & 52.69 & 56.94 & 82.01 & 54.51 & 36.85 & 43.73 & 33.87 & 32.23 & 24.27 & 25.80 & 55.25 & 47.80 & 14.88 & 45.19 & 14.85\\%& 16.78 \\
    \texttt{ROID} & 89.99 & 89.99 & 89.99 & 89.99 & 89.99 & 89.99 & 89.99 & 89.99 & 89.99 & 89.99 & 89.99 & 89.99 & 89.99 & 89.99 & 89.99 & 89.99 & 89.95\\
        \bottomrule
    \end{tabular}}
\end{table*}

\begin{table*}
  \caption{Classification error rate ($\%$) for the standard ImageNet-1k $\rightarrow$ ImageNet-C online test-time adaptation task. Results are
evaluated on ViT-base-patch16-224 with the largest corruption severity level 5. Here, SO means source only.}
  \label{tab:vit_base_16_224}
  \resizebox{1\linewidth}{!}{% 
    \begin{tabular}{l cccc cccc cccc cccc}
    \toprule
    ImageNet-C & Glass & Fog &Defoc. &Impul. &Contr. &Gauss. &Elastic. &Zoom &Pixel &Frost &Snow &JPEG &Motion &Shot &Brit. &Mean \\
    \midrule
    SO & 73.92 & 63.76 & 65.54 & 79.18 & 89.84 & 79.88 & 53.38 & 62.42 & 39.26 & 60.70 & 63.66 & 45.06 & 61.20 & 77.38 & 33.38 & 63.24\\
     \midrule
    \multicolumn{17}{c}{\centering \textcolor{blue}{Batch size = 16}}\\
    \texttt{Tent} & 77.86 & 97.88 & 73.88 & 97.24 & 97.28 & 95.28 & 57.54 & 85.92 & 33.12 & 97.64 & 96.06 & 42.40 & 92.60 & 90.88 & 29.30 & 77.66\\
    \texttt{CoTTA-ALL} & 73.72 & 94.82 & 92.80 & 99.14 & 92.94 & 99.06 & 78.06 & 84.14 & 79.40 & 88.92 & 96.70 & 54.48 & 86.58 & 98.10 & 58.48 & 85.16\\
    % \hline  % A: update all, P: parameter reset, L: update LN, N: No parameter reset
    \texttt{CoTTA-LN} & 77.06 & 68.50 & 75.40 & 92.10 & 87.60 & 92.06 & 63.28 & 67.24 & 50.28 & 71.36 & 80.74 & 52.46 & 71.58 & 89.46 & 46.40 & 72.37\\
    % \hline
    \texttt{CoTTA$*$-ALL} & 90.30 & 95.84 & 95.08 & 99.28 & 95.24 & 99.12 & 87.90 & 88.20 & 90.20 & 94.84 & 97.50 & 68.04 & 95.82 & 98.68 & 62.08 & 90.54 \\
    % \hline
    \texttt{CoTTA$*$-LN} & 77.26 & 68.60 & 75.32 & 91.58 & 87.08 & 92.28 & 63.30 & 67.18 & 50.72 & 70.74 & 80.64 & 52.02 & 71.48 & 89.72 & 46.46 & 72.29 \\
    \texttt{SAR} & 52.78 & 48.34 & 54.74 & 71.92 & 85.74 & 73.18 & 37.54 & 46.48 & 31.70 & 61.14 & 49.24 & 36.74 & 48.48 & 64.92 & 27.52 & 52.70 \\
    \texttt{Conj-CE} & 68.30 & 42.96 & 76.74 & 93.70 & 84.52 & 96.60 & 34.46 & 45.64 & 31.00 & 83.90 & 55.12 & 35.28 & 53.32 & 84.88 & 28.60 & 61.00 \\
    % \hline
    \texttt{Conj-Poly} & 68.26 & 47.48 & 54.08 & 93.68 & 75.82 & 96.68 & 34.08 & 47.42 & 29.94 & 81.26 & 65.06 & 36.06 & 68.36 & 87.70 & 29.16 & 61.00 \\
    \texttt{MEMO-LN} & 96.10 & 95.62 & 96.26 & 98.68 & 99.18 &	98.08 & 97.48 & 87.80 & 35.40 & 93.58 & 96.30 & 83.56 & 94.50 & 98.26 & 84.06 & 90.32 \\
    % \hline
    \texttt{RoTTA} & 73.50 & 63.24 & 65.14 & 78.40 & 89.58 & 79.02 & 53.10 & 61.94 & 39.12 & 60.34 & 63.24 & 44.62 & 60.80 & 76.88 & 33.12 & 62.80\\
    \texttt{TAST} & 80.60 & 73.16 & 75.26 & 82.22 & 94.32 & 81.68 & 60.98 & 71.34 & 47.48 & 67.62 & 71.34 & 48.98 & 71.68 & 79.66 & 40.46 & 69.79 \\
    \texttt{ROID} & 51.06 & 43.02 & 50.68 & 61.02 & 56.70 & 61.94 & 36.04 & 46.02 & 30.66 & 48.44 & 47.30 & 35.88 & 46.70 & 58.38 & 26.72 & 46.70\\
    \hline
    \multicolumn{17}{c}{\centering \textcolor{blue}{Batch size = 1}}\\
    \texttt{Tent} & 99.66 & 99.46 & 97.90 & 99.64 & 99.84 & 99.80 & 98.16 & 98.60 & 97.56 & 99.28 & 97.46 & 96.90 & 98.18 & 99.66 & 90.34 & 98.16\\
    \texttt{CoTTA-ALL} & 99.24 & 99.64 & 99.06 & 99.54 & 99.78 & 99.80 & 97.94 & 99.20 & 97.48 & 98.08 & 99.20 & 98.60 & 99.46 & 99.02 & 94.82 & 98.72\\
    % \hline  % A: update all, P: parameter reset, L: update LN, N: No parameter reset
    \texttt{CoTTA-LN} & 80.38 & 69.62 & 82.82 & 96.40 & 90.80 & 95.30 & 65.80 & 67.46 & 46.52 & 72.82 & 81.56 & 51.42 & 71.74 & 93.66 & 42.28 & 73.91\\
    % \hline
    \texttt{CoTTA$*$-ALL} & 99.62 & 99.70 & 99.54 & 99.60 & 99.82 & 99.80 & 98.82 & 99.48 & 99.30 & 99.28 & 99.58 & 99.20 & 99.60 & 99.64 & 98.70 & 99.45\\
    % \hline
    \texttt{CoTTA$*$-LN} & 82.84 & 72.14 & 84.56 & 94.98 & 92.66 & 94.66 & 66.16 & 70.64 & 48.88 & 78.10 & 83.78 & 51.62 & 76.08 & 91.92 & 45.12 & 75.61\\
    \texttt{SAR} & 70.58 & 66.64 & 72.40 & 83.60 & 97.76 & 82.96 & 57.54 & 62.98 & 41.02 & 64.54 & 78.52 & 46.34 & 66.96 & 82.72 & 37.78 & 67.49\\
    \texttt{Conj-CE} & 99.38 & 96.42 & 98.18 & 99.76 & 99.74 & 99.76 & 91.54 & 96.02 & 96.74 & 98.86 & 97.86 & 96.90 & 98.26 & 99.14 & 90.38 & 97.26\\
    \texttt{MEMO-LN} & 99.82 & 99.58 & 99.78 & 99.84 & 99.72 & 99.90 & 99.54 & 99.30 & 98.04 & 99.66 & 99.48 & 98.46 & 99.52 & 99.80 &97.30 & 99.32  \\
    % \hline
    \texttt{RoTTA} & 73.46 & 63.26 & 65.10 & 78.40 & 89.58 & 79.04 & 53.14 & 61.94 & 39.12 & 60.36 & 63.18 & 44.64 & 60.86 & 76.88 & 33.12 & 62.81 \\
    \texttt{TAST} & 80.66 & 73.44 & 75.50 & 82.26 & 94.40 & 81.74 & 60.96 & 71.60 & 47.64 & 67.96 & 71.60 & 49.06 & 72.04 & 79.62 & 40.48 & 69.93 \\
    \texttt{ROID} & 99.90 & 99.92 & 99.90 & 99.92 & 99.92 & 99.92 & 99.90 & 99.90 & 99.90 & 99.90 & 99.90 & 99.90 & 99.90 & 99.92 & 99.90 & 99.91 \\
    \bottomrule
    \end{tabular}}
\end{table*}
\begin{table*}[!ht]\caption{Classification error rate ($\%$) for the CIFAR-10 $\rightarrow$ CIFAR-10-Warehouse online test-time adaptation task. Results are
evaluated on ViT-base-patch16-224 with the Google split and diffusion split. Here, SO means source only} \label{tab. cifar-w-exp}
    \centering
    \resizebox{1\linewidth}{!}{% 
    \begin{tabular}{l cccc cccc cccc c}
    \toprule
        Google & G-01 & G-02 & G-03 & G-04 & G-05 & G-06 & G-07 & G-08 & G-09 & G-10 & G-11 & G-12 & Mean  \\ 
        \midrule
        SO & 20.41 & 12.42 & 13.39 & 11.53 & 19.80 & 17.74 & 26.51 & 26.4 & 23.12 & 17.36 & 15.61 & 11.37 & 17.97 \\
 \midrule
    \multicolumn{14}{c}{\centering \textcolor{blue}{Batch size = 16}}\\
    \texttt{Tent} & 19.23 & 11.08 & 13.59 & 11.07 & 18.70 & 17.07 & 27.25 & 26.63 & 25.17 & 16.80 & 15.35 & 9.80 & 17.65\\
    \texttt{CoTTA-LN} & 20.52 & 12.50 & 13.56 & 11.43 & 20.09 & 18.02 & 27.29 & 26.87 & 23.36 & 17.93 & 15.73 & 11.43 & 18.23 \\
    \texttt{CoTTA-ALL} & 22.28 & 21.89 & 29.76 & 10.55 & 22.47 & 20.96 & 40.86 & 44.11 & 41.56 & 26.07 & 31.80 & 24.21 & 28.04\\
    \texttt{CoTTA$*$-LN} & 20.55 & 12.48 & 13.54 & 11.47 & 20.04 & 17.93 & 27.25 & 26.71 & 23.30 & 17.81 & 15.88 & 11.41 & 18.20\\
    \texttt{CoTTA$*$-ALL} & 40.13 & 60.93 & 61.92 & 33.15 & 50.54 & 48.50 & 56.21 & 58.46 & 61.05 & 61.48 & 59.86 & 63.37 & 54.63 \\
    \texttt{SAR} & 20.41 & 12.42 & 13.39 & 11.53 & 19.80 & 17.74 & 26.51 & 26.40 & 23.15 & 17.36 & 15.61 & 11.37 & 17.97 \\
    \texttt{Conj-CE} & 17.91 & 9.87 & 11.51 & 10.11 & 20.18 & 16.41 & 27.22 & 26.40 & 24.96 & 14.89 & 13.57 & 9.30 & 16.86 \\
    \texttt{MEMO-LN} & 20.70 & 10.84 & 12.61 & 10.88 & 19.08 & 16.46 & 28.11 & 26.87 & 20.91 & 17.21 & 14.99 & 10.83 & 17.46 \\
    \texttt{MEMO-ALL} & 74.62 & 90.31 & 91.95 & 89.39 & 92.81 & 92.49 & 90.51 & 92.71 & 87.48 & 92.28 & 89.94 & 90.07 & 89.55 \\
    \texttt{RoTTA} & 20.29 & 12.37 & 13.39 & 11.51 & 19.73 & 17.58 & 26.40 & 26.40 & 23.09 & 17.30 & 15.52 & 11.35 & 17.91 \\
    \texttt{TAST} & 21.30 & 16.37 & 14.26 & 13.71 & 23.24 & 20.89 & 28.11 & 28.50 & 27.30 & 23.21 & 18.90 & 16.13 & 20.99 \\
    \texttt{ROID} & 18.02 & 10.76 & 12.06 & 10.55 & 17.70 & 15.11 & 24.55 & 25.56 & 19.81 & 15.41 & 13.93 & 9.90 & 16.11\\
        \midrule
    \multicolumn{14}{c}{\centering \textcolor{blue}{Batch size = 1}}\\
    \texttt{Tent} & 25.90 & 44.81 & 15.22 & 12.77 & 21.57 & 65.91 & 26.55 & 32.14 & 60.05 & 59.22 & 33.68 & 18.70 & 34.71 \\
    \texttt{CoTTA-LN} & 20.44 & 12.99 & 13.76 & 11.78 & 21.07 & 19.05 & 27.22 & 27.27 & 24.20 & 18.30 & 15.90 & 11.83 & 18.65 \\
    \texttt{CoTTA-ALL} & 84.36 & 74.83 & 71.15 & 83.16 & 78.50 & 87.64 & 81.61 & 80.46 & 84.98 & 85.87 & 83.04 & 82.22 & 81.49 \\
    \texttt{CoTTA$*$-LN} & 20.70 & 13.50 & 14.11 & 11.97 & 21.35 & 19.31 & 27.29 & 28.18 & 24.57 & 18.49 & 15.73 & 12.33 & 18.96 \\
    \texttt{CoTTA$*$-ALL} & 86.81 & 85.35 & 86.21 & 87.08 & 88.44 & 87.41 & 72.75 & 90.88 & 84.03 & 89.93 & 85.44 & 85.36 & 85.81 \\
    \texttt{SAR} & 20.29 & 12.40 & 13.46 & 11.45 & 19.82 & 17.70 & 26.66 & 26.24 & 23.17 & 17.30 & 15.64 & 11.33 & 17.95 \\
    \texttt{Conj-CE} & 24.63 & 11.08 & 14.94 & 9.37 & 19.58 & 47.47 & 23.47 & 43.52 & 52.68 & 19.84 & 32.57 & 12.52 & 25.97 \\
    \texttt{MEMO-LN} & 15.35 & 9.53 & 11.48 & 8.18 & 27.78 & 21.33 & 21.99 & 43.48 & 70.91 & 12.14 & 20.35 & 8.06 & 22.55 \\
    \texttt{MEMO-ALL} & 74.62 & 90.96 & 91.95 & 89.39 & 92.81 & 92.49 & 97.44 & 92.75 & 87.48 & 92.28 & 89.94 & 90.07 & 90.18 \\
    \texttt{RoTTA} & 20.29 & 12.37 & 13.36 & 11.51 & 19.73 & 17.56 & 26.44 & 26.40 & 23.09 & 17.30 & 15.52 & 11.37 & 17.91 \\
    \texttt{TAST} & 21.41 & 16.34 & 14.21 & 13.67 & 23.24 & 20.84 & 28.11 & 29.09 & 27.49 & 23.32 & 18.90 & 16.08 & 21.06 \\
    \texttt{ROID} & 96.87 & 97.85 & 95.11 & 96.92 & 92.81 & 94.33 & 97.48 & 92.87 & 91.85 & 94.17 & 96.16 & 95.94 & 95.20\\
        \bottomrule
    \end{tabular}}
\end{table*}

\end{sloppypar}
\end{document}